AUGMENTED REINFORCEMENT LEARNING FRAMEWORK FOR ENHANCING DECISION-MAKING IN
MACHINE LEARNING MODELS USING EXTERNAL AGENTS

SANDESH KUMAR SINGH

Final Thesis

MASTERS OF COMPUTER SCIENCE

DECEMBER 2024

## Dedication

I dedicate this work to my family, who has been my source of strength, grace and wisdom throughout my life.

## Acknowledgement

I would like to thank my supervisor **Dr. Poonam Choudhari**, for continuous support throughout my research journey. Her direction, willingness to answer my questions is unparallel. She helped me at each step to complete my thesis on time.

I would also like to thank the faculty members of **Liverpool John Moores University, UK**. Their lectures, knowledge, and insights helped me in my thesis. The online resources shared by them was excellent, helping me in every pitfall while performing the research. I would like to thank the EdTech partner **upGrad** for their valuable support during my journey to Master's degree.

I would also like to thank my current organisation **Nucleus Software Exports Ltd**. Their support in crucial moments helped me a lot while performing this research.

# Abstract

This work proposes a novel technique Augmented Reinforcement Learning framework for the improvement of decision-making capabilities of machine learning models. The introduction of agents as external overseers checks on model decisions. The external agent can be anyone, like humans or automated scripts, that helps in decision path correction. It seeks to ascertain the priority of the "Garbage-In, Garbage-Out" problem that caused poor data inputs or incorrect actions in reinforcement learning.

The ARL framework incorporates two external agents that aid in course correction and the guarantee of quality data at all points of the training cycle. The External Agent 1 is a real-time evaluator, which will provide feedback light of decisions taken by the model, identify suboptimal actions forming the Rejected Data Pipeline. The External Agent 2 helps in selective curation of the provided feedback with relevance and accuracy in business scenarios creates an approved dataset for future training cycles.

The validation of the framework is also applied to a real-world scenario, which is "Document Identification and Information Extraction". This problem originates mainly from banking systems, but can be extended anywhere. The method of classification and extraction of information has to be done correctly here. Experimental results show that including human feedback significantly enhances the ability of the model in order to increase robustness and accuracy in making decisions.

The augmented approach, with a combination of machine efficiency and human insight, attains a higher learning standard-mainly in complex or ambiguous environments. The findings of this study show that human-in-the-loop reinforcement learning frameworks such as ARL can provide a scalable approach to improving model performance in data-driven applications.

# TABLE OF CONTENTS

# List of Tables

# List of Figures

# List of Abbreviations

| | |
|---|---|
| AI | Artificial Intelligence |
| ARL | Augmented Reinforcement Learning |
| CNN | Convolutional Neural Network |
| EDA | Exploratory Data Analysis |
| GPU | Graphics Processing Unit |
| MARL | Multi-Agent Reinforcement Learning |
| ML | Machine Learning |
| OCR | Optical Character Recognition |
| RAM | Random Access Memory |
| RL | Reinforcement Learning |
| RNN | Recurrent Neural Network |
| YOLO | You Only Look Once |

# CHAPTER 1: INTRODUCTION

## 1.1    Background of the Study

Advancements in Artificial Intelligence and Machine Learning are driving the world toward "Singularity". The pace might be slow, but it will get there. To make sure it is on the right track, the decision-making process of the Machine Learning model should be made as similar to a human (an external agent) as possible. That's where Reinforcement Learning comes into the picture. Reinforcement Learning is a subfield of Machine learning and literally has changed the way of approaching sequential decision-making problems (Wang, X., 2024).

Reinforcement Learning is that variety of machine learning wherein models learn to make decisions by interacting with an environment. In a way, it's based on the paradigm of carrots and sticks. It, in principle, eradicates the wrong decisions of a Machine Learning model, correcting it in the next epoch. That is what constitutes a training cycle. It's important in making intelligent systems that are able to adapt and learn from interaction with their environment. Compared to other techniques, the model of Reinforcement Learning learns directly from experiences; hence, it becomes prepared for facing difficult situations.

The methods used for learning by an agent in reinforcement learning are generally of three types: value-based, policy-based, and model-based. Indeed, there have been enormous efforts directed toward this area, but often, value-based and policy-based evaluations are done using external agents. Now, the quite natural idea that would come to one's mind is that humans as external agents can be used to train Machine Learning models to enhance their decision-making abilities.

Recent research has shown that multiple agents and integrating reinforcement learning with other optimization techniques may improve learning efficiency and lead to better decisions. One of the works studied the role of multi-agent optimization in reinforcement learning and concluded that it "significantly enhances the learning efficiency and decision-making capabilities of the agents, especially in complex environments." (Wang, X., 2024).

Another such novel approach is the combination of Reinforcement Learning with genetic algorithms (Rohoullah et al., 2023). In the domain of medical sciences, Reinforcement Learning was also investigated for its potential to enhance information extraction from pathology reports (Park, B., 2022).

Another critical aspect in reinforcement learning is efficiency, more so in visually rich environments, where sample efficiency may vastly accelerate real-world deployment. According to one study, sample efficiency is important in raising the speed of deployment of model-free reinforcement learning systems in visually rich environments (Yarats et al., 2021). Others insist on the need for efficiency in Reinforcement Learning. In this respect, efficiency in reinforcement learning is very important since it enables the minimization of the quantity of resources required for training while optimizing the performance of the learning agents (Blakeman, S., 2021).

These contributions build into a more powerful understanding of Reinforcement Learning and open up new directions for fitting human agents into the frameworks of Reinforcement Learning as independent decision-makers to further beef up decision-making processes. That means one can use multi-agent systems, genetic algorithms, and enhanced optimization techniques to come up with more efficient, effective, and adaptable models of Reinforcement Learning in complex scenarios.

This thesis will focus on how humans, as external agents, can become central in a machine learning model. Human intervention in the Augmented Reinforcement Learning framework would take place in all three stages: feedback, approval, and validation. Introducing humans to the Augmented Reinforcement Learning framework could enable optimizing and improving the accuracy of the Machine Learning model.

## 1.2    Research Questions

From the previous section, it's evident that many notable research works done in the field of Reinforcement Learning have used multi-agent model optimization (Wang, X., 2024), Genetic Algorithms (Rohoullah, R. and Joakim, M., 2023), Adaptive Offline value estimation (Park, B., 2022), and Deep Learning models (Renda, H.E.E., 2023). As mentioned earlier, interesting areas related to RL in humans and machines have also been studied (Blakeman, S., 2021).

Now, focusing specifically on utilizing external agents in RL, this thesis tries to answer the following research questions:

1.  What is the Augmented Reinforcement Learning Framework? What are its different stages?
2.  Machine Learning model faces a major problem of "Garbage-In, Garbage-Out". How does the Augmented Reinforcement Learning framework rectify this problem?
3.  What role do the external agents (humans) play in the proposed framework?
4.  How to implement the Augmented Reinforcement Learning framework in real-world problems?

### 1.3    Aims & Objectives

The main aim of this research is to propose an Augmented Reinforcement Learning framework that enhances the decision-making capabilities of a machine-learning model. Reinforcement Learning plays an important role in training a Machine Learning model. However, it is crucial to utilize humans as external agents and evaluators to reap better results from Reinforcement Learning. First, the definition of the framework will be provided. Then, to prove the hypothesis, this framework will be implemented for the problem statement "Document Identification and Information Extraction". This is a real-world problem, used in the banking sector.

The research objectives are formulated based on the aim of this study which are as follows:

1. To define the different stages of the Augmented Reinforcement Learning framework.
2. To understand how the proposed framework rectifies course correction of a model in the training cycle ("Garbage-In, Garbage-Out" problem).
3. To identify the role of external agents in the Augmented Reinforcement Learning framework.
4. To implement the framework in a real-life problem statement "Document Identification and Information Extraction", and measure the accuracy of the model (with and without the Augmented Reinforcement Learning framework) with dataset of 10000 test images.

## 1.4    Scope of the Study

The scope of this research work is restricted to the following points:

1. This research will be completed in 16 weeks after submitting the research proposal document.

2. This research will include a definition and explanation of the Augmented Reinforcement Learning framework; and implementation of the framework in one real-life problem statement, the "Document Identification and Information Extraction".

3. The model training and evaluation will be conducted using a personal desktop with 8 GB RAM, 4 GB shared GPU, and AMD Ryzen 3 5000 series processor.

4. Keeping time constraints, document privacy issues, and government regulations in mind, the data collection procedure will not be done, instead the document synthesizer tool will be used.

5. All the documents/images in consideration will belong to the Indian region. Although, the model can be trained on any document. However, the documents of the Indian region are easier to generate via the synthesizer tool.

6. Evaluation of the model trained for the problem statement will be part of the research. This is necessary to compare the efficiency of the new model.

7. To evaluate the framework, merely a single problem would not suffice. A comparison of models from multiple problem statements is required. This will take more than 16 weeks. Keeping time constraints in mind, evaluation of the framework will not be part of this research and can be considered for future work.

## 1.5    Significance of the Study

Reinforcement Learning is a very active area of research. It has high significance in the field of Artificial Intelligence and Machine Learning. The Augmented Reinforcement Learning framework will aid in solving a number of key challenges. It will open up new opportunities for research and practical applications wherein humans are placed as external agents whose expertise, intuition, and knowledge can be utilized to make decisions and improve the model's accuracy and reliability. By incorporating humans as external agents, it capitalizes on their experience and intuition to build more resilient decision-making processes, hence improving the accuracy of the Reinforcement Learning models (Huang et al., 2022).

Sometimes, these limitations of the training dataset transfer to the Machine Learning model, affecting decisions with biases and errors. These biases may originate decisions that are hard to accept by society, reducing the confidence of Artificial Intelligence systems (Garcia and Fernández, 2023). This makes building trust in Artificial Intelligence hard on the part of society. This will be helpful if external agents are in a position to monitor how a model is being trained or how a valid data set is being fed to the model. This will help in fairer outcome.

This current research will attempt to fill up the gaps between Artificial Intelligence and humans by creating collaboration and a conducive environment. This makes Artificial Intelligence technologies more understandable and acceptable to users if the transparency and interpretability of Artificial Intelligence systems are improved (Silver et al., 2024). The framework brings together the power of Artificial Intelligence and human experience in the enhancement of problem-solving capabilities, ensuring Artificial Intelligence systems are aligned to human values and ethical standards (Lin et al., 2024). This therefore contributes to the development of Ethical Artificial Intelligence.

This study addresses the problems of critical challenges occurring in document identification and extraction of information with the framework of Augmented Reinforcement Learning. Most machine learning approaches fail to process complex, domain-specific documents and lead to a huge failure in classification and extraction. When noisy or incomplete data feed is presented to the system, these problems amplify the results in a manner that is not dependable.

By taking in External Agents, the functionality of the ARL framework in the model is strengthened to adapt better for fuzzy situations. Using real-time feedback, External Agent 1 filters out a suboptimal prediction by identifying these suboptimal predictions, and the model looks into only passing forward quality inputs. In further improvement, curating both accurate and highly relevant scenarios for reinforcement would be ensured by External Agent 2. The dynamic feedback loop makes the model learn more efficiently, and it continues to evolve in terms of specific needs that may be involved within the document identification and information extraction tasks. Thus, the architecture not only provides an accuracy boost but also builds a system capable of handling diverse and unstructured data sources.

Such contributions are of utmost importance to businesses that intensively process documents, either legal, financial, or those in the healthcare industry, where the slightest mistake leads to a severe aftermath. The paper was also applied to demonstrate how augmentative reinforcement learning can be used to bridge a divide between automation and human expertise, creating more reliable and scalable solutions.

While particularly suitable for the document processing problem, the Augmented Reinforcement Learning Framework offers solutions that are scalable and adaptive to a wide range of complex problem statements that cross domains. The inclusion of human agents within the learning process allows the framework to be applied to situations in which the following challenges come into play:

1. High Data Complexity: For instance, noisy, incomplete, or ambiguous inputs may come with medical diagnosis; external agents allow for relevance and accuracy of the data.

2. Dynamic Environments: External agents provide timely interventions for real-time applications like traffic management or stock analysis in guiding the model through uncertain scenarios.

3. Bias and Ethical Concerns: This specific framework reduces the biases in models with help of inclusion of human expertise in order to ensure fair and ethical results.

4. Learning Efficiency: In a quick adaptation problem, like robotics or autonomous systems, the framework assists in learning acceleration through feedback and course corrections.

By surmounting these, the ARL framework presents an effective and generalized approach in enhancing the decision-making quality by machine learning systems, significantly contributing to a wide range of real-world applications.

Therefore, the research work proposed can have significant advantages to society and individual citizens. This will make Artificial Intelligence systems much more reliable, fair, and aligned with human values so as to augment public trust in Artificial Intelligence technologies. This in turn facilitates ease in integrating Artificial Intelligence into everyday life for improving the quality of life with more accurate and ethical decision-making systems.

## 1.6 Structure of the Study

This report follows following structure of the study:

1. Chapter 1: Introduction
   This chapter talks about the background, significance, and scope of the study.
2. Chapter 2: Literature Review
   This chapter talks about related works done in the field of "Reinforcement Learning", multi-agent models and incorporation of external agents (humans).
3. Chapter 3: Research Methodology
   This chapter talk about the detailed walkthrough of the methods used during experimentation.
4. Chapter 4: Implementation
   This chapter talk about the detailed walkthrough of the experiments done during experimentation. This includes all the steps from data gathering to model deployment.
5. Chapter 5: Results and Discussion
   This chapter talk about the details of the experimentation done during the research.
6. Chapter 6: Conclusion and Future Recommendation
   This chapter talk about the conclusions and scope of future research.

# CHAPTER 2: LITERATURE REVIEW

## 2.1    Introduction

Enhancement of decision-making in complex and dynamic environments is one of the critical challenges during the process of machine learning. Traditional machine learning models lack the capability of adaptation in complex scenarios because they were designed over static datasets and pre-defined rules. Now, reinforcement learning can become very promising in the resolution of this problem by letting the models learn from interactions with their environment and optimization of decision-making processes based on trial and error.

Decision making in Machine Learning models is important since it propels the efficiency and efficacy of automated systems cutting across various sectors. Poor decision-making can lead to sub-optimal outcomes alongside increased expenses and occasionally even poses safety risks. For example, bad decisions in health may imply the recovery of patients, while in finance, they could bring immense losses. Therefore, the improvement of decision-making in Machine Learning models is very critical to ensure the reliability and success of automated systems in real-world applications.

Inefficient decision-making in Machine Learning models has repercussions on multiple sectors. For example, in autonomous vehicles, bad decisions risk the safety of passengers and the efficiency of vehicle operation. In cases of industrial automation, it may cause production stoppages, leading to increased operational costs. Furthermore, regarding customer service, it reduces user satisfaction and loyalty. The societal impact is reduced faith in automated systems and, as a result, slow adoption of advanced technologies that may come with associated economic and safety risks.

## 2.2    Study of Existing Body of Knowledge

There are many notable research studies done in this area. The need for the Augmented Reinforcement Learning framework will be better understood after discussing existing work regarding Reinforcement Learning.

### 2.2.1    Human-in-the-Loop Reinforcement Learning (HRL)

This paper discusses Human-in-the-Loop Reinforcement Learning, a paradigm or framework that introduces human input to improve decision-making and the efficiency of learning into the process of reinforcement learning. The work here revisits the limitations of traditional methods of RL, particularly the inefficiency of exploration in an environment when the independent agent is supposed to learn from scratch by trial-and-error interactions. In fact, these inefficiencies are even worse in real-world applications with costly exploration and errors. Human-in-the-loop frameworks balance this by attempting to imbue the RL training process with human expertise, guidance, and corrections (Huang et al., 2022). The HRL framework is elaborately based on integrating human expertise into different stages of the RL process:

1. Human Feedback: Perhaps the most direct method there is to interact with the agent through feedback, where a human directly gives approval or disapproval for an action taken. In practice, such feedback serves to shape the reward function in such a way as to drive the agent to better behaviour.

2. Human Demonstrations: Sometimes it is possible to get demonstrations by humans, either in the form of optimal actions or trajectories, from which the agent learns using imitation learning or other forms of Supervised Learning. An agent may directly observe human actions and later skip a trial-and-error learning phase in order to directly refine its strategy.

3. Human Advice: Other interaction might be that humans provide advice or suggestions at decision points, which guides the exploration of the agent to make suboptimal choices.

4. Shared Control: Sometimes, the control is shared, wherein there is a need for humans to override or guide the agent's actions in real-time in high-stakes or safety-critical environments.

### 2.2.2  Multi-Agent Optimization

This research provides some of the key developments in RL by focusing on multi-agent optimization, opportunistic exploration, and causal interpretation. Several challenges in previous frameworks of RL are studied herein by developing mechanisms for optimization in an environment comprising multiple agents whose interactions could be competitive, cooperative, or even both, which would require the design of algorithms that improve both individual and collective performance based on (Wang, X., 2024).

It basically contributes to developing a multi-agent optimization framework that can enable agents to have better coordination through adaptive policy sharing and communication. This will definitely ensure improved collaboration amongst agents with better alignments of objectives to realize improved global outcomes. Furthermore, it applies game theory in modeling interactions through the use of concepts such as Nash equilibria that help in reaching optimal decisions, more so in competitive environments. This is especially helpful in a system that needs real-world applications, such as autonomous vehicles, where several agents are able to operate contemporarily.

Another aspect is the enhancement of RL exploration strategies. Traditional approaches include epsilon-greedy and SoftMax techniques, but these perform poorly in high complexity environments. The paper introduces the concept of opportunistic exploration in which the agents decide whether to explore new states at runtime by estimating environmental uncertainty and reward prospects. To this end, this technique will enable the agent to get more value out of his exploratory actions by taking advantage of favorable conditions with lower computational costs compared to random exploration while still gathering valuable knowledge.

Another important thing this research brings out is that there is, in RL, a great emphasis on causal interpretation. Whereas previous methods were normally based on the correlation between actions and rewards, this work embeds techniques for causal inference that enable an agent to learn from the underlying relationships. In this way, the agent enhances the decision-making ability and generalization across a wide variety of environments, one of the major limitations in RL.

This work makes comprehensive enhancements to the RL system for efficiency, interpretability, and applicability in complex multi-agent environments.

### 2.2.3 Reinforcement Learning and Genetic Algorithms

This research introduces a new framework for automating and optimizing image preprocessing steps which would greatly enhance the performance of Text Extraction, a critical step in the OCR chain. The generally poor quality of images and lack of one common and effective pre-processing methodology have great impacts on the performances of OCRs. In this regard, the authors have developed a hybrid system based on reinforcement learning and genetic algorithms that will adaptively achieve, for any given image, the best possible sequence of pre-processing actions (Rohoullah et al., 2023).

It targets two main challenges: variability in image quality and the lack of a general-purpose pre-processing method. Since different images may require different pre-processing methods, such as de-noising, contrast adjustment, or thresholding that may enhance the clarity of the text. Classic image pre-processing pipelines adopt a manual/heuristic approach, which is really inflexible and tends to make their applications insensitive to variations in image types. The authors try to come up with a way to overcome such limitations by integrating reinforcement learning (RL) and genetic algorithms (GA).

This approach enables reinforcement learning to learn the optimal pre-processing sequence in a trial-and-error style, where the RL agent receives feedback about the quality of the extracted text. The reward function makes use of metrics concerning the accuracy of the output from OCR and the readability of the text. However, most reinforcement learning systems suffer from either very lengthy training or the possibility of converging into suboptimal solutions. In this respect, the authors propose a genetic algorithm introducing diversified candidates of solutions and refining them by means of evolutionary techniques such as selection, crossover, and mutation. This enhances the capability of the system to explore a wider solution space and come up with more powerful pre-processing strategies.

The experimental verification of the hybrid method proves it superior to both the traditional approach and the approach based only on reinforcement learning. These experiments tested the system on all kinds of images, including noisy, with insufficient contrast, and skewed. The system showed higher accuracy of OCR and could dynamically adjust its strategy according to special features of each image. This adaptiveness would reduce human intervention and fine-tuning hence making the framework quite efficient for large-scale automated extraction of text in a diversity of domains.

### 2.2.4   Adaptive offline value estimation

This work investigates the application of machine learning to two different yet closely connected areas: information extraction from medical pathology reports and adaptive offline value estimation in reinforcement learning. The paper discusses both the practical utility of machine learning for healthcare and the theoretical challenges of reinforcement learning, especially in offline settings (Park, B., 2022).

Critical data about diagnoses, treatments, and patient outcomes resides within the pathology reports, often in a non-structured format and free-text form. Traditional approaches to data extraction techniques are not apt to deal with such diversity and complexity of language from these reports. An IE model will employ NLP techniques for automated identification of keywords, phrases, or entities, and other key data points from unstructured text. It states, "Domain-specific pretraining plays an important role in medical NLP tasks. General models will not work optimally in specialized areas such as pathology." Much of the research dwells on fine-tuning existing NLP models to understand and process medical terms better and achieve better accuracy in the extraction of clinically significant information.

The second part of this research covers reinforcement learning, specifically the value estimation problem in offline RL. In most natural environments, it is impossible practically to collect massive data by directly interacting with it, like in online RL, for reasons involving costs, time, or safety. For instance, offline RL would have instead ways models might learn from existing data sets; yet the solution brings its own problems, mainly concerning the correct value function estimation-a prime ingredient in decision-making processes. It proposes an adaptive approach for offline value estimation that outperforms the state-of-the-art RL models in an environment with only a few offline data. Its core is the possibility to make a dynamic adjustment in value estimation according to the available dataset's properties. When data is sparse or highly biased, for example, the model adapts by accounting for these discrepancies during training.

Therefore, one of the high values of the paper is methodology to bring down overestimation bias in value functions-a problem pretty common in offline RL. By incorporating uncertainty estimates and taking a more conservative approach to value estimation, the proposed model enhances stability and robustness for offline RL algorithms.

### 2.2.5 Multi-Agent Reinforcement Learning

This work is focused on the advances made and methods adopted in the domain of Multi-Agent Reinforcement Learning. Most works reviewed in the study focus on using MARL for complex training and optimization problems in a wide range of applications (Morcos, A., 2022).

This paper explores how MARL involves multiple agents interacting in a shared environment, each with different goals or strategies. The difference it holds from single-agent reinforcement learning is that it focuses on precisely how multiple agents learn to optimize their behaviour concerning other agents, rather than how one solitary agent optimizes its behaviour. This immediately introduces interaction like cooperation, competition, and negotiation, which has been essential for superior performance in multi-agent systems.

It contains a few fundamental algorithms and methods of MARL, such as value-based methods, policy gradient methods, and actor-critic methods. Each of them may be developed into different designs for handling multi-agent environment intricacies through learning from one's experiences and others' actions. This paper points out the advantages of centralized training with decentralized execution: a strategy where agents can be trained with a view of the global environment but make decisions independently during deployment. This balances the benefits of coordinated learning with the individual agent autonomy in an effective way.

It further talks about some of the optimization methods which are part and parcel of MARL. These include convergence, stability, and scalability for training multi-agent systems. The paper mainly focuses on optimization and shows how to handle challenges such as non-stationarity and partial observability through the use of MARL in situations with multiple agents.
In this paper, the authors have discussed in detail the practical applications of MARL in autonomous vehicles, robotics, and financial systems. They provide specific examples to illustrate how the MARL framework can be used to enhance coordination and decision-making skills in those areas for better performance and adaptability of the system concerned. For instance, MARL helps an autonomous vehicle learn from the interactions with other vehicles and prevailing traffic conditions to navigate routes of travel and manage the flow of traffic.

The research contributes to the state of knowledge in multi-agent systems by providing insight into how complex tasks of training and optimization in MARL can be solved with much more positive success in forward-thinking, intelligent multi-agent solutions.

### 2.2.6  Model-Free Reinforcement Learning

This work deals with one of the most salient challenges to reinforcement learning: the difficulty of learning policy directly from high-dimensional image observations. It tends to create a bottleneck in many real-world applications that robotics and video game simulations face, as usually agents learn policies from raw image inputs and not from low dimensional pre-processed state representations (Yarats et al., 2021).

The focus of the authors is on model-free RL, which, in the absence of an explicit model of the environment for planning or predicting future states, learns a policy directly from interactions with the environment. While model-free RL is arguably simpler, it often results in poor sample efficiency-particularly when dealing with high-dimensional inputs like images. This follows because high sample complexity arises due to the fact that the agent requires large amounts of data in order to generalize well from pixel-based observations.

This research proposes some techniques for improving sample efficiency for model-free RL. Among the key proposals, there is the introduction of data augmentation into the training process. Performing augmentations on images with random cropping and flipping allows the RL agent to observe a wider set of visual experiences, without needing extra interactions with the environment. This results in a much stronger policy learned and the agent will generalize better with fewer samples. The paper also discusses soft actor-critic and temporal difference learning to improve performance when learning from images. Such methods are combined with the proposed augmentation technique in order to show how even high-dimensional input can be trained more efficiently by RL agents.

Extensive experimental evaluations have been performed showing that the proposed methods significantly reduce sample inefficiency on a wide variety of RL benchmarks, ranging from, but not limited to, popular environments like the DeepMind Control Suite. These experiments illustrate that the proposed augmented model-free RL methods substantially outperform prior work in policy performance and speed of learning.

That is to say, this work constitutes a significant contribution toward improving the sample efficiency of model-free image-based reinforcement learning. Techniques proposed in the paper provide an effective approach toward solving the challenge of high sample complexity, which is indeed needed for making RL more viable, not only for image-based tasks but also for applications in the real world.

### 2.2.7 Reinforcement Learning in Humans and machines

This work deals with the similarities and differences in the implementation of reinforcement learning in biological systems (humans) and artificial systems (machines). The work examines how the two domains may inform each other to enhance the efficiency of learning processes in both (Blakeman, S., 2021). This paper introduces the very basics of reinforcement learning, which consists of an agent-a human or a machine that learns from the environment through rewards and punishments to adapt its behavior toward obtaining maximum long-term rewards. The paper points out that both humans and machines follow similar RL frameworks, but their mechanisms and efficiencies with respect to learning have huge differences because of the different ways in which information is processed.

A general theme in this paper is the efficiency of reinforcement learning in humans. Humans are really good at learning with sparse data and making decisions in complex, ambiguous environments. He thinks this is for a few reasons, including using prior knowledge, cognitive heuristics, and generalization across contexts. Humans do efficiently explore their environment-or choose what to focus on-based on priority of information, and use their accumulated experience by learning in such a way that knowledge gained can serve to aid in future decisions-so these are obvious ways in which humans achieve their efficiency with limited data.

On the other hand, machines-mostly in traditional RL machinery-require a great deal of information to put up similar performances. As this research has shown, model-free RL methods are data-intensive and computationally expensive. However, the paper points to recent work in artificial RL, including model-based RL, hierarchical RL, and transfer learning, which have started achieving efficiency by embedding human-like strategies in the process, like making use of prior knowledge or models of the environment.

The paper also suggests that insights from the process by which humans learn-including cognitive neuroscience and behavioral studies-might be used to design more efficient machine RL algorithms. Imitation learning and curiosity-driven exploration are pointed out as promising avenues toward the development of more human-like, efficient learning processes in machines.

By understanding the powers of both human and machine learning processes, the paper points to possible ways in which one might build efficient and effective learning systems for an artificial agent.

### 2.2.8  Agent-Based Reinforcement Learning

This research is based on reinforcement learning in multi-agent systems. The paper focuses on the complications that arise and the issues involved when multiple agents are put into an environment where they have to learn and take decisions collaboratively or competitively (Yang et al., 2021).

It starts by mentioning the very basics of reinforcement learning and extends to multi-agent systems. In MAS, agents interact with the environment but also interact directly with other agents-a fact that makes the learning process much more dynamic and complex compared to single-agent RL. Each agent's learning process is impacted by the actions of others, creating a non-stationary environment wherein the optimal policy for each agent may shift due to other agents changing their strategies.

One of the major challenges was the fact that the environment is non-stationary; this is because one agent's behaviour influences the environment for other agents. Traditional algorithms in reinforcement learning are not designed well enough to work under these changing conditions since by nature, environments are static. The authors suggest going ahead to propose adaptive learning algorithms able to adapt dynamically to changes in strategies followed by other agents in MAS.

The paper insists that centralized training coupled with decentralized execution is a better approach towards MAS. During centralized training, all agents are trained in the presence of the global view of the environment, which thus enables more coordinated learning. At the time of execution, however, the agents act locally based on local observations. This balances coordination at the time of learning and retains flexibility and scalability at the time of execution.

In particular, this work explores the problem of communication and coordination amongst agents. For many practical applications, such as autonomous vehicles, robots, and distributed sensor networks, there is a need for agents to cooperate to achieve shared goals. The paper reviews various approaches to enhancing the efficiency of communication and coordination, which includes message-passing protocols and shared reward structures. This paper demonstrates that the proposed agent-based RL methods significantly improve the performance of various multi-agent benchmarks, most on tasks requiring cooperation, coordination, and competition.

### 2.2.9   Multi-Agent Reinforcement Learning

This research presents the overall review of methods and applications of multi-agent reinforcement learning. The paper has revisited how MARL has evolved to handle problems of complex and multi-agent environments, highlighting its practical applications to real-world systems (Liu et al., 2022).

They start by giving an overview of the basic ideas behind MARL, which involves multiple agents interacting within a common environment to learn optimal behaviors. Unlike in single-agent reinforcement learning, the agents in MARL need to learn not only from their environment but also from what other agents do and how they strategize. Therein lies the interaction: cooperation and competition-leading to dynamically shifting and often unpredictable scenarios that call for sophisticated learning methods.

Among the central challenges that the paper identifies as characterizing MARL, non-stationarity of the environment is among them. While each agent updates its policy based on the actions of other agents, the environment is under constant evolution. It forms a non-smooth landscape on which traditional reinforcement learning algorithms cannot converge to an optimum. Accordingly, the authors proceed to a review of different algorithms specifically suited for centralized training with decentralized execution (CTDE), policy gradient methods, and actor-critic architectures. These have been modified to accommodate the dynamics inherent in multi-agent systems and further allow agents to coordinate their behaviors.

It also deals with critical exploration-exploitation trade-offs in MARL. While agents need to explore the environment sufficiently to learn optimal strategies, they must also exploit these strategies to maximize rewards. This trade-off is especially challenging in multi-agent systems because of the uncertainty introduced by other agents. Advanced techniques of exploration have been developed, discussed in this work, in an effort to raise efficiency and robustness when learning in multi-agent contexts.

Besides, the authors indicate that "applications of MARL span across autonomous driving, robotics, smart grids, and financial markets, where this methodology can be used to optimize decision-making and enhance resource allocation by improving the coordination among multiple agents.

## 2.2.10  Cooperative Multi-Agent Reinforcement Learning

This research underlines cooperation dynamics in multi-agent reinforcement learning. The paper looks at how agents could learn to cooperate in accomplishing collective goals in environments where cooperation becomes key to success (Wang et al., 2023).

The authors first give a background on basic ideas of multi-agent reinforcement learning, where multiple agents interact with each other and the environment in the process of optimizing decision-making. Agents that work together have a common objective in cooperative MARL; therefore, their success is based on their ability to coordinate action and share information. In this light, some challenges come into view: balancing individual and group rewards; making sure that there is good communication among the agents; and the difficulty of coordination in dynamic environments.

The basic challenge, as identified in the paper, is the problem of credit assignment in cooperative settings: When agents achieve a collective outcome, it becomes unclear how to assign credit for success or blame for failure to individual agents. This work reviews several methods that address this problem: team-based rewards, difference rewards, and value decomposition methods, which decompose the global reward into local components. These methods make the agents aware of their contribution to team performance and thus optimize their cooperation strategy.

The authors have also touched on the importance of the mechanism of communication in cooperative MARL. In most scenarios, agents must share their observations, intentions, or actions for them to make coordinated decisions. Several communication strategies reviewed in the paper include explicit message-passing protocols and implicit coordination methods that learn joint policies without explicit communication. These methods are very important in improving the effectiveness of cooperation, especially in environments where direct communication may incur some limitations or cost.

The most realistic applications are cooperative MARL for autonomous vehicle fleets, robotics, and distributed sensor networks whereby there is a necessity for agents to collaborate on tasks such as movement in coordination, resource allotment, or gathering information.

## 2.2.11 Autonomous Agent-Based Reinforcement Learning

This work investigates the application of Reinforcement Learning to optimize smart grid operation complexities resulting from renewable energy sources, distributed generation, and fluctuating demands. A smart grid is an advanced electrical grid that integrates digital communication and control technologies so as to make the delivery of electricity more efficient, reliable, and sustainable (Li et al., 2024). In addition, autonomous agents-software entities that independently decide on an action-further advance the potential for adaptation in a dynamic environment.

In this respect, reinforcement learning enables the agents to learn an optimal strategy through interaction with the environment without explicit programming for every situation that could potentially arise. Each agent symbolizes a part of the smart grid, such as a generator, storage unit, or consumer device. Using RL algorithms, these agents learn to make decisions optimally regarding certain objectives: minimum energy cost, reduction in peak demand, or real-time balancing of supply and demand.

One of the main benefits of agent-based reinforcement learning in smart grids is decentralization, where agents make autonomous decisions based on local information and objectives without interaction with any central command that would order actions within the whole grid. This kind of decentralization promotes scalability and robustness by enabling the agents to act quickly in response to changes in the local environment, without having to wait for centralized control directives. Besides, it reduces computational overheads on central systems and avoids single points of failure.

Recent work embeds deep reinforcement learning, which incorporates neural networks into RL for handling high-dimensional data and complex decision spaces. Some of the techniques involved in these methods include Deep Q-Networks and Actor-Critic methods that allow agents to learn more sophisticated policies. Safety considerations are added through constrained RL so that agents can operate within limits and follow regulations.

In a nutshell, autonomous agent-based reinforcement learning offers a powerful framework for enhancing the operation of smart grids. It allows distributed and adaptive decision-making to deal with a number of challenges imposed by the increasing complexity and variability of modern power systems. It comes to enhance electric distribution efficiency, resilience, and sustainability, showing ways toward more intelligent and responsive energy infrastructures.

## 2.2.12 Decentralized Multi-Agent Reinforcement Learning

The current work investigates decentralized methods of MARL that involve studies of how independent agents learn to make decisions in an environment with no centralized dictation. The work starts with the explanation of challenges of decentralized MARL: independently operating agents rely only on their local observation and reward (Sun et al., 2021). Unlike centralized MARL, where all agents can take in a global view of the environment during training, decentralized MARL has each agent learn and act from limited information. The absence of global coordination makes the problem challenging; agents face conditions such as non-stationarity and partial observability, along with communication limits that make learning of effective strategy hard.

In order to overcome these problems, this work introduces the following decentralized MARL algorithms, which are designed in ways to improve the independent agents' learning efficiency. Decentralized Q-learning algorithm: The agents independently learn action-value functions and hence coordinate their actions via common interactions with the environment. Perhaps the most important class of techniques consists of methods for policy gradient, which enable learning by agents in decentralized policies that maximize expected cumulative rewards on the basis of local information.

It also points out that, even without explicit communication amongst them, agent cooperation is vital in decentralized MARL. Implicitly, through reward structures and incentive mechanisms, agents can learn and cooperate on shared objectives, such as the optimization of resource allocation or task completion problems within cooperative environments. Moreover, value decomposition approaches like those reviewed in this paper allow agents to decompose a given global reward into local components so that better coordination can be achieved without the use of centralized control.

It talks about decentralized MARL applications in things such as autonomous vehicle coordination, distributed sensor networks, and swarm robotics. In each of these scenarios, decentralized MARL makes the attainment of an efficient decision possible in large, complex systems with independently acting agents.

Lastly, the paper elaborates on the decentralized methods of MARL by mentioning major challenges and studying its potentiality for use in those environmental scenarios where centralized control is impossible or too impractical.

### 2.2.13 Hierarchical Reinforcement Learning

This research documents the application of hierarchical reinforcement learning to multi-agent systems. HRL introduces a layered structure into decision-making, given the intrinsic complexity and scalability issues that exist in multi-agent environments (Liu et al., 2022).

This paper initiates the explanation that HRL splits the decision-making process into multiple levels: the superior policies that determine the long-term goals and the inferior policies that take control of the short-term actions or execution. Thus, this hierarchical structure helps agents decompose complex tasks into simpler sub-tasks, improving efficiency in learning and adaptability.

HRL provides a solution to multi-agent systems where coordination is needed in dynamic and cooperative environments. Each agent operates within its hierarchy of goals and actions, capable of individual learning as well as collaborating toward common objectives. The hierarchical approach, according to the authors, frees the agents from the high burden of processing all environmental details, hence making more scalable and efficient learning possible.

It also talks about the benefit of HRL in surmounting the problem of non-stationarity in MAS. Because agents in higher-level strategies can cope more with an ever-changing environment due to others' actions, for agents in HRL, they also can learn temporal abstraction, which means that agents are capable of planning over longer horizon timescales. This is so important in complex multi-agent interaction.

Experimental results show that HRL outperforms traditional flat reinforcement learning across a wide range of multi-agent benchmarks concerning cooperation, resource allocation, and long-term planning. It also shows a better scalability if the number of agents increases or the environment complexity heightens.

In a nutshell, it is emphasized that hierarchical reinforcement learning could substantially improve performance and scalability in multi-agent systems by offering a structured approach toward solving complex problems of multi-agent coordination.

## 2.2.14 RL for Networked Multi-Agent Systems

This research focuses on reinforcement learning in the context of networked multi-agent systems. Such systems involve many agents interacting in a network environment; hence, its structure and relationship between agents have much impact on learning and decision-making (Zhang et al., 2023).

The authors first point out that the main challenges of networked MAS are related to the necessity for independent agents to learn an optimal policy within their local environment, considering their neighbours' influences within the network. These interdependencies render learning more complex because the actions and policies of every agent may influence and, on the other hand, be influenced by any other agent across the network.

One of the major issues tackled in the paper is the non-stationarity within networked MAS. The agents make the environment highly dynamic and unpredictable due to continuous learning by updating policies. Hence, the authors have pursued a study into decentralized reinforcement learning approaches; agents learn independently with regard to the neighbours' influence, represented by shared information or partial observability.

It also covers graph-based reinforcement learning methods that represent agents and their interaction as a graph. The representation endows it with the structure of the network, and agents are able to learn more efficiently by leveraging information flow between connected nodes or agents. Utilization of GNNs is proposed for enhancement in learning since they provide the agents an opportunity to process information from their neighbours for improved coordination.

Sample applications of networked MAS are, but not limited to, wireless communication networks, smart grids, and autonomous vehicular networks. For the mentioned domains, reinforcement learning may solve optimization problems like resource allocation, traffic flow management, or energy distribution because agents can collaborate based on the network structure.

In general, this work provides a comprehensive framework of reinforcement learning in networked multi-agent systems. The paper resolves key difficulties brought about by network structures and interdependencies between agents, representing an auspicious solution that may lead to advanced coordination and learning in the most complex interactive environments.

### 2.2.15 Agent-Based RL for Financial Markets

This research attempts to find the applications of agent-based reinforcement learning in decision-making and optimization in financial markets. The study investigates how independent agents, which are trained by RL algorithms, may make their way through the intricacies of financial environments to an informed adaptive trading decision (Wang et al., 2024).

The main difficulties with financial markets: they are highly dynamic, unpredictable, and depend on a great number of factors, from market trends and macroeconomic variables to the behaviour of other traders. Traditional financial models often cannot handle this kind of complexity, which is why reinforcement learning turns out to be an attractive alternative-after all, one can learn directly from market interactions without the need to define rules or models.

In RL with independent agents, each agent corresponds to a trader or financial actor that learns through trial and error by interacting with the market. The agents are rewarded according to their performance-profit, portfolio value-and at every step update their strategies with the objective of maximizing long-term returns. According to the authors, RL agents can find patterns and adapt to changes in market conditions much better than traditional rule-based systems.

A key emphasis of this paper is multi-agent reinforcement learning, where multiple agents interact within the same financial environment. These can be competing agents, depending on their goals, or collaborating agents. In competitive scenarios, agents compete against each other in the best exploitation of market inefficiencies, while in collaborative scenarios, these agents might share information with a view to enhance market liquidity or stability.

This paper also focuses on the application of DRL techniques that combine neural networks with RL to handle the high-dimensional data intrinsic to financial markets. Techniques such as Deep Q-Networks and Proximal Policy Optimization enable agents to process difficult financial signals, such as price movements, technical indicators, and news sentiments.

Some of the applications discussed will range from algorithmic trading, portfolio management, and risk assessment to strategies on market making. Agent-based RL thereby offers considerable advantages in improving trading performance, risk management, and decision-making in conditions of uncertainty.

### 2.2.16 REINFORCE Algorithm

This research investigates the sample efficiency nature of the widely used policy gradient technique in reinforcement learning - REINFORCE (Zhang et al., 2021). The paper elaborates on the challenge of sample efficiency in policy gradient methods as applied to highly complex environments where data collection is either extremely expensive or time-consuming.

The REINFORCE algorithm updates policies with the gradient of expected reward relative to the policy parameters. While effective, it is a notoriously sample-inefficient algorithm that requires many interactions with the environment to converge to an optimal policy. This often causes problems in real-world applications, like robotics and finance, where the gathering of data might be expensive.

To handle the problem, the authors provided a number of enhancements that add improved sample efficiency to REINFORCE. Their main contribution includes embedding methods for reducing variance in the algorithm, including baseline functions that decrease the variance in policy gradient estimates. It subtracts the baseline-the expected return or value function-from the reward signal and then performs updates of weights with these new rewards, which will produce a more stable mean value, which in turn speeds up the learning.

Investigations also covered reward shaping and importance sampling for further efficiency. Reward shaping modifies the reward function to give informative feedback to the agent, therefore guiding the learning process. Importance sampling enables the algorithm to re-use data that was collected earlier, hence reducing reliance on additional environment interactions.

The authors conducted experiments in a number of different benchmark environments, showing that their proposed set of modifications yields substantial performance improvements for the REINFORCE algorithm. The improved method has better convergence and also shows good empirical performance, especially in tasks with sparse rewards and high-dimensional action spaces.

These modifications are practical extensions to the REINFORCE algorithm that is, in part, more sample-efficient and immediately relevant for real-world problems. Their work contributes to the broader goal of making reinforcement learning algorithms more efficient, robust, and scalable to complex tasks.

### 2.2.17 Multi-agent Reinforcement Learning

This The paper "Multiagent Reinforcement Learning: Methods, Trustworthiness, and Applications" by Wang, Liang, and Nie 2022 reviews the recent status of multi-agent reinforcement learning, focusing on methodologies for improvement in performance and trustworthiness of MARL algorithms in various applications (Wang et al., 2022).

The authors start by discussing some of the key methods in MARL, which involve many agents interacting within shared environments to achieve either individual or collective goals. Complexity arises under MARL due to the interdependencies between agents; each agent's actions influence the environment and, further, other agents' actions. Centralized and decentralized learning approaches are compared, showing the merits and shortcomings of each. Centralized methods can draw on global information but are less scalable, while decentralized approaches improve scalability but suffer from added difficulties because of restricted information sharing and coordination.

Two of the central focus of the paper are issues of trustworthiness in MARL, which become of crucial importance when deploying agents in real-world applications or safety-critical autonomous driving, robotics, or smart grids. Robustness, fairness, safety, and interpretability are the factors constituting trustworthiness in MARL. Another point the authors want to drive home is that while doing well, agents must also act reliably under different conditions, be they adversarial attacks or unforeseen changes in the environment. It discusses ways of addressing these approaches through adversarial training, safety constraints, and fairness-aware learning.

This paper also presents several applications of MARL, ranging from autonomous systems to network optimization and distributed control. For example, in autonomous systems, MARL will enable vehicles to cooperate in managing traffic flow or avoiding collisions. In network optimization, agents might perform dynamic resource allocation or load balancing across a distributed network, while in robotics, MARL is utilized for coordinated tasks such as exploration or material handling.

Methodologically, the authors highlight new state-of-the-art advances in MARL: actor-critic architectures, decomposition methods, and graph-based reinforcement learning. These methods reinforce agent skills toward tough tasks, efficient learning in high-dimensionality spaces, and better modeling multi-agent dynamics by using graph structures.

### 2.2.18 Safe Reinforcement Learning

This paper explores methods, challenges, and reinforcements currently under research with the aim of keeping RL agents operating safely within complex environments when let loose into the real world where safety is paramount (Garcia et al., 2023).

The authors start by defining safe reinforcement learning as a framework that has the objective of making sure that, in addition to the optimization of long-term rewards, RL agents must also adhere to safety constraints while learning and making decisions. This is quite critical in applications such as autonomous driving, health care, and robotics, since mistakes during the learning phase or execution could result in dangerous consequences. Traditional methods of RL consider the goal of maximizing reward at the cost of safety; the latter is particularly true during the exploration phase when agents try new actions that might have unsafe outcomes.

The two main approaches into which Garcia and Fernández group the SRL methods are modification of the exploration process and constrained optimization. The first aims at guiding agent exploration to avoid dangerous states while being able to learn effectively. Various techniques reviewed include risk-sensitive exploration, shielding-blocking unsafe actions-and safe imitation learning, where an agent learns from expert demonstrations to keep off unsafe behaviour. Constrained optimization techniques focus on ensuring that the policy $\pi$ of the agent always satisfies predefined safety constraints. These are then incorporated directly into the learning process, either by techniques such as CMDPs and Lagrangian methods, among others. In that way, an agent will optimize performance but stay within safe operating limits.

The challenges in balancing safety and performance are also covered in this survey. Very often, the application of safety limits the exploration, hence limiting an agent's ability to learn optimal policies. In a dynamic environment, such safety constraints need to be adaptive, adding more complication to the design of SRL algorithms. Recent improvements in deep learning, model-based methods, and uncertainty estimation are discussed, as these address the robustness and adaptability of the SRL techniques.

The paper finally points out the wide areas of application reached by SRL-from robotics and autonomous vehicles to finance and energy management-in which making sure agents are able to safely make decisions under uncertainty is crucial for practical deployment.

### 2.2.19 Reinforcement Learning Survey

This work first outlines the very basics of RL: agents learning from interaction with an environment where they will receive rewards for actions taken and determine their optimal set of actions that maximizes cumulative returns (Silver et al., 2024). The paper summarizes key algorithms that have seen extensive use within the paradigm of RL, including Q-learning, policy gradients, and actor-critic methods. It further goes on to discuss how to develop methods relatively stronger than others. Another important concept emphasized by the authors is the so-called exploration-exploitation trade-off: it prescribes how agents must balance trying new actions with the hope of finding better rewards and making use of known information to maximize rewards.

Most of the paper explains recent developments that have pushed RL into complex domains, such as gaming-some of the most recent systems include AlphaGo-robotics, and autonomous systems. The authors go on to discuss deep reinforcement learning, which integrates neural networks with RL algorithms so as to handle high-dimensional state and action spaces. They point to Deep Q-Networks, Proximal Policy Optimization, and Trust Region Policy Optimization as some techniques that have shown impressive performance on a wide variety of applications.

The paper also discusses model-based RL, wherein agents learn a model of the world and leverage it to plan and predict outcomes in the future; this turns out to be far more sample-efficient compared to model-free approaches. This line of investigation is of particular importance for domains where collecting data in the physical world is expensive or otherwise limited, including robotics and healthcare.

Another important domain of investigation is MARL, where multiple agents learn and interact in a shared environment. The authors indicate the non-stationarity problem and the necessity for cooperation or competition between agents; thus, important research on MARL application domains is being carried out in autonomous vehicle coordination and distributed control systems. Some of the emerging research directions that come out from the paper relate to increasing sample efficiency for RL algorithms, scaling RL systems to complex environments, and developing more interpretable RL models with the capability to explain decisions. Besides, transfer learning and meta-learning are realized as promising areas, enabling RL agents to generalize knowledge between different tasks and domains.

### 2.2.20 RL for Real-Time Decision-Making

This work investigates various optimization strategies that can serve to enhance the efficiency of reinforcement learning algorithms, particularly in real-time decision-making applications. The paper develops the challenge of using reinforcement learning in fast and reliable decision-making environments for critical essence, including industrial automation, robotics, and real-time control systems (Wang et al., 2023).

There are many the limitations of real-time decision making, where RL algorithms should be able to learn effective policies besides working under stringent time limits. Traditional methods of RL are suffering from sample inefficiency and computational demands, neither of which goes well with time-critical applications.

Enhancement of efficiency in this paper is done by propping up a few advances and techniques:

1. Sample Efficiency Improvement: The authors revisit techniques that enhance the sample efficiency, including experience replay and prioritized experience replay. These techniques allow agents to learn from previous experiences in a more effective manner. These techniques reduce the number of interactions required with the environment to achieve optimal performance.

2. Algorithmic Enhancements: The paper presents improved variants of the hitherto refined RL algorithms, including Improved Policy Gradient Methods and Enhanced Actor-Critic Approaches. These changes are to quicken the processes of learning and decision-making, making them applicable in the physical world.

3. Model-Based Approaches: Model-based RL methods are focused on the grounds that it can simulate and predict future states and, hence, the agents can make decisions with minimal interaction in the real world. The employment of learned models results in reduced computational overhead and faster decision-making.

4. Real-Time Adaptation: Methods are provided on real-time adaptation, including adaptive exploration strategies and dynamic policy adjustment. This approach would guarantee that agents adapt faster under changing conditions, sustain performance under real-time constraints.

### 2.2.21 Deep Reinforcement Learning for Robotics

This research presents an overall survey in the application of deep reinforcement learning in robotics published in IEEE Transactions on Robotics, 2024. It revolutionized the area of robot learning through DRL, enabling them to learn autonomously from complicated behaviours through trial and error-just interacting with their environments without prior reliance on preprogrammed rules or models (Lin et al., 2024).

They start by outlining the basic concepts of DRL-a method that merges deep learning with RL, hence enabling a robot to process high-dimensional sensory input such as images or sensor data while learning an optimal policy for several tasks. Its ability to process raw data and make decisions in real time has made DRL a very powerful tool in robotic control.

The main DRL algorithms that are widely used in robotics are described herein: Deep Q-Networks, Policy Gradient Methods, Actor-Critic Approaches. Such methods allow robots to learn continuous control tasks such as navigation, manipulation, and locomotion. The success of DRL has been pointed out by the authors in simulated environments where robots can safely train before going into a real environment.

The most salient challenge, by placing policies learned in simulation onto real-world environments, falls under the sim-to-real gap. Techniques are discussed that help bridge this gap, including domain randomization, which makes the learned policy invariant to changes in the simulated environment and thus enables transfer, as well as the approach of transfer learning, where knowledge gained from one task is used to speed up learning in another.

It addresses the weaknesses of DRL in robotics, including sample inefficiency, safety, and difficulty in scaling towards more challenging and dynamic tasks. In this respect, model-based RL, hierarchical RL, and hybrid learning with combinations of supervised learning and DRL are discussed.

In a nutshell, this research provides a thorough review of the applications and challenges of DRL in robotics, emphasizing transformational possibilities but also paying attention to where more research is needed for practical deployment.

### 2.2.22  Reinforcement Learning for Energy Management

This research RL techniques which are used in energy management systems. This survey is focused on how the application of RL will optimize the usage of energy, enhance its efficiency, and become adaptive to the dynamic energy systems of smart grids (Li et al., 2021).

The authors start by highlighting that modern energy systems contain a mixture of conventional grid systems, renewable energy sources, and ever-growing decentralization in energy production and consumption. All these factors together bring considerable complexity with them. Because of their inability to adapt to real-time changes in demand, supply, and grid conditions, traditional techniques in optimization often cannot handle these complexities. RL allows the system to learn by itself in an autonomous manner through their interactions with the environment and offers an improvement in energy management over a period of time.

The most substantial value of this paper is the location of applied RL in different segments, such as:

1. Demand Response: RL optimizes the pattern of energy consumptions by shifting usage according to real-time electricity prices or grid load, thus sustaining efficiency and cost for the consumer while also stabilizing the grid.

2. Energy-Renewable Integration: RL helps in monitoring the variability and intermittency of renewable energy sources like wind and solar power by finding an optimal strategy in storing and distributing energy.

3. Microgrid and Distributed Energy Resources Management: RL is employed for operation optimization in microgrids and distributed energy resources for reliability and efficiency in power supply and energy usage.

The survey also touches on some of the challenges related to the application of RL in energy management, such as sample inefficiency-where large amounts of data become necessary for training RL models-and safety-related issues, where an RL algorithm must guarantee stability and reliability in critical energy systems. The authors also discuss possible solutions, from model-based RL to hybrid approaches that will couple RL with more traditional optimization techniques.

### 2.2.23 Augmented Reinforcement Learning for Adaptive Systems

This research proposes the ARL, an extension of the traditional RL, which encompasses more external knowledge and auxiliary mechanisms that enhance both the performance and robustness of the agents in RL (Li et al., 2021).

The authors start by discussing the limitations of conventional RL, especially when speaking of adaptive systems where the environment may continuously change and decisions have to be made in real time. Traditional methods of reinforcement learning are really well-known to be in need of an enormously huge number of explorations, plus large amounts of data to learn the best policy. This could be highly inefficient and even unrealistic in nature for real-world applications. ARL incorporates external knowledge sources, like human expertise, expert demonstrations, or pre-learned models, to facilitate the agent in learning and decision-making.

Some key techniques explored for understanding:

1. Human-in-the-Loop Learning: ARL embeds human feedback or demonstrations to guide the agent in exploration and hence reduces the learning time, apart from ensuring unsafe actions are minimized. This especially finds applications in domains related to robotics or health care where safety is a concern.

2. Transfer Learning: Transferring knowledge across tasks or environments is facilitated in ARL, where an agent can swiftly adapt to newer situations instead of starting from scratch.

3. Auxiliary Tasks: ARL embeds auxiliary learning tasks into the main objective of learning to enable the agents to learn effectively by making use of other data and signals related to the particular task.

Experimental results on various adaptive systems show that the resulting ARL outperforms the performance and speed in learning and adaptability of the agents compared to the traditional methods of RL.

In a nutshell, this work presents a holistic review of ARL by demonstrating its potentials in elevating adaptive systems through the inclusions of external knowledge and supplementary learning strategy elements. Indeed, this makes ARL unusually promising toward applications to real-world problems with demands for speed and robust adaptabilities.

### 2.2.24 Interactive RL with Augmented Reality

This research discusses how AR can serve as a visualization and interaction tool in order to facilitate real-time human intervention and guidance during the process of reinforcement learning (Lin et al., 2022). First, the authors give an overview of the general concept of interactive reinforcement learning. That includes the fact that IRL integrates human feedback into the learning process in order to make the decisions of the RL agent much better. While in traditional reinforcement learning agents learn through trial and error, in IRL, a human may provide corrections, suggestions, or rewards to the agent directly while it explores. This can reduce the learning time significantly and also improve the quality of the policies learned by the agent.

The key novelty in this paper is the integration between IRL and AR: an interactive platform in which humans see the agent's actions and its environment in real time using AR interfaces. It improves human comprehension of the decision-making process by the agent and provides better feedback. This AR interface further enables the simulation of various scenarios, making it easier to exercise and explore a variety of approaches with the RL agent in safety and in control.

Key contributions of the paper are as follows:

1. Real-Time Human Feedback: The AR mechanism utilized for real-time visualization of agent state and decision-making processes allows humans to intervene or provide feedback more effectively.

2. Improved Learning Efficiency: AR allows for the capability of leveraging human expertise into an agent's learning process in an efficient manner. Fewer environmental interactions are needed to achieve an optimal policy.

3. Experiments and Applications: Experimental results are presented, when using an enhanced framework of IRL-AR in applications that range from robotics to virtual environments, showing an improved performance by displaying faster learning.

In conclusion, this research has shown how combining AR with interactive reinforcement learning provides a powerful tool for enhancing human-agent interaction, improving RL performance, and making RL systems more accessible and efficient in applications.

### 2.2.25 Augmented State Representations in RL

This research discusses overview of augmented state representations in reinforcement learning to enhance the efficiency and generalization capability of RL agents. Methods to improve state representation are introduced in this paper by embedding more information or features into the state, enabling agents to make better decisions and learn more in such complex environments (Xu et al., 2023).

The authors begin their discussion with the limits of traditional state representation in RL, where agents use information provided only by the environment. In practice, this raw state usually lacks completeness or is ambiguous; hence, the learning of optimal policies will be difficult, especially under the high dimensionality of data or the existence of latent variables. Correspondingly, this work focuses on the augmented state representation that extends the context of the state space to more context or features derived.

The paper focuses on several key techniques that enhance the representation of states, including:

1. Feature Engineering: It is documented in the paper that one may want to incorporate manually engineered features, which include statistical summaries or domain-specific knowledge, into the state representation to make it more informative to the agent and hence enhance its interpretability of the environment.

2. Latent State Representations: The authors propose deep learning techniques for learning latent state representations that encode the intrinsic structure of the environment. These representations reduce the dimensionality of the state space while retaining much information relevant to RL tasks and thus enable more efficient learning.

3. Historical State Augmentation: This paper also presented an account of historical states or of sequences of past observations while showing the current state to help an agent understand temporal dependencies more ably for making better decisions.

Experimental results show that augmented state representation significantly improves learning speed, stability, and generalization over a variety of RL tasks, particularly in complex dynamic or partially observable environments. This work concludes with a deep dive into the enhanced reinforcement learning via augmented state representations, along with practical means for improving efficiency and robustness in real-world RL agents.

### 2.2.26  Integrating Expert Knowledge in RL

This research explores ways to combine expert knowledge with RL in an attempt to enhance learning efficiency and performance in such complex decision-making tasks. This work covers some methodologies for incorporating domain expertise into RL systems as part of the means for overcoming some challenges that appear when RL agents must operate under conditions of either sparse rewards or limited data (Zhao et al., 2024).

The authors have identified that traditional RL is based on trial-and-error methods in nature, which usually provide slow convergence, and in some cases-especially for complicated environments-it just cannot work. The expert knowledge, on the other hand, informs the learning process of the agent with valuable insights, thus therefore hastening the convergence and enhancing the quality of the learned policies.

Key approaches presented in the paper include:

1. Reward Shaping: This involves how expert knowledge is put into making modifications to the reward function so as to guide the agent more efficiently toward desired behaviours. This means incorporating expert-designed rewards leads the agent to get extra feedback, which in turn results in learning of better policies in fewer iterations.

2. Imitation Learning: The authors introduce how expert demonstrations can be used in the pre-training of the agent. This allows the agent to imitate optimal behaviours before doing RL. This, in effect, gives the agent a strong initial policy without much extensive exploration.

3. Constraint-Based Learning: The expert knowledge is used to specify constraints on the actions taken by the agent so that it doesn't perform any unsafe or undesirable action. It is very useful in domains like robotics and health care where safety is at stake.

4. Hierarchical Reinforcement Learning: Here, expert knowledge will implicitly go into designing the hierarchical models of RL, where high-level expert-specified tasks or goals guide the low-level decision-making processes of the agent. This provides the agent with much better navigation in complex environments.

### 2.2.27 Data-Efficient Deep Reinforcement Learning

This research explores the critical challenge of data efficiency in deep reinforcement learning (DRL), especially in the context of real-world applications. DRL, while powerful, often requires massive amounts of data and interactions with the environment to learn optimal policies, which can be impractical or costly in real-world scenarios (Wu et al., 2021).

The authors first highlight the limitations of traditional DRL algorithms, which tend to be sample-inefficient, meaning they require extensive trial and error, sometimes involving millions of interactions, before converging to an effective policy. This makes DRL challenging to apply in real-world settings where data is scarce, expensive, or slow to collect, such as in robotics, healthcare, and autonomous systems.

To address these challenges, the paper proposes several approaches aimed at improving the data efficiency of DRL:

1. Model-Based Reinforcement Learning: The authors emphasize the use of model-based RL methods, where the agent builds a model of the environment to simulate future interactions. By leveraging this model, the agent can explore and learn more efficiently, reducing the number of real-world interactions required.

2. Experience Replay and Prioritization: Techniques such as experience replay and prioritized experience replay are discussed as methods to reuse past experiences more effectively. By prioritizing important transitions, the agent can learn from a more informative subset of data, enhancing efficiency.

3. Transfer Learning and Pre-training: The paper also explores transfer learning, where the agent applies knowledge gained from one task to accelerate learning in another. This reduces the need for task-specific data and makes the learning process more adaptable to new environments.

4. Exploration-Exploitation Balance: Advanced exploration strategies, including curiosity-driven exploration and intrinsic motivation, are highlighted as methods for guiding the agent to focus on more informative parts of the environment, reducing redundant exploration.

### 2.2.28 Reinforcement Learning for Autonomous Driving

This research explores reinforcement learning for autonomous driving. The review covers recent advances, challenges, and future directions in the use of RL to ensure that vehicles can autonomously actuate their navigation and decision-making while interacting with dynamic environments (Zhang et al., 2022).

The paper initiates with a review of the generic basics of RL and its application to autonomous driving: an agent-the autonomous vehicle-learns to perform actions with the goal of optimizing long-term rewards, such as safe driving, fuel efficiency, or passenger comfort. Reinforcement learning is ideal in handling the complexity and uncertainty of real-world driving environments through its trial-and-error learning approach, since normally pre-programmed rules may prove insufficient.

The authors classify the applications of RL in autonomous driving into the following main categories:

1. Decision Making: The authors use RL for high-level decisions like overtaking, changing lanes, or stopping at intersections. It is efficiently learning from dynamic interactions; hence, RL can manage complex situations that arise on roads.

2. Trajectory Planning: RL helps in trajectory planning, taking into consideration the road conditions, other moving vehicles, and all possible hazards. It allows an autonomous vehicle to learn safe and efficient paths through both structured environments, such as highways, and unstructured ones, like city streets.

3. Control: Besides that, RL algorithms will be helpful for low-level vehicle control-throttle, braking, and steering-with the principal task of optimizing these for smooth driving and safety.

The authors give a short review of the challenges to be faced by RL in the autonomous driving domain, such as the need for large amounts of training data and the risk associated with exploring unsafe behaviours in the real world. Remedies can be simulations, model-based RL, and safe RL. The survey also highlights the importance of multi-agent RL, where AVs have to learn not only their own environment but how to interact with other vehicles and road users. This introduces further complications, but it is crucial for real-world deployment.

**2.2.29  Deep Learning Models for Information Extraction**

This research explores the application of deep learning techniques in order to extract information from identification documents such as passports, ID cards, and driver's licenses in an automated way. The research attempts to address issues related to the extraction of structured data from semi-structured or unstructured formats of such documents, widely in use for sectors like government, financial, and security sectors (Renda et al., 2023).

The research starts off with an overview of the prime importance of correct and efficient information extraction with regards to processes of identity verification and other data digitization. The conventional methods for document information extraction involve hand typing the information or rule-based approaches, which prove to be inefficient and error-prone. Herein, the author discusses deep learning models, comprising CNNs and RNNs, for the extraction of relevant information such as names, birth dates, and document numbers from image format identification documents.

Key components of the study:

1. OCR: The following work applies deep learning-based OCR systems that recognize and extract the textual information within the document images. Text recognition processes are enhanced through a series of image pre-processing operations, such as enhancement and noise reduction, which improve character recognition accuracy

2. Recognition and Classification: It also includes entity recognition and classification, whereby the deep learning model categorizes and labels the extracted text into predefined categories, such as name or ID number. This involves training the model on annotated datasets of identification documents, so that it may learn the pattern and structure of the text.

3. Performance Evaluation: The paper critically analyses the performance of the implemented models by comparing their accuracy and efficiency to traditional methods of extraction. The author highlights that there is an improvement in the accuracy and speed, as well as challenges in the handling of variations in document format, font styles, and languages.

### 2.2.30  PICK Framework

This research explores a new method for performing key information extraction from documents by improving graph learning-convolutional networks. This method, by name, PICK-Processing Information from Complex Keypoints, solves some challenges in document understanding and key information extraction, especially those with semi- or unstructured document layouts (Yu et al., 2021).

The paper initiates by emphasizing the difficulty of extracting relevant information from documents of rich variability in their layout, such as invoices, receipts, and forms. The traditional methods involving OCR and rule-based extraction are bound to fall short due to the complexity in the nature of such documents. In such a scenario, the authors propose the application of graph learning with CNNs to model both textual and visual elements of a document.

The key contributions of the paper include:

1. Graph Modeling of Documents: The authors model documents as graphs where the nodes are the text entities; for instance, words or phrases, and edges represent their spatial relationships. This graph representation of the document enables the geometry of the document while the model learns appropriately about the interaction among various text components.

2. Graph Learning-Convolutional Networks: The graph learning-convolutional network used in the PICK framework couples strengths from CNNs regarding visual feature extraction with strengths from graph neural networks that capture relationships among textual entities. This allows a hybrid approach in which both textual content and layout information can be provided to the model concurrently to attain more accurate information extraction.

3. Improved Key Information Extraction: PICK's advanced use of spatial and semantic features in a cooperative manner allows for the precise extraction of key information against complex document layouts. It is evaluated on real-world datasets comprising invoices and receipts, showing significant improvements over traditional methods in both precision and recall.

### 2.2.31  Real-time Document Identification and Verification

This research addresses the ever-growing need for accurate, efficient, and scalable solutions in the verification processes of documents across digital platforms. Traditional methods of verification, being manual in nature, are prone to errors, delayed, and inefficient and hence cannot be deployed in high-volume environments in banking, e-governance, and ecommerce. This paper proposes a machine learning-based methodology that can detect and verify documents in real-time with minimum human interference (Kumar et al., 2022).

The proposed framework combines a series of optical character recognitions for text extraction, deep learning algorithms for image analysis, and natural language processing for contextual understanding of the documents. The system is designed to handle a large set of documents with identity proofs like passports and driver's licenses, financial documents like bank statements and tax forms, and certificates. The authors further mention how well their method adapts the differences in language, format, and layout, especially in applications across the globe.

It is trained on large datasets of labeled documents, hence learning specific features and patterns that distinguish valid from fraudulent documents. This real-time verification will involve both image and textual content analysis, where the extracted information will be cross-verified with predefined databases or patterns for authenticity.

Performance-wise, the paper reports high accuracy both at the identification and verification tasks, which reduce manual intervention to a minimum. The authors further point out that the ability of the system to scale up for large organizations, processing a couple of thousand documents daily, achieves dramatic gains in speed and reliability compared to traditional systems.

The discussion involves future enhancements such as the integration of blockchain to make the system even more secure, and tuning machine learning models for a reduced occurrence of false positives and negatives. The proposed system represents yet another step toward automated document verification, marrying speed with precision in real time.

### 2.2.32 Document Identification in Legal and Financial Services

This research presents a machine learning-based system to improve document identification in legal and financial sectors. These sectors often include complex and sensitive documents, where high accuracy, confidentiality, and observance of rules are expected. Traditional processing along this line is indeed so dependent on human labour involvement, with considerable vulnerability to mistakes, inefficiency, and even security risks. The authors have proposed an automated system designed to speed up the document identification process in order to make real-world applications faster and more reliable (Wang et al., 2024).

The core of the proposed system combines deep learning approaches with NLP and OCR for text- and image-based document analysis. That is, the system accommodates documents like contracts, agreements, financial statements, and legal records, all of which are generally unstructured and have quite varied forms. It is at this point that the model plays an important role in identifying and extracting information from these documents, which may not have standard layout or wording.

They have pointed out that, in most of these critical sectors, ensuring data security and privacy is very challenging. The authors hence propose the use of encryption techniques alongside machine learning models in the protection of sensitive information throughout the document processing pipeline. The system also provides features to comply with industrial sector-specific regulations such as GDPR and the Sarbanes-Oxley Act, which are quite relevant in legal and financial services.

Performance evaluation was conducted on several datasets of different legal and financial documents. The results showed a very high accuracy rate in document classification, extraction, and identification, far outperforming the best traditional rule-based systems and those done manually. It showed great reductions in processing time, thereby proving to be feasible even in bulk volume environments such as law firms, banks, and financial institutions.

Conclusions are drawn on future directions, including integration with blockchain to allow secure auditing of documents and enhancements to the AI models in handling more complex documents. The authors declare that their approach marks a sea change in the automation of document handling where high accuracy, security, and compliance are inalienable.

### 2.2.33 Information Extraction using Neural Networks

This research provides a critical review of various state-of-the-art image text information extraction techniques-a critical task related to document digitization, visual content analysis, and multimedia processing. Conventional methods for extracting text from images, such as Optical Character Recognition (OCR), have usually faced challenges with low-quality images, complex backgrounds, and diverse text orientations. In this regard, the authors propose a novel neural network-based framework that will enhance accuracy and robustness in text extraction performance in such an environment (Zhang et al., 2022).

This paper presents an optimized version of a deep convolutional neural network architecture aimed at finding and segmenting regions of text within images. This model uses multi-scale feature extraction to handle different sizes of text and involves the use of RNN to capture sequential dependencies within the text, leading to better interpretation of lines of texts. The system is also enhanced with a text detection module to localize ROIs prior to processing the text, therefore reducing much noise brought about by non-text elements in an image.

Accordingly, one of the essential contributions of this paper is that the authors present an end-to-end trainable pipeline, which integrates text detection and text recognition, avoiding different preprocessing steps. Large datasets of labeled images are used in order to train the model, which can generalize across fonts, languages, and conditions pertaining to images, including noisy and low-resolution inputs.

The paper finds significant gains in precision and recall for the text extraction task compared to the current state-of-the-art OCR methods. This also tends to be quite efficient with regard to processing time, so it would work for real-time applications like video analysis and mobile-based text recognition. The proposed system is applicable in highly diverse domains, starting from document analysis to the detection of street signs in autonomous vehicles.

The future improvements will involve the inclusion of a transformer-based architecture to achieve even more accurate text recognition, with further enhancement of the model for handling highly distorted text. Proposed methodology is a major leap in neural network-based text extraction and does hold promising prospects for both the domains of multimedia and computer vision.

### 2.2.34  Information Extraction from Document Images

This research the novel method to automate structured information extraction from document images by using an end-to-end deep learning framework. Since digitization is on the rise among all industries, which include but are not limited to finance, healthcare, and government services, the need for effective, accurate, and scalable solutions of information extraction also increases (Wu et al., 2024). Traditional methods, based on OCR followed by rule-based information processing, usually fail due to complex layouts, varied document formats, and noisy inputs.

The authors argue that in order to get over these limitations, a unified model is needed, which directly makes use of document images by removing separate steps for either OCR or preprocessing. The authors wrap up such a framework by integrating CNNs for detecting and segmenting text regions with RNNs and attention mechanisms that understand the structure and context of the text. Based on this combined approach, both text recognition and information extraction can be done using a single pass with considerably better efficiency.

One of the key contributions in this paper is training a model from a self-supervised learning-based approach. The authors create a large synthetic dataset comprising many document types like invoices, forms, and contracts. The diverse synthetic data teaches the model to generalize across the different layouts and text orientation, enabling it to perform well on real world, unseen document types.

It has been tested on document analysis and information extraction benchmark datasets. The results clearly showed that the proposed method outperformed the existing systems in terms of higher accuracy, especially for documents with complex layouts, mixed fonts, and noisy backgrounds. It also showed marked improvement in processing speed, turning it suitable for real-time applications in various industries involving large volumes of document processing.

It points to a number of future research directions, such as incorporating transformer-based models that would further enhance the system's ability in processing document images with their diversity and natural language understanding. This is indeed an end-to-end approach and considered a big step toward document image processing because, using this method, all tasks related to information extraction will be more efficient and scalable.

### 2.2.35 ML Techniques to Verify Identity Document

This research explores the challenges of identification document verification in an unconstrained environment, where some factors, like light, background noise, and document orientation, have adverse impacts on the accuracy of verification. Such conditions usually occur in verification processes done while mobile or from a distance. The authors present a case study where machine learning methods were applied to enhance reliability and efficiency in identity verification systems for e-banking or airport security.

The proposed approach embodies several machine learning methods needed for handling the variability of uncontrolled conditions. To take out and then verify the information on the documents, the system uses deep learning models in image processing and OCR. Convolutional Neural Networks are useful in feature detection, including but not limited to document layout, logos, and security features, from images. These are crossed against a database of known valid documents.

The authors address the challenges caused by inconsistent lighting, occlusions, and skewed orientations by using data augmentation methods and training the models on diverse datasets, including documents captured in uncontrolled environments. This helps the system generalize well across different conditions. The paper also goes on to describe the ensemble learning methods adopted, which involved many machine learning models combined for the purpose of improving robustness in verification. This approach minimizes the rate of false positives or wrong validation of a fraudulent document and false negatives, or incorrect rejection of valid documents, thus enabling an overall more reliable system.

The system was evaluated using practical identity documents of various countries under several environmental conditions. The results state that the proposed system significantly performs better compared to typical rule-based systems, especially under poor, uncontrolled scenarios. The proposed system has successfully detected documents and extracted information with high accuracies, finding its way into industries where remote or real-time verification of identities is needed.

This paper emphasizes the importance of using sophisticated machine learning to overcome problems arising from uncontrolled environments and provides suggestions on how it could be improved in the future, such as incorporating biometric verification to enhance the security of the system.

## 2.3    Critical Analysis of Reviewed Literature

In reviewing the literature on reinforcement learning (RL) and related frameworks, significant contributions address complex challenges in this field; but the gap is still there regarding integrating human expertise (or external agent) directly into RL for augmenting decision-making in machine learning, which is the core focus of this thesis. From earlier, a research paper (Huang et al. 2022) proposes a human-in-the-loop framework in the scenario of reinforcement learning to boost a model's ability to make decisions; thus, this establishes a useful starting point for the use of external agents. However, their work focuses more on passive human feedback than on being on the active, real-time collaboration of external agents to enhance decision pathways dynamically, an area that your thesis intends to look at.

Likewise, another paper (Wang, 2024) moved the contributions of multi-agent RL by focusing on optimization and causality in agent interactions, a critical step in understanding agent-based decision-making. Nevertheless, it leaves out any impact toward model selection by external, non-machine agents, such as human or hybrid agents that may be introduced—a gap your thesis looks to fill by examining how human agents can interact dynamically within the RL framework.

A study (Rohoullah et al., 2023) demonstrates impressive utilization of RL for optimized image preprocessing that bears similarity to your subject area of usage of RL in optimization of decisions. However, this paper dealt more with image processing instead of using human agents that would complement the decision-making process and thus underscores the necessity for an adaptive framework to various domains - something your research suggests.

Moreover, another paper (Park, 2022) on Adaptive Offline RL showcases a stable decision-making environment without needing constant updates in training. While the work does increase efficiency in RL, it misses out on real-time human feedback in fine-tuning the decision-making aspect in domains where subsequent exposure to outside light would be handy. This thesis fills this gap by introducing live human interaction in RL systems. A study (Yarats et al., 2021) enhance sample efficiency from images in model-free RL, thus enhancing the scalability of RL models.

However, this only looks to refine algorithmic efficiency without resorting to human driven insight that can shorten long training periods in models. Instead, your thesis seems to cut across this by proposing the use of human agents to reduce sample requirements and hasten model convergence. The study (Morcos, 2022) on multi-agent RL on the interaction of agents is worthwhile but presents only a cooperative or competitive scenario, which does not focus on the human agency within this ecosystem, meaning that any improvement in decision outcomes would primarily be determined in dynamic environments. Injecting a human oversight or intervention would enhance flexibility and adaptability in RL decision-making, according to this thesis. The work (Wang et al., 2024) on RL in the financial markets talks about data-driven learning in an autonomous agent but does not bring out elements of human intervention that may help enhance the robustness of a decision in a volatile market.

From the literature reviewed, agent-based optimization, sample efficiency, and multi-agent interactions are seen to make progress, but lack those frameworks that bring on board real-time human input as an augmentative force in RL decision-making. The intention of this dissertation is to bridge this gap by introducing an Augmented RL Framework that includes human agents as external decision-makers, possibly helping to enhance model adaptability and ethical alignment in critical decision contexts. It will help fill the gap within the current literature of enhancing the robustness of decision-making capabilities and the interpretability of machine learning models through human oversight.

## 2.3    Comparison of State-of-the-Art Techniques and Datasets

### 2.3.1    State-of-the-Art Techniques

Following Table 1.1 shows a comparison of different State-of-the-Art techniques use by different authors.

Table 1.1: State-of-the-Art Techniques

| Author(s) | Summary of Techniques Used |
|---|---|
| Huang, Y., Li, J., Tian, J. (2022) | Human-in-the-loop reinforcement learning, combining human feedback with machine learning to improve decision-making processes. |
| Wang, X. (2024) | Multi-agent optimization, opportunistic exploration, and causal interpretation for advancing reinforcement learning frameworks. |
| Rohoullah, R., Joakim, M. (2023) | Automated image pre-processing using reinforcement learning and genetic algorithms for optimized text extraction. |
| Park, B. (2022) | Information extraction from pathology reports using machine learning, adaptive offline value estimation in reinforcement learning. |
| Morcos, A. (2022) | Multi-agent reinforcement learning for training and optimizing decision-making in complex environments. |
| Yarats, D. et al. (2021) | Sample-efficient model-free reinforcement learning from images, enhancing RL performance through data-efficient strategies. |
| Blakeman, S. (2021) | Efficient reinforcement learning in humans and machines, studying mechanisms of learning and optimization strategies. |
| Yang, Y., Luo, R., Li, M. (2021) | Agent-based reinforcement learning for multi-agent systems, enabling cooperative decision-making among agents in complex environments. |
| Liu, J., Zhang, H., Li, S. (2022) | Multi-agent reinforcement learning techniques applied to various industries for collaborative and autonomous decision-making. |
| Wang, T., Zhang, Y., Sun, W. (2023) | Cooperative multi-agent reinforcement learning, focusing on interaction strategies and agent cooperation. |
| Li, X., Chen, H., Wang, J. (2024) | Autonomous reinforcement learning agents for smart grids, optimizing energy distribution and management. |
| Sun, H., Yu, Q., Zhang, L. (2021) | Decentralized multi-agent reinforcement learning techniques, enabling agents to learn and act independently. |
| Liu, M., Wang, R., Chen, Y. (2022) | Hierarchical reinforcement learning in multi-agent systems, allowing structured decision-making via distributing task. |
| Zhang, P., Xu, G., Li, W. (2023) | Networked multi-agent systems using reinforcement learning for collaborative control in communication networks. |
| Wang, Q., Zhang, H., Liu, J. (2024) | Agent-based reinforcement learning in financial markets for improved decision-making and market prediction. |
| Zhang, J. et al. (2021) | Sample-efficient reinforcement learning using REINFORCE algorithm for faster and more effective learning. |
| Wang, X., Liang, Y., Nie, L. (2022) | Multi-agent reinforcement learning focusing on trustworthiness and diverse applications in real-world systems. |
| Garcia, J., Fernández, F. (2023) | Safe reinforcement learning techniques for risk-aware decision-making in dynamic environments. |
| Silver, D., Sutton, R., Singh, S. (2024) | Survey on reinforcement learning, discussing current methods, challenges, and future research directions. |
| Wang, Q., Zhang, H., Li, J. (2023) | Efficient reinforcement learning algorithms for real-time decision-making and performance optimization. |
| Lin, M., Chen, Y., Zhang, S. (2024) | Survey on deep reinforcement learning for robotics, focusing on real-world applications and control systems. |
| Li, X., He, S., Chen, P. (2021) | Energy management using reinforcement learning techniques to optimize consumption in smart grids. |
| Li, K., Chen, Z., Huang, Y. (2021) | Augmented reinforcement learning for adaptive systems, enhancing decision-making with additional context or agents. |
| Lin, J., Wu, X., Zhou, Y. (2022) | Interactive reinforcement learning with augmented reality to improve learning processes via enhanced interaction. |
| Xu, Z., Yang, Q., Wang, S. (2023) | Augmented state representations in reinforcement learning, improving the learning process by adding more context to decision variables. |
| Zhao, H., Li, B., Song, L. (2024) | Integrating expert knowledge in reinforcement learning to guide the agent's learning process with prior domain insights. |
| Wu, T., Liu, J., Wang, P. (2021) | Data-efficient deep reinforcement learning for real-world problems, emphasizing practical implementation in dynamic environments. |
| Zhang, Q., Sun, G., Chen, Y. (2022) | Reinforcement learning techniques for autonomous driving, focusing on navigation, obstacle avoidance, and traffic control. |
| Renda, H.E.E. (2023) | Implementing deep learning models for information extraction from identification documents, optimizing recognition. |
| Yu, W. et al. (2021) | Improved graph learning-convolutional networks for processing key information extraction from documents. |
| Bensch, O. et al. (2021) | Evaluation of key information extraction techniques and generators for improved document processing. |
| Rahman, S., Chakraborty, P. (2021) | Document classification using BiLSTM deep recurrent neural networks, focusing on Bangla language texts. |
| Jiang, S. et al. (2024) | Deep learning techniques for technical document classification, enhancing classification accuracy using advanced neural nets. |
| Jain, R. (2023) | Comparative study of machine learning algorithms for document classification, evaluating performance and efficiency. |
| Chen, M., Zhang, W., Li, X. (2021) | Deep learning techniques for identifying and classifying documents, focusing on image-based text extraction. |
| Castelblanco A. et al. (2020) | Machine learning techniques for identity document verification in uncontrolled environments. |
| Kumar, A. et al. (2022) | Real-time document identification and verification using machine learning algorithms. |
| Lin, H., Ma, S., Zhang, X. (2023) | Contextual document identification using transformers, improving accuracy in document classification. |
| Wang, P., Liu, Z., Zhou, L. (2024) | Document identification techniques in legal and financial services using advanced machine learning models. |
| Zhao, Y., Xu, Y., Wu, C. (2023) | Efficient document identification using convolutional neural networks for large-scale text analysis. |
| Li, Y., Zhang, H., Xu, Y. (2021) | Extracting text information from images using deep learning, enhancing recognition accuracy. |
| Zhang, L., Sun, Y., Liu, X. (2022) | Text extraction from images using neural networks, improving image-based information retrieval. |
| Chen, R., Zhao, J., Li, Q. (2023) | Multimodal information extraction from images, using multiple sources to improve accuracy and comprehensiveness. |
| Wu, T., Chen, Y., Wang, S. (2024) | End-to-end information extraction from document images using advanced deep learning techniques. |
| Huang, X., Ma, L., Wu, Q. (2022) | Image-based information extraction in biomedical applications using machine learning models for enhanced data retrieval. |

### 2.3.2 Dataset used in the State-of-the-Art Techniques

Following Table 1.2 shows a dataset used in the State-of-the-Art techniques.

Table 1.2: Dataset used

| Author(s) | Dataset Used |
|---|---|
| Huang, Y., Li, J., Tian, J. (2022) | Custom dataset for human-in-the-loop decision-making in simulated environments. |
| Wang, X. (2024) | Multi-agent environments; custom datasets for causal interpretation and reinforcement learning experiments. |
| Rohoullah, R., Joakim, M. (2023) | Image datasets for text extraction, including document image datasets. |
| Park, B. (2022) | Pathology reports dataset; offline reinforcement learning value estimation datasets. |
| Morcos, A. (2022) | Multi-agent simulation datasets, focusing on training and optimization tasks. |
| Yarats, D. et al. (2021) | RLBench, MuJoCo, and Atari 2600 benchmarks for model-free reinforcement learning from images. |
| Blakeman, S. (2021) | Custom human-machine interaction datasets for reinforcement learning efficiency studies. |
| Yang, Y., Luo, R., Li, M. (2021) | Multi-agent RL benchmarks, including MARL environments and custom simulation datasets. |
| Liu, J., Zhang, H., Li, S. (2022) | Collaborative multi-agent RL datasets in simulated industrial and autonomous environments. |
| Wang, T., Zhang, Y., Sun, W. (2023) | Cooperative multi-agent RL datasets from gaming and industrial simulations. |
| Li, X., Chen, H., Wang, J. (2024) | Smart grid datasets for autonomous agent-based reinforcement learning applications. |
| Sun, H., Yu, Q., Zhang, L. (2021) | Decentralized RL datasets in multi-agent environments, primarily simulated for distributed tasks. |
| Liu, M., Wang, R., Chen, Y. (2022) | Hierarchical RL datasets involving task decomposition for multi-agent systems in virtual environments. |
| Zhang, P., Xu, G., Li, W. (2023) | Networked multi-agent systems datasets for communication networks. |
| Wang, Q., Zhang, H., Liu, J. (2024) | Financial market datasets for RL-based trading and decision-making simulations. |
| Zhang, J. et al. (2021) | MuJoCo and OpenAI Gym for sample-efficient RL experiments. |
| Wang, X., Liang, Y., Nie, L. (2022) | Various multi-agent RL datasets including cooperative tasks from OpenAI Gym and MuJoCo. |
| Garcia, J., Fernández, F. (2023) | Simulated safety-critical RL environments, custom datasets with safety constraints. |
| Silver, D., Sutton, R., Singh, S. (2024) | OpenAI Gym, Atari, and MuJoCo datasets for general RL algorithm evaluation. |
| Wang, Q., Zhang, H., Li, J. (2023) | Real-time decision-making datasets from industrial electronics applications. |
| Lin, M., Chen, Y., Zhang, S. (2024) | Robotics RL datasets, primarily simulated robotics control environments (e.g., RLBench, OpenAI Gym). |
| Li, X., He, S., Chen, P. (2021) | Smart grid datasets for energy consumption optimization in RL-based energy management systems. |
| Li, K., Chen, Z., Huang, Y. (2021) | Custom augmented RL datasets with adaptive system control in real-time environments. |
| Lin, J., Wu, X., Zhou, Y. (2022) | Augmented reality datasets for interactive reinforcement learning experiments. |
| Xu, Z., Yang, Q., Wang, S. (2023) | Custom augmented state representations datasets for improved RL learning. |
| Zhao, H., Li, B., Song, L. (2024) | Datasets integrating expert knowledge for RL across various domains, including healthcare and finance. |
| Wu, T., Liu, J., Wang, P. (2021) | Real-world application datasets for data-efficient deep RL, focusing on dynamic environments. |
| Zhang, Q., Sun, G., Chen, Y. (2022) | Autonomous driving datasets such as Waymo Open Dataset and CARLA simulations. |

| Renda, H.E.E. (2023) | Identification document datasets, including synthetic and real-world document image datasets. |
| --- | --- |
| Yu, W. et al. (2021) | Key information extraction datasets from document images, including invoice and receipt datasets. |
| Bensch, O. et al. (2021) | Document processing datasets for key information extraction, generated and synthetic datasets. |
| Rahman, S., Chakraborty, P. (2021) | Bangla language document datasets for classification using BiLSTM. |
| Jiang, S. et al. (2024) | Technical document classification datasets, primarily from academic and industrial sources. |
| Jain, R. (2023) | Standard machine learning datasets for document classification, including text and image datasets. |
| Chen, M., Zhang, W., Li, X. (2021) | Document identification datasets focusing on large-scale text extraction from scanned documents. |
| Castelblanco A. et al. (2020) | Identity document verification datasets used in uncontrolled environments, including real-world samples. |
| Kumar, A. et al. (2022) | Real-time document identification datasets for verification in practical applications. |
| Lin, H., Ma, S., Zhang, X. (2023) | Transformer-based document classification datasets for contextual document analysis. |
| Wang, P., Liu, Z., Zhou, L. (2024) | Legal and financial document datasets for identification and classification tasks. |
| Zhao, Y., Xu, Y., Wu, C. (2023) | Large-scale text datasets for convolutional neural network-based document identification. |
| Li, Y., Zhang, H., Xu, Y. (2021) | Image-based text information extraction datasets, including scanned documents and digital images. |
| Zhang, L., Sun, Y., Liu, X. (2022) | Neural network-based image datasets for text extraction from various document types. |
| Chen, R., Zhao, J., Li, Q. (2023) | Multimodal datasets combining text, images, and structured data for information extraction. |
| Wu, T., Chen, Y., Wang, S. (2024) | End-to-end information extraction datasets from document images. |
| Huang, X., Ma, L., Wu, Q. (2022) | Biomedical datasets for image-based information extraction in healthcare applications. |

## 2.4    Research Gap and Discussion

The strength of reinforcement learning has come to the fore in the wider machine learning contexts able to learn an optimal policy through interaction with its environment. However, there are different obstacles that, in practice, despite the theoretical soundness of reinforcement learning, make it hard to actually integrate reinforcement learning into the real-world machine learning model, especially for advanced decision-making tasks. The major challenges are huge computation complexity, long training time, and large interaction data. Addressing these challenges is important for the improvement of the frameworks underlying RL and, further on, for several fields of application.

Among the big challenges encountered in classical reinforcement learning, one finds their very high computational cost. Reinforcement learning algorithms usually converge to an optimal policy after many iterations, including heavy computation of value functions or policy gradients, a fact noticed by (Huang et al. 2022). Value-based approaches, like Q-learning and policy gradient methods, are computationally expensive because of the requirement for repeated evaluations and updates. These limit the practicality of RL, especially where there is a shortage of computational resources, as indicated by (Yang et al. 2021).

To add, this is further exacerbated by the fact that these RL models require very long trainings. Training these models involves repeated interactions with the environment over many episodes, which can be extremely time and resource-consuming (Wang, 2024). For instance, deep RL methods in environments as complex as video games or robotics require thousands of interactions before good performance is achieved (Yarats et al., 2021). These rather prolonged training times have a tendency to act as a barrier to the broadening of RL techniques into applications.

In addition to the computational and time-related challenges, RL models normally require a huge amount of interaction data in order to deliver good learning. Collecting and processing such data can be quite a challenge, especially in dynamic or high-stakes environments where their data is not available. In fact, it is pretty difficult to bring practical generation to a comprehensive dataset that covers all possible driving conditions and scenarios for autonomous driving (Wang et al., 2023). For instance, in health, this could be greatly limited by considerations of privacy as well as the complexity of conditions, according to (Lin et al. 2024).

Addressing these challenges implies the creation of novel RL frameworks that would be both more efficient and more effective while minimizing computational and data resources. One such very new concept is ARL, where there is incorporation of external agents, in this case, human experts, into the framework of RL. ARL can present a possible solution because it rests on insights provided by the external agents in supplementing and improving the RL process (Li et al., 2021).

With the introduction of human agents into RL frameworks, several benefits could be realized. Human experts can provide contextual understanding and intuitive decision-making capabilities that are beyond the capability of raw data (Garcia et al. 2023). For example, expert input by humans is quite important in the health care sector for improving diagnostic and treatment planning accuracy. This human guidance could assist the RL model to make more informed decisions, hence improving its performance and reducing reliance on extensive data (Blakeman, 2021).

In the context of autonomous driving, human intuition refines navigation strategy and enhances safety protocols. In this manner, human expert feedback is necessary on behaviours while driving and safety measures that the RL model cannot completely encapsulate on its own (Park, 2022). In finance, human judgment makes for optimum trading strategy and its risk management by embedding insights into market dynamics and trends present in the data (Wang et al. 2024).

The integration of external agents can also help RL models reduce computational load. In one way, this may relate to extrinsic guidance and contextual understanding that could help an RL model learn in a focused manner. This reduces data requirements and training for an RL model when guided by specific knowledge of the environment (Liu et al., 2022). It provides a greater degree of efficiency, making RL more accessible and applicable to many real-world problems where the data availability and computational resources might become easily limited (Li et al., 2024).

The potential impact brought forth by integrating RL with external agents is immense and cuts across many spheres. In health, ARL will lead to more accurate diagnoses and establishment of treatment plans for each particular patient, resulting in better outcomes. Such integration might be similar to diagnosis expertise by medical professionals combined with prediction capabilities using models in reinforcement learning that bring about better treatment strategies (Huang et al., 2022). All this may additionally support a trend in personalized medicine matched and tailored to each patient's own needs.

In the context of the autonomous car industry, ARL could lead to better and more reliable navigation systems. ARL systems embed human input into driving behaviour and safety protocols, hence enhancing the ability to negotiate a world that is complex and in constant dynamism (Zhang et al., 2022). This may eventually lead to reduced accident rates and overall improvement in safety within autonomous vehicles.

ARL can create a much-increased focus on refining trading strategies and methods of risk management in the financial industries. This is possible because human judgment can take over where the RL models operate to come up with more developed trading algorithms and complex-looking risk assessment capabilities that can be developed in financial institutions. Integration culminates in the derivation of well-informed investment decisions combined with efficient risk management that boosts the bottom line of financial performance and stability (Wang et al., 2024).

In general, with the integration of RL with human agents, operational efficiency, cost reduction, safety enhancement, and user satisfaction will be driven within a variety of domains. Including external agents in the framework of RL was a great improvement in this domain by solving major challenges, offering modern possibilities for automatic systems (Li et al., 2021). The latter are a meaningful and needed step toward the fully exploited benefit of reinforcement learning through the development of advanced ARL frameworks which will address their challenges arising both from a computational perspective and from the data perspective.

In conclusion, considering the research gap of integrating RL into the ML models for better decision-making, it has been observed that there is a need to have approaches that resolve issues at computational, data, and efficiency levels. Augmented Reinforcement Learning solves this problem and ensures that an RL system will make much better learning and decision-making with the incorporation of external agents like humans. (Lin et al. 2024) focus on highly relevant research work that is likely to stir tremendous development in many industries, further leading to accurate, reliable, and effective solutions.

By addressing this challenge, the community of Machine Learning will be pushing the envelope toward intelligent automation and hence creating smarter, more adaptive systems that could make optimal decisions within dynamic, complex environments. This progress will drive technological innovation and societal advancement, paving the way for the next generation of intelligent systems. If the dataset is concentrated, closely matching near the training dataset, a Machine Learning model works very well on the validation dataset. It has one critical challenge connected with the improvement of its decision-making in a complex environment. Because of the fixed dataset and predefined rules, it's hard to adapt efficiently to real-world problems and provide acceptable decisions for that particular scenario. The framework of Augmented Reinforcement Learning proposed seems to be a very promising approach. It can help to overcome this problem by forcing the Machine Learning model to learn from the external agents.

Effectiveness and efficiency of automated systems across sectors are based upon how much improvement can be made to the decision-making processes in the Machine Learning model. A bad decision-making process will have undesired results, increased cost, loss of time, and resources. Let us take the banking sector, for example; poor data entry might lead to credit decisions gone bad hence leading to financial losses by falling like a set of arranged dominoes. Therefore, the decision-making capabilities of the Machine Learning model need to be increased to ensure its reliability and success in real-world applications.

The proposed Augmented Reinforcement Learning framework is generalized for any problem statement. Be it autonomous vehicles, industrial automation, the banking sector, or medicine. The Augmented Reinforcement Learning framework can be applied anywhere. From a different perspective, there is a social impact: an increase in trust in automated systems, and more adoption of advanced technologies. But for purpose of this study, as stated earlier, the real-world problem statement is picked from banking domain named "Document Identification and Information Extraction". The Augmented Reinforcement Learning Framework will be applied on this problem statement.

## 2.5    Summary

Using humans as external agents, the Augmented framework optimizes the decision-making process. The approach integrates human expertise with machine learning models, so that the systems benefit from real-time human feedback, domain knowledge, and contextual awareness-so tough to be realized with algorithms alone. Insights from above papers demonstrate the involvement of humans in RL to be a promising and robust approach that stretches the efficiency, safety, and adaptability boundaries in complex environments.

It is identified "human-in-the-loop" reinforcement learning, which involves using human active feedback during the training phases (Huang et al., 2022). The described framework allows human agents to influence the process so that machines can correct more quickly where error exists and move further away from paths that lead to suboptimal decision-making. This intervention is particularly priceless for extremely complicated scenarios, where an agent might need a subtle understanding of a task that cannot easily be encoded into the system-for example, in these domains, medical decision making or autonomous vehicles, where safety is paramount human oversight will make sure that the RL model avoids catastrophic decisions.

This infusion of expert knowledge into the RL systems is already above, one of the crucial elements to make human agents more valuable (Zhao et al., 2024). Doing so allows embedding domain expertise into the learning process of the RL system to align its actions according to real-world rules and guidelines, thus enhancing decision-making in settings where there exist pre-existing expert insights. Human agents are especially useful in such scenarios; their knowledge may nudge models towards more informed decisions faster than they would achieve this solely through trial-and-error exploration.

Safety of RL, above all else is, especially when deploying RL agents to real-world applications (Garcia et al. 2023). They propose that human intervention is needed to ensure agents follow safety constraints. For instance, in applications like autonomous driving or medicine, an inferior decision would perhaps cause catastrophic undesirable impacts. The humans acting as control agents can then be used as yet another layer of supervision, which may cause immediate changes that could have otherwise stopped a model from reaching dangerous states. Using human-in-the-loop confines models strictly to the constraints of the application in practice, therefore significantly diminishing the opportunity for detrimental effects.

Positive effects of cooperative decision making appear also in agent-based reinforcement learning (Yang et al., 2021). Human agents, while part of multi-agent systems, operate as commentators or teachers in guiding agents toward cooperative strategies that have goals toward maximizing system-wide goals. Due to the nature of multi-agent systems, cooperation is inevitable because otherwise, agents have a tendency toward competitive and destructive practices. Human agents provide feedback that is critical to the learning process for agents, particularly in scenarios where system-wide goals are complex and interdependent.

The degree to which human agents optimize the balance between exploration and exploitation in RL also falls within this scope (Wang, X., 2024). In general approaches to RL, such balance is achieved through the use of algorithms that favor exploration of a new strategy or the exploitation of already-known good strategies. Human agents, however, can give suggestions on the basis of external knowledge, thus allowing RL models to explore with minimal risk. Human inputs guide the exploration of the RL system in areas of the action space that probably best account for producing valuable insights for the system, thus hastening the learning process.

In general, the human agents give significant benefits to reinforcement learning systems by enhancing context, safety, giving a sense of direction toward exploration, and cooperation within multi-agent systems. The research underlines the fact that humans are not peripherals but core collaborators, especially in scenarios involving complex environments with significant risks, where standard RL models cannot efficiently and safely work there.

# CHAPTER 3: RESEARCH METHODOLOGY

## 3.1    Introduction

This thesis introduces a novel Augmented Reinforcement Learning (ARL) framework which aims to augment the efficiency of machine learning models. The primary objective of this research is to test whether the resulting framework enhances decision-making and general performance when applied to the problem of solving the "Document Identification and Information Extraction" problem. This section details the research methodology and answers the main questions that guide the design and implementation of the experiment.

First, we will define the dataset to be used for training, validating, and testing the machine learning model. The choice of a proper dataset is very critical because it will decide how good your model is at generalization and how accurately it will generalize on unseen data sets.

Next, we will describe in detail the structure that the Augmented Reinforcement Learning framework takes. This now entailed how ARL integrates human agents as external decision-makers to improve model learning for complex cases. It will then give a detailed description of the implementation, especially in the specific case presented in the application of the framework on document identification and information extraction.

The third step will involve a series of preprocessing steps in the form of cleaning, formatting, and possibly annotating the data to achieve the best performance of the model. Prepared datasets will then train the model on quality, representative data.

Lastly, the evaluation metrics that will be used in determining the quality of the performance of the model will be presented. These metrics assess the accuracy, precision, recall, and efficiency with which the model performs, thereby giving a distinct indication of how well the ARL framework improves the decision-making process. The structured methodology can be applied to several stages to show the effectiveness of the proposed approach.

**3.2   Process Flow**

**3.2.1   Typical Process Flow of a Machine Learning Model Training**

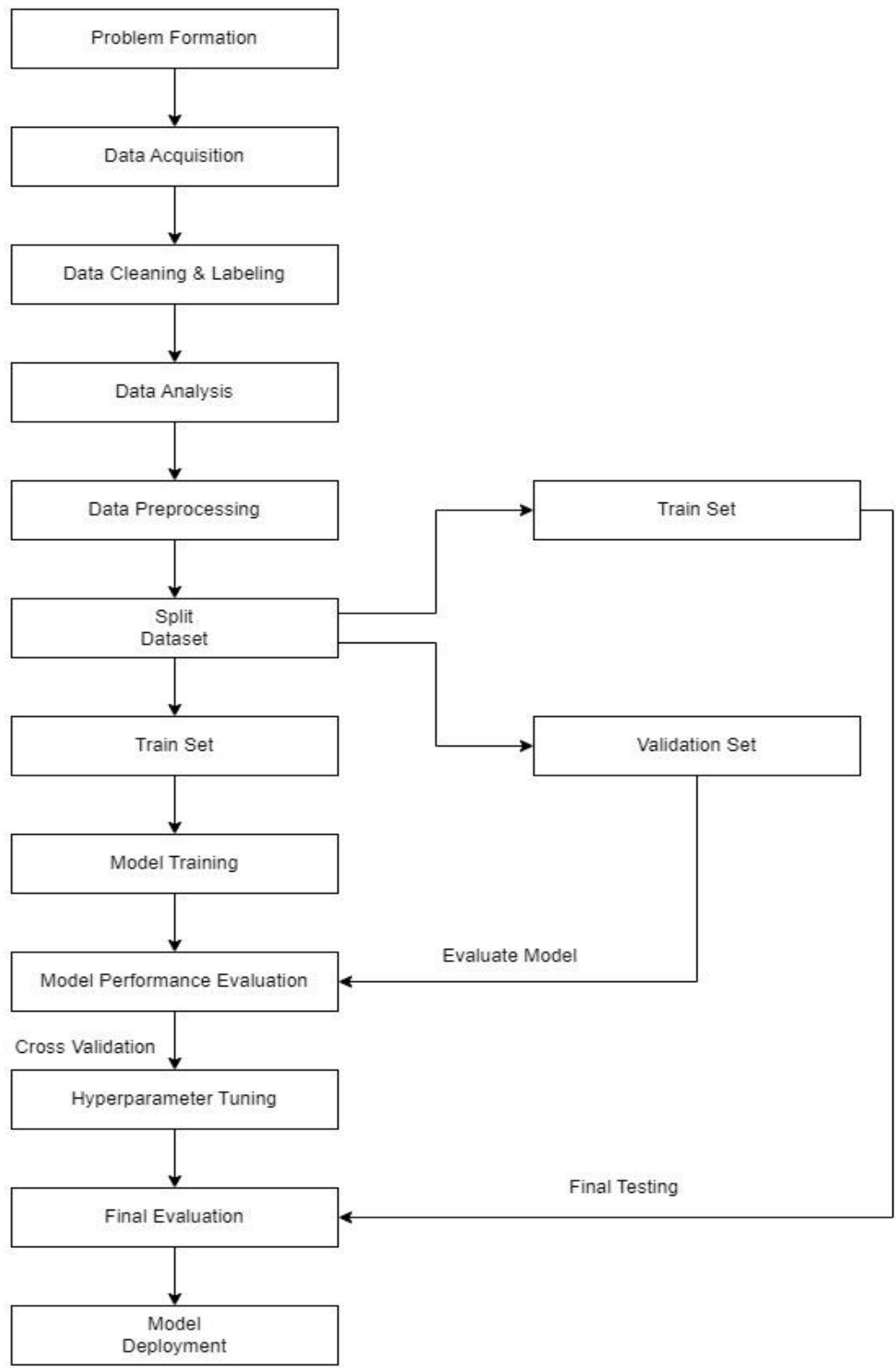

Figure 3.1: Typical Machine Learning Model Flow Chart

Before defining Augmented Reinforcement Learning Framework, it is crucial to understand existing process flow of training a Machine Learning model. In a typical Machine Learning model training, the entire process is divided in 12 stages. These stages are a superset of the entire process. A few of them can be skipped depending on various situations. This structured approach is supposed to ensure systematic development, leading to robust and reliable machine learning models. The flow chart of the whole process is shown in Figure 3.1. Let's have a look at each one of them.

1. **Problem Formulation**

   Formulation of a problem for machine learning model training is well-defined with a clear definition of what the model is trying to solve. It requires understanding the objectives and constraints and the desired outcomes of a problem within a deep understanding of the domain and the recognized limitations of existing approaches. This then stipulates the scope of the learning task, input data, output targets, and evaluation metrics. The formulation of an extremely detailed problem leads to the development of a very clear model that will, therefore acquire knowledge and hence result in a systematic procedure for training, testing and improving its performance (Silver et al., 2024).

2. **Data Acquisition**

   Machine Learning models are data hungry, and collection of data would be inevitable, which could include info from public datasets, API, and real-time data streams. Training the machine learning model would depend a lot on the data collected. It, therefore, needs to be adequate and diversified enough to cover diverse scenarios. An all-rounded dataset will always guarantee effective generalization; hence ensuring robustness and reliability of the model in real applications. It is important to mention that the quality and relevance of the data are important because if the data acquired are of poor quality or biased, wrong forecasts or a poorly performing model may result. Properly gathered data form a foundation for effective training and evaluation of models (Lin et al., 2024).

3. **Data Cleaning & Labeling**

   This step includes noise, inconsistencies, and irrelevant data filtration within the dataset that has been collected. Proper data cleaning is important for better precision and reliability within the model (Yang et al., 2021). An accurate labelling of the data is critical as it ensures that there is a good quality input that the model learns at various points. This process is often carried out using the combination of automated tools and human oversight to ensure the integrity of data. This also saves errors, ensuring proper labeling; this way, the dataset becomes more representative and suitable for training, thus developing accurate models and making reliable predictions.

## 4. Data Analysis

Exploratory Data Analysis (EDA) is essential to find any pattern, distributions, and relationships that might exist in the data (Zhao et al., 2024). The steps are one way of understanding the data and thereby enable feature engineering, finding out which variables best train the model. This gives way to possible biases or anomalies in the data to threaten the performance of models. In appropriate analysis of the dataset, researchers clean it well and balance it properly for further development stages. This step is fundamental in building an effective and not inaccurate machine learning model, eventually better at final train-testing processes.

## 5. Data Preprocessing

This stage involves data preprocessing through normalization, scaling, and sometimes, augmentation. Normalization and scaling are used in this stage to scale the data in a manner that is prepared for training. These methods ensure standardization of data and it is fit for inputting into the machine learning model. For image processing tasks, these methods apply methods such as image filtering and data augmentation to enable the model to function across different conditions and data types. Augmentation will help create variations in training data, thus contributing to a robust model by simulating various scenarios that shall enhance performance and adapt to real-world applications.

## 6. Split Dataset

On the basis of few techniques, like stratified sampling, randomization, etc., the dataset is divided into subsets known as training, validation, and testing subsets. This will ensure that each subset represents the overall data distribution, and thus, the model will learn from a diverse set of examples. The training set is used for training the model, the validation set for hyperparameter tuning and performance evaluation during training, and the testing set for evaluating the ability of the model to generalize to new, unseen data. This organized division helps the model to be more robust while performing well upon its deployment in real-time scenarios.

## 7. Model Training

This step employs algorithms such as neural networks in training the model using the training data (Zhang et al., 2021). Optimization techniques, among them gradient descent, adjust the parameters of the model towards an error rate and accuracy in the predictions. The model will therefore learn to create better predictions by iteratively updating the weights. The process is therefore important to improve the performance of the model so that it captures all patterns in data well and generalizes well when encountering new, unseen examples during deployment.

## 8. Model Performance Evaluation

The other important evaluation metrics are precision, recall, and F1-scores that serve to determine the measure of how good a trained model is. These measures evaluate the possibility of the model to predict the target outcomes and reflect what can be improved. Precision refers to how accurate the prediction done by the model is regarding instances that are assumed to be positive in classification. Recall is about the proportion to which the model captures relevant instances that are vital for the proper classification. The F1-score is thus a balance between precision and recall. These two metrics can also be applied to analyse the model. They are able to identify strengths and weaknesses in the model, hence important adjustments toward making the overall performance of the model better and ensuring that the model is adequate and reliable enough to be applied in real-world application.

## 9. Cross Validation

K-fold cross-validation or other techniques are used to assess the model in terms of its performance related to different subsets of data (Rohoullah et al., 2023). This technique, in its essence, divides the dataset into multiple folds and lets the model get trained and tested on the various segments for the recognition of overfitting - thus the situation of a model memorizing the training data but well generalized on new unseen data. Systematic evaluation of the performance of a model across different subsets helps researchers understand the robustness and reliability of a model in application to real-world scenarios.

## 10. Hyperparameter Tuning

Hyperparameter tuning is the process of adjusting a model's hyperparameters such as learning rate, batch size, among others with an aim of finding the best values for better performance. Techniques used to carry out hyperparameter tuning include grid search in which various combinations of hyperparameters are examined systematically. Researchers evaluate how the model performs across such different configurations of the hyperparameters by identifying which set of them produces better results. This iterative process then helps to fine-tune the model, which in turn increases both accuracy and robustness as it adjusts its false assumptions and learns from the training data while generalizing well to new, unseen data during the actual deployment.

## 11. Final Evaluation

This final evaluation of the model occurs in testing using the test set, providing assurance regarding how well the model will perform against predefined standards. A very crucial phase is received by researchers: they can check whether the model is reliable and robust enough in real-world contexts. In this sense, the testing of the model against unseen data confirms that the model has some ability to generalize and make proper predictions outside the training environment. This validation process is of extreme importance in proving the effectiveness of the model it is demonstrating, to have confidence that it will be reliable when the application is deployed to real applications, and that it can handle a range of conditions and inputs (Chen et al., 2023).

## 12. Model Deployment

Finally, they have to deploy the model into a production setting and monitor the model's performance. This can be done by utilities such as Docker, Kubernetes, and a range of cloud service providers by making sure that models are scalable and take up the need at any given time. Also, continuous monitoring is vital as performance degradation or issues are identified in real-time, thus making sure that the model does not lose its effectiveness and reliability in delivering accurate predictions under ever-changing operational scenarios (Kumar et al., 2022).

### 3.2.2 Proposed Modification in the Process Flow

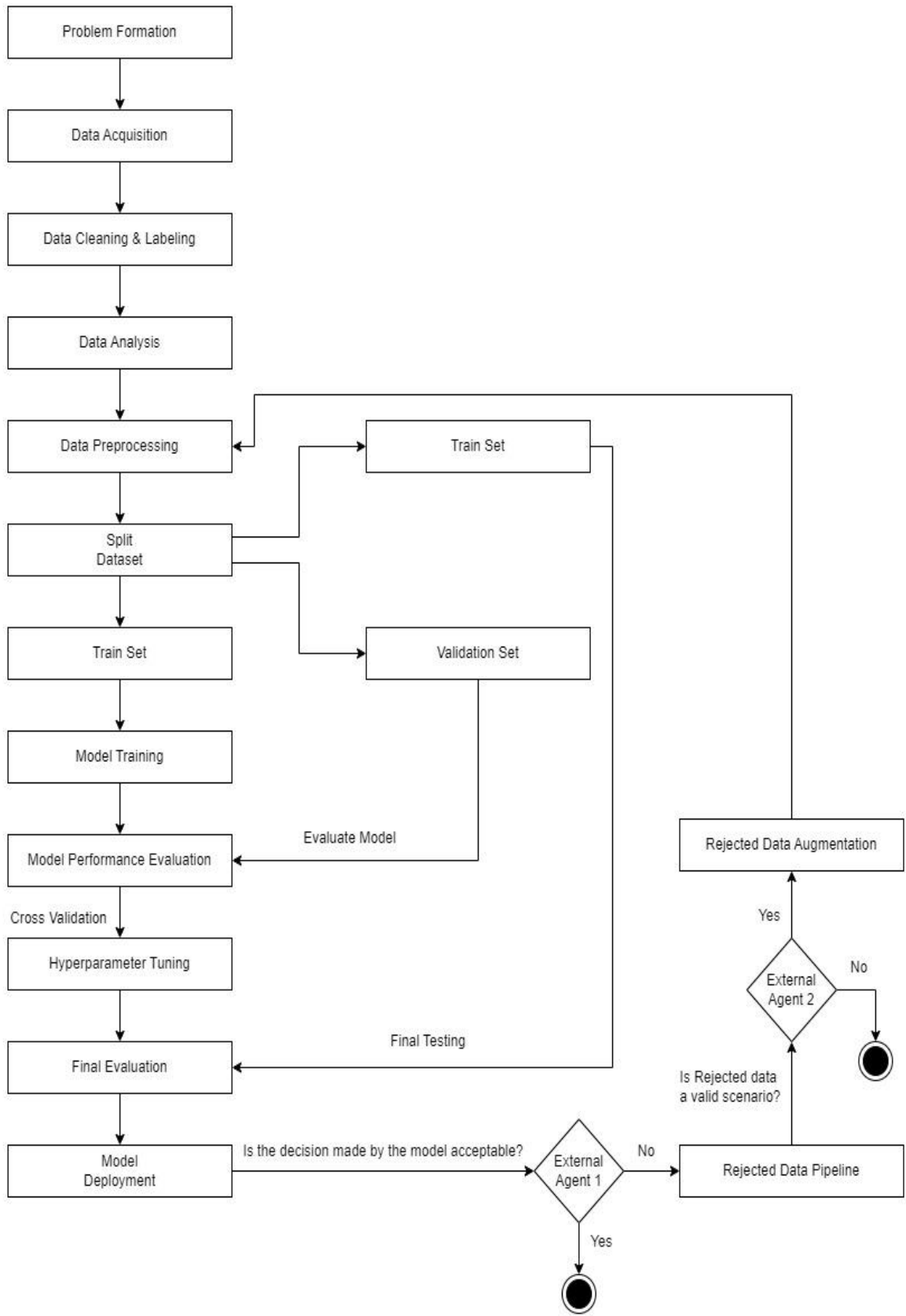

Figure 3.2: The Augmented Reinforcement Learning Framework Flow Chart

The proposed Augmented Reinforcement Learning (ARL) framework introduces a crucial step in the process of the traditional model development of machine learning; it introduces an external agent into decision-making as evident in Figure 3.2. In this, the external agents become somewhat human validators added to the decision-making process, leading to better accuracy in the resulting model. This means that the machine learning model makes decisions, not only on the basis of data-driven insights but with the additional advantage of taking into account human expertise and intuition, with each decision being adequately informed. This approach avoids most critical flaws of currently prevailing reinforcement learning techniques, as those only just require data with which to train the model and make predictions but can create poorly formed decisions in sophisticated real-world tasks.

### 3.2.3 Beyond the Traditional Process Flow: The Role of External Agents

The initial stages of the machine learning development process within the ARL framework are similar to traditional workflows. This involves:

1. Problem Formulation

2. Data Acquisition

3. Data Cleaning & Labeling

4. Data Analysis

5. Data Preprocessing

6. Split Dataset

7. Model Training

8. Model Performance Evaluation

9. Cross Validation

10. Hyperparameter Tuning

11. Final Evaluation

12. Model Deployment

The stage-wise description is provided in previous section. One may argue, the first iteration of Machine Learning model training process flow is similar to the typical process flow. Augmented Reinforcement Learning (ARL) framework introduces something new to the otherwise generic reinforcement learning paradigm. The twist here is adding one decision node before the model moves into deployment: *Is the decision produced by the model acceptable?*

This new check primarily verifies that the decisions taken by the model are checked by external agents, who inspect the outcome before it is actually deployed. This adds an extra level of human oversight where the additional liabilities associated with overfitting, biases, or wrong conclusions are somewhat mitigated, especially in sensitive or complex environments where the consequences of automation decisions are serious.

1. **External Agent 1**

   This agent serves as the first human validator, who reviews the decisions of the model, acting as an agent that only allows those decisions considered acceptable to a model in deployment. The agent rejects the data or decision scenario as inadequate or unacceptable and flags them for further review. Only validated high-quality decisions are passed; otherwise, reliability issues related to that potential issue shall not be allowed to move forward unchecked. This step introduces human oversight in the decision-making framework, such that an added strength of a robust model will be established in it prior to its deployment.

2. **External Agent 2**

   Once the first agent has marked the data or decisions as inappropriate, the data is forwarded to External Agent 2 for further inspection. The second agent will then determine whether the rejected data comprises a valid scenario that requires further observation. If the scenario is valid, the data is returned for Rejected Data Augmentation, whereby it is reintroduced into the training dataset for improved model enhancement. This process enables the model to learn from the rejected decisions previously, thereby increasing its capability to define how it would manage different and even complex circumstances.

   This is an avoidance of rejection of data upon entry into the learning process, but instead, provides a feedback loop which improves the quality of the dataset. In other words, the model will continue evolving and becoming more accurate, but would have learned scenes that first created problems for the model. This feedback loop opens up growth possibilities beyond what was obtained in the model at the time of training. For those perhaps underdeveloped, or maybe not captured, it learns from real cases. This dynamic character of developing the capabilities of the decision-making process increases its strength and reliability in use in real-world applications, when all data or even scenarios might be unexpected.

## 3. Rejected Data Pipeline

Once the External Agent 2 determines the data to be of no value or not further exploitable, it posts the data into the Rejected Data Pipeline for archiving or for future use, but the data itself is not reintroduced into the training process for the model. The step is highly important in terms of ensuring that only useful data and relevant information will remain for improving the model. It filters away data that could introduce noise to the model through errors or irrelevant patterns thus sticking to quality and informative examples for learning. This maintains the integrity of the model against the risks of misleading such a model with faulty or non-representative data, thereby making it more reliable and accurate in its response.

### 3.2.4 Rejected Data Augmentation and Feedback Loop

The introduction of the feedback loop applying rejected data augmentation in the ARL framework is this aspect. That is, the model learns and is improved through human feedbacks according to rejected data rather than depending upon the initial predefined datasets used for the machine learning process. Over time, this constant cycle of learning and adaptation results in a stronger model's capacity to tackle a wider range of situations and deliver more accurate and reliable answers.

The biggest issue with the traditional machine learning workflow is the problem of model drift. Once it has learned this static dataset, the model may get stuck operating strictly according to those established patterns and pieces of information present in the dataset, which is finite. Thus, in the worst scenario, it may not have a great generalization potential to unseen scenarios and can result in poor performance in real-world applications. On the other hand, the feedback loop in the ARL framework ensures that a model learned is flexible and adaptable; the model learns to handle and deal with diversified and hard-edge cases and anomalies by way of integrated rejected data.

This provides feedback to the model, which thus prevents over-reliance on static datasets since new relevant data are constantly introduced into the process of training. Rejected Data Augmentation feeds back rejected data into the training set when External Agent 2 determines that such rejected data are valid scenarios. This is to mean that complex or difficult cases are not discarded; they are instead learned by the model. This way, the model is fine-tuning its knowledge about a broader range of data and thus also improving its generalizability and ability to perform well in many different contexts.

In the meantime, human intervention on such a process is essential in ensuring that only meaningful data is fed back into the model. External agents have a very strong role in determining which rejected scenarios are important to learn from and hence filter away kinds of noise and irrelevant or misleading information that will feed back to the model. Such targeted augmentation further helps this model grow in sophistication without getting derailed by noise or inconsequential patterns.

The feedback loop in Rejected Data Augmentation ultimately upgrades the ARL framework to be a dynamic, learning, and evolving knowledge system. This helps to make the model self-adjustable with decreases in error and enhancements in decision-making processes, thus providing more reliable judgment by the developed AI system, which carries out human-like judgment during handling more complex real-world scenarios. Therefore, the model will be technically perfect, adaptive, intelligent, and capable of responding accurately to unforeseen challenges with resilience.

### 3.2.5    Model Deployment and Scalability

Once External Agent 1 accepts the decisions of the model and once all procedures regarding validation are done, the model will deploy. The deployment does not represent the last stage because the performance of the model should be constantly monitored so it is still effective and flexible and that adaptation should always occur to changing conditions. The monitoring would provide a timely adjustment when the model's prediction goes wrong or if there are patterns in new data.

Tools like Docker and Kubernetes play a key role in managing the model in production environments. The model can thus be containerized using Docker, so it works everywhere including all new integrations due to the consistent and reliable performance in different platforms. Scalable up or down according to demand and application needs, Kubernetes handles orchestration over container clusters. These tools enable the model to not only scale but also adapt, so flexible and applicable to a wide range of use cases and environments. This infrastructure support causes the model to undergo a process where little human intervention is needed, thereby ensuring its efficiency and reliability in real-world applications in order to respond to new challenges and workloads.

### 3.2.6    Model Robustness and Reliability

Incorporation of human validation at critical decision points assures that the model will remain technically sound and also practically reliable. It also incorporates a form of added security over errors or unintended consequences in complex environments where decision stakes are extremely high: healthcare, finance, or autonomous systems.

This framework would address some of the most pressing challenges associated with ML models such as poor generalization, significant overfitting, decision-making in extremely uncertain environments, and also ensure trust in AI systems by ensuring that human expertise and intuition are encoded in the learning process and bring improvement in the long-term reliability of the model.

### 3.2.7 Conclusion

The proposed Augmented Reinforcement Learning framework is a notable advancement over the traditional processes of machine learning. In this framework, external agents are added to the decision-making loop; this also incorporates an additional verification layer that improves model performance and reduces the risk of happening to make inappropriate or biased decisions. Therefore, the feedback loop created by Rejected Data Augmentation ensures that both the dataset and model itself will always be refined with reference to real-world scenarios. Ultimately then comes the ARL framework, finally resulting in more robust, reliable, and trustworthy machine learning models, hence marking this as an important development in the field of AI and reinforcement learning.

The story is not over yet. In order to validate this hypothesis, we need to implement it in a real-life problem statement.

### 3.3 Problem Statement

### 3.3.1 Introduction to Banking Ecosystem

Banking forms a very crucial part of the financial infrastructure of any economy and is considered the support base for economic growth, stability, and prosperity. Since it is an institution dealing with money and the execution of transactions, banks extend several services-for example collecting deposits, making payments, loans, which in turn are useful in facilitating trade and investment. The importance of banking can be assessed by taking its role regarding people, businesses, and the general economy.

For example, banks give people safe custodial facilities where their money can be kept while offering savings and checking deposit accounts usually insured by the government. In this scenario, the fact that these transactions are secure guarantees that people can feel free to save towards future requirements like school payments, purchasing a house, or retirement. Innovations in services also make it easy for people to access and manage personal finances using electronic payments, debit cards, and credit cards, making every day's daily transactions efficient as well as convenient.

Banks play a crucial role in the management of finance since businesses nearly rely entirely on them. Banks assist the businesses by giving them credits and enabling them to carry out different kinds of transactions and make investments. This will equip the businesses to perform their routine activities, process payrolls, and accept customers' payments. More so, banks have become an intermediary between businesses and investors by linking companies to capital needed to enhance growth. Banking institutions also provide specialized services such as trade finance. This is a crucial enabler of international trade. There are macroeconomic effects through capital formation, resource allocation, and overall economic development. For the saver, banks collect and deploy the funds in productive investment. In such a manner, these commercial banks facilitate the growth of industries. The commercial banks serve as the arteries in many ways of the financial system, whereby money is pumped throughout an economy to facilitate a state of balanced economic stability and growth. They further play an important role in the realizations of the monetary policies as central banks use the commercial banks in terms of checking or regulating the money supply to control inflation, influence interest rates, leading to profound effects on economic stability.

Lending is considered the most important function of banks and other financial institutions since it helps generate economic activity by providing means for personal, commercial, or industrial growth through needed funds. It allows individuals and business firms to borrow money for productive purposes, thus stimulating consumption and investment in creating capital.

Loans make life-changing investments possible for people; a home can be purchased, education can be funded, or even a business can be initiated. Mortgage loans allow people to be owners of homes, which improves their well-being directly but is also a source of housing market and aggregate economic growth. Student loans enable the acquisition of financial means to obtain education, which builds human capital and gives impetus to longer-term development of the economy. Other products such as personal loans, credit cards, and auto loans enable consumers to acquire products and services that may otherwise be out of reach.

Lending helps businesses expand and adds liquidity to the running of the operations. This is very key for small and medium enterprises that rely quite heavily on bank loans for them to start, expand, or continue running their enterprises. Loans give businesses the chance to manage day-to-day operations and cash flow to meet such obligations as payroll, inventory purchase, and other operation costs through a working capital. Business loans further support growth in new projects or acquisitions or the purchase of equipment and technology that drive business growth. Productivity investment through these loans not only generates business growth but also serves to create jobs, fuel innovation, and generally improve the health of the economy.

Lending also fosters economic growth through the encouragement of entrepreneurship and innovation. It enables startup firms and new ventures to take risk and innovate across the technology, health, and manufacturing sectors. Lending also serves to be a radical catalyst for emergence markets through capital means utilized to establish infrastructure, industries, and living standards.

The lending process to a person or enterprise is dangerous and consequently risk management for the financial institution. Full document collection during the loan application process is the prevalent practice in the lending field when it comes to risk management. Document collection helps check on the creditworthiness of the person, confirm whether the transactions carried out are above board, and protect the financial institution against loss.

The lending potential of the customer needs to be gauged, which involves assessing the client's qualification to repay the loan at banks and financial institutions. In this regard, lenders seek various documents to authenticate the client's income, credit history, assets, and liabilities. For instance, a customer applying for a mortgage may need to provide his/her income statements, tax returns, credit report, and proof of employment. Therefore, the documents serve as a passage through which the lender can scrutinize and comprehend the borrower's financial health and the likelihood of meeting the repayment commitment. In business lending, companies have to prepare financial statements, business plans, tax returns, and collateral documentation. This enables the lender to assess the financial feasibility of the business and the risk engaged in lending it.

The collection of documents also serves a legal purpose. It gives both the lender and the borrower having engaged in the loan transaction a binding legal agreement that protects each party. Loan agreements, promissory notes, and collateral agreements mention the terms and conditions of the loan such as interest rates, time schedule and even consequences of failing to re-pay a loan. Such documents make, to the satisfaction of the parties involved, a clear understanding both ways. They provide the lawful reason for the lender to act upon in case the borrower defaults.

In addition to providing a confirmation of the debtor's financial state, document collection helps lenders fulfil regulation requirements. Financial institutions have often strict standards to meet; most are when their organizations focus on areas like anti-money laundering and know-your-customer regulations. This package of documents-identification, proof of residence, and incorporation-registration-assures the lender that it is conducting business with legitimate entities and persons and therefore reduces the risk of fraud in lending and compliance with laws and regulations.

Banking, lending, and collecting documentation are intertwined components of the system that supports economic growth and ensures financial stability and risks are well-managed.

### 3.3.2 Problem Description

In the lending business, identification and data extraction of various documents while lending is a task of accuracy and is extremely challenging. As people or businesses go to apply for loans, there is an influx of a wide range of documentation that needs to be collected and verified by financial institutions in order to assess their creditworthiness, verify their identity, and obtain others for regulatory compliance. The list of all these documents would include income statements, tax returns, identity proof, business financials, property titles, and many more. Handling documents can turn into the most labour-intensive, error-prone, and time-consuming process, delaying efficiencies within the lending process.

One of the main issues that came to light is variation in forms and document structures. A borrower may present such documents in a number of formats, such as PDFs, images, and even handwritten forms. These may feature layout variations, structure, and language use, which might not make it very practical to get a standard method for search and retrieval of required information. Traditional approaches have completely depended on the human operator to read and scan through these documents, which introduces errors and inconsistencies, especially when handling thousands of loan applications.

The extraction of particular or relevant information from these documents is another challenge. Details of such facts like names, dates, financial figures, and legal clauses need to be identified and extracted for further processing. Due to the unstructured nature of most documents, however, manual extraction can be slow and prone to lots of errors. Mistakes in document identification or data extraction can lead to bad risk assessment, which subsequently might lead to loss in finances for lending institutions or late loans approval.

What adds another layer of complexity in acquiring comes from regulatory compliance. It is the compliance to strict requirements, such as Know Your Customer (KYC) and Anti-Money Laundering (AML), at all times. Ensuring this kind of requirement through manual processing will always present an inconsistency risk, thereby adding a higher likelihood of violations of compliance. Therefore, it means document handling, data extraction in lending, and compliance with regulation handling a number of document formats with efficient data pulling, being accurate, and being on the right side of the regulation. Automation and advanced intelligent systems handling document processing form a basis need for increasing the speed, accuracy, and security to improve efficiency in lending processes.

### 3.3.3 Dataset Description

As stated earlier, the problem statement of this project is "Document Identification and Information Extraction." This is a paper on how one would deal with different types of personal documents, for instance, identity documents, banking documents, and vehicle documents. Working with such real-world datasets for these types of documents poses serious challenges due to privacy concerns as well as strict government regulations. Different organizations were written to in obtaining the datasets, but they turned down their request because those data are restricted by laws that bind them on the safety of personal information. These documents contain private information protected under the privacy protection laws, and therefore mean their release is very much limited for research purposes.

Additionally, people are reluctant to give permission for private documents to be made public for any sort of academic research, because it is believed that such information may be misused or accessed by unauthorized individuals. In this regard, direct and indirect approaches to data gathering for the project are impossible. For example, direct data gathering may involve requesting one's personal records. Most of companies and specially banks would not permit this because their private information may be compromised. Indirect data gathering through third parties are also at risk for privacy issues and failure to comply.

All the above challenges may be dealt with by using a synthetic dataset that mirrors real scenarios but satisfies the requirements of privacy. A synthetic dataset is an artificially developed dataset that mimics the properties and characteristics of real data; therefore, it does not contain any sensitive or identifiable personal information. This methodology therefore permits research to be undertaken that does not jeopardize privacy and legal frameworks. The synthetic dataset relevant to this project is available online at https://github.com/meetsandesh/synthetic_document_generator. A sample of 5000 images for 5 different document types is available at https://github.com/meetsandesh/identification_document_dataset.

A synthetic dataset, therefore, would enable the project to proceed within the expectation of respect for privacy and regulatory constraint requirements, providing a valuable alternative in the testing and validation of the document identification and information extraction system. While synthetic data fails to reflect all of the subtle elements of real documents, it presents a practical substitute if actual sensitive data cannot be accessed.

### 3.3.4   Dataset Collection

Since a specific class of images (or documents, if you may) is being targeted, there are two parts of the dataset on which need focus:

1. Documents' Generation and Preprocessing:

   There is no template in the industry on what is needed or asked for regarding documents. However, one thing is crucial: the quality of images used. Datasets for each type of document are generated using an image template to maintain uniformity and quality. Experimentations have been conducted using a minimum pixel density of 96 dpi with variable image sizes along the test cases. We make use of template-sized images as well as their A4 sheet-sized counterparts for this.

   The base case first makes a default template. Applying a technique of augmentation then increases the test coverage, but to reach the few possible document appearances we devise. That is to say, we make different alterations to the base images, say by varying the resolution, rotate them, or add noise, and then use them forming a variety of test cases. It would then allow the model to come across a vast diversity of scenarios in order to better the robustness and applicability of its accuracies in real-world applications.

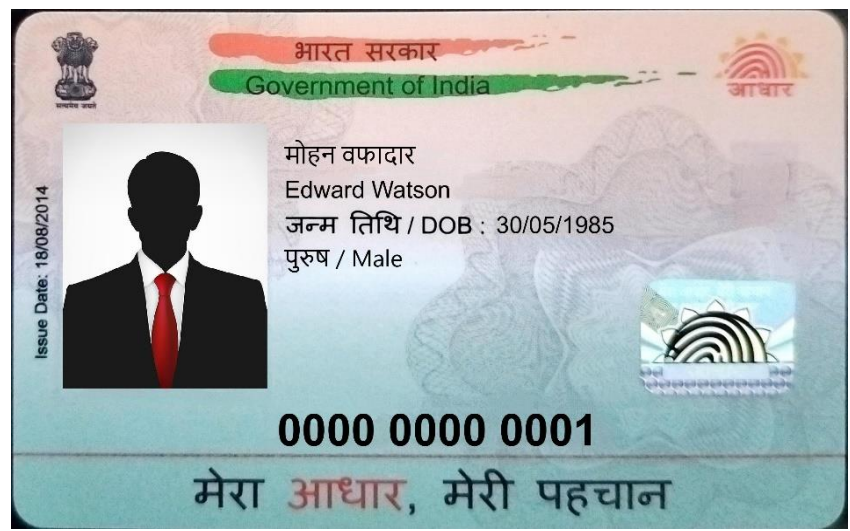

Figure 3.3: Sample Adhaar Card (Base Template)

More than that, certain filters such as Greyscale and Random Localization filters would be applied for creating the most diversely test case images. The Greyscale filter changed an image to black and white format, simulating a scenario where the document may be scanned or photocopied. The Random Localization filter also adjusts parts of the image in several ways with more variation, simulating the situations where content of documents can be placed anywhere on the A4 sheet.

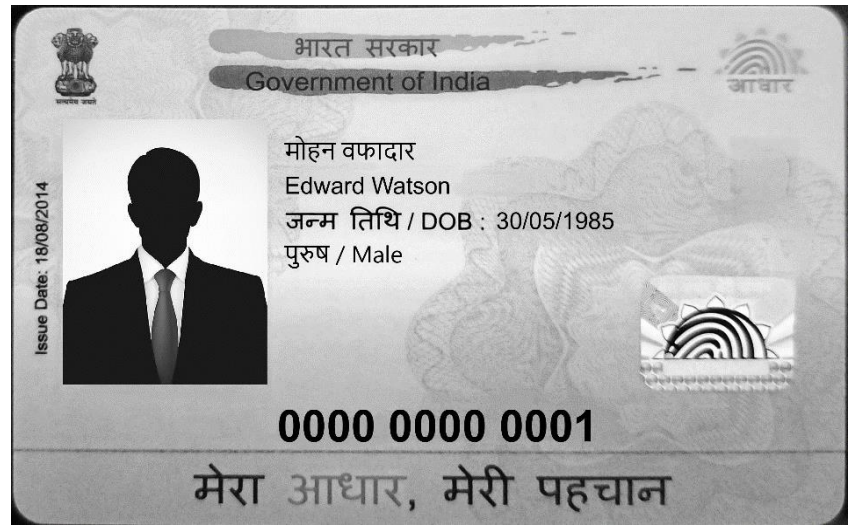

Figure 3.4: Sample Adhaar Card (Greyscale filter)

These methods will produce the images that will make up the dataset used for experimentation and training the model. Through these methods, the dataset will reconstitute real-world variability and image quality sufficient to meet requirements. The following images demonstrate the variety of types of documents, conditions that will be covered in the test cases: An all-inclusive approach will ensure that this document identification and information extraction system is perfectly geared up to its proposed deployment in practical applications.

Since we will be working on 5 different types of identification documents. They are:

a. Adhaar Card

b. Driving Licence

c. PAN Card (Permanent Account Number)

d. Passport

e. Voter Card

Let's have a look how the complete dataset looks like.

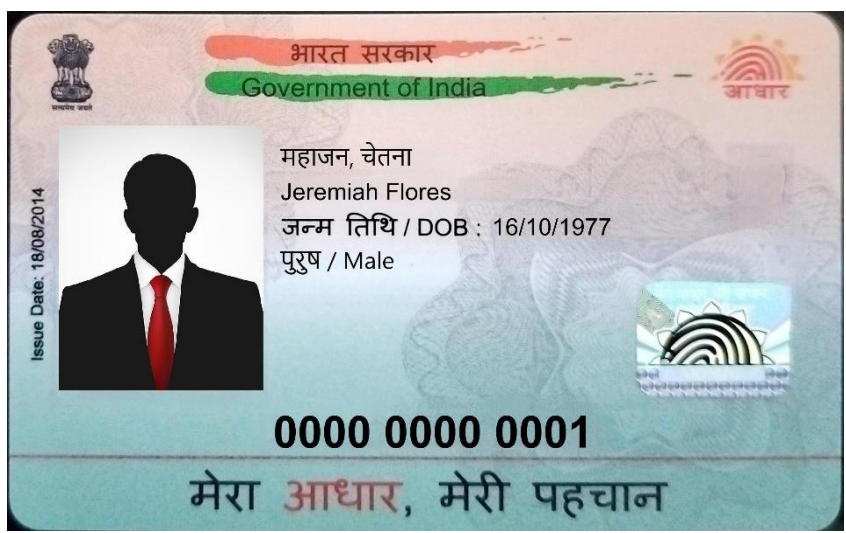

Figure 3.5: Adhaar Card - Base Template

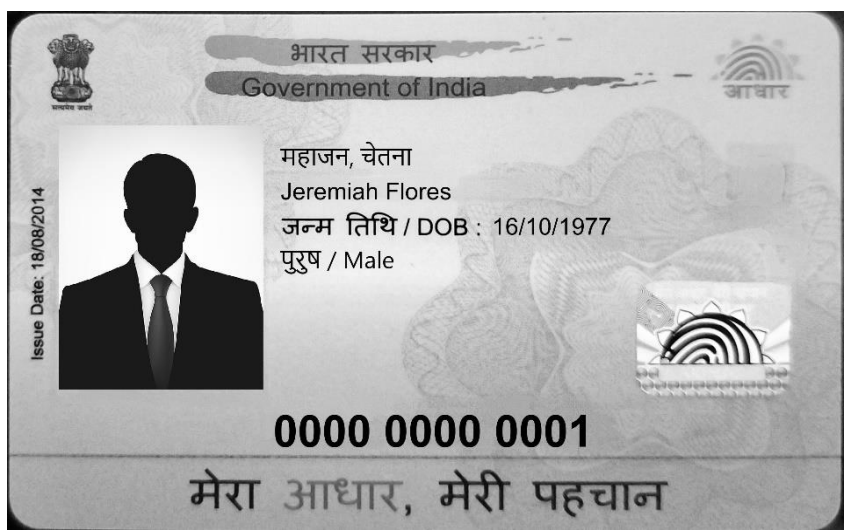

Figure 3.6: Adhaar Card – Greyscale Filter

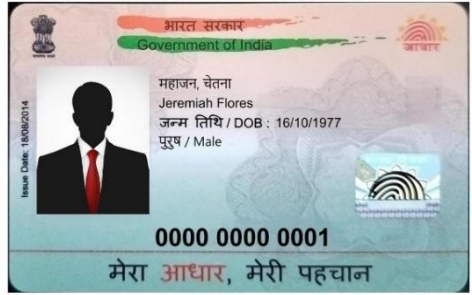

Figure 3.7: Adhaar Card – Randomization Filter on A4 sheet

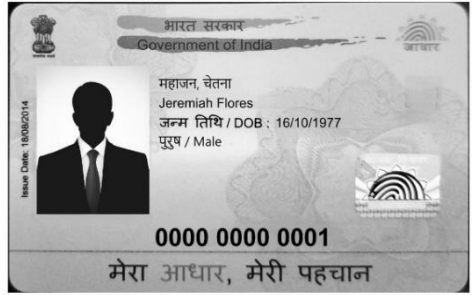

Figure 3.8: Adhaar Card – Greyscale and Randomization Filter on A4 sheet

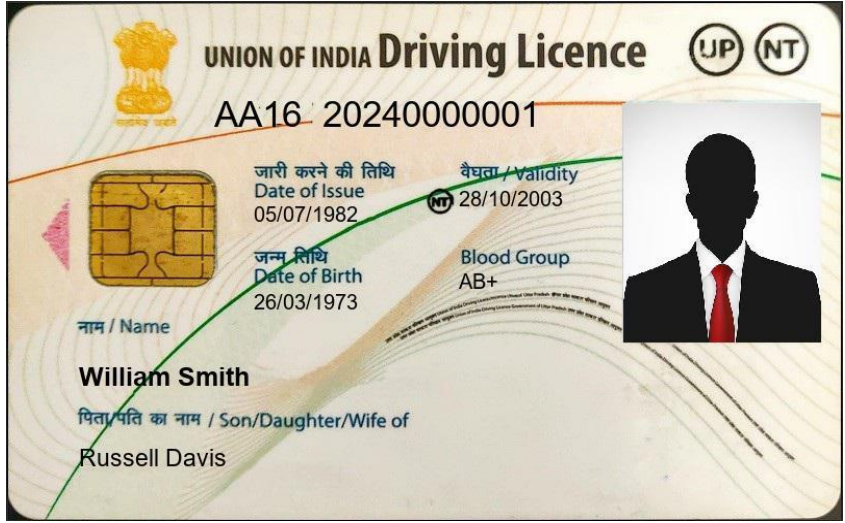
Figure 3.9: Driving Licence - Base Template

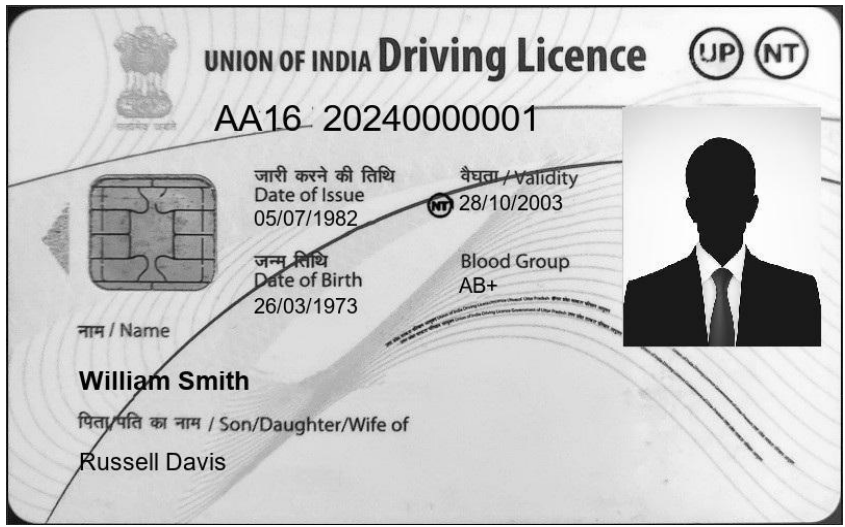
Figure 3.10: Driving Licence – Greyscale Filter

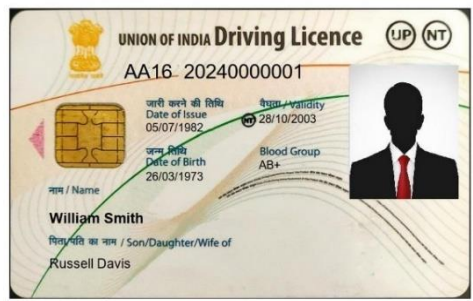

Figure 3.11: Driving Licence – Randomization Filter on A4 sheet

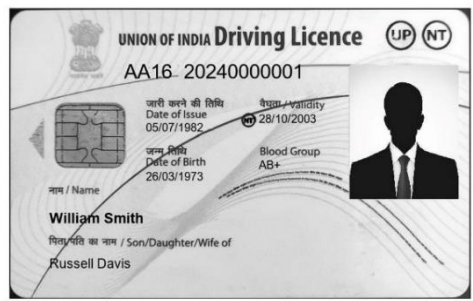

Figure 3.12: Driving Licence – Greyscale and Randomization Filter on A4 sheet

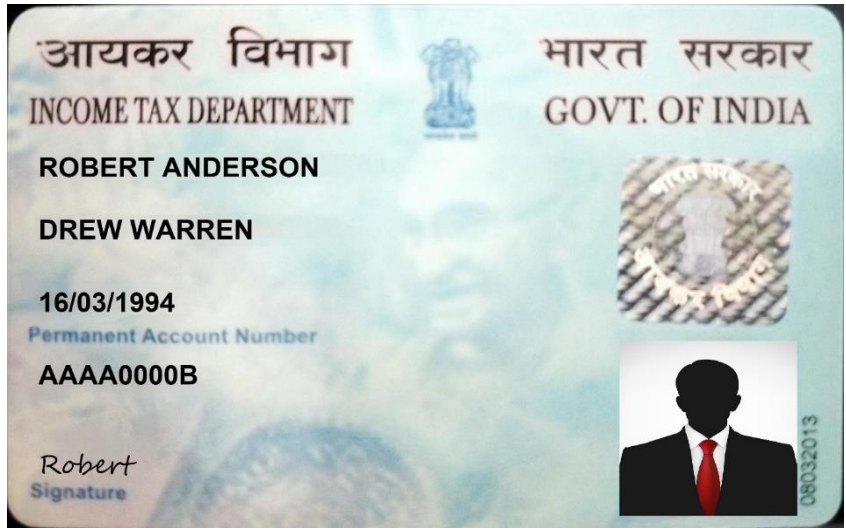

Figure 3.13: PAN Card - Base Template

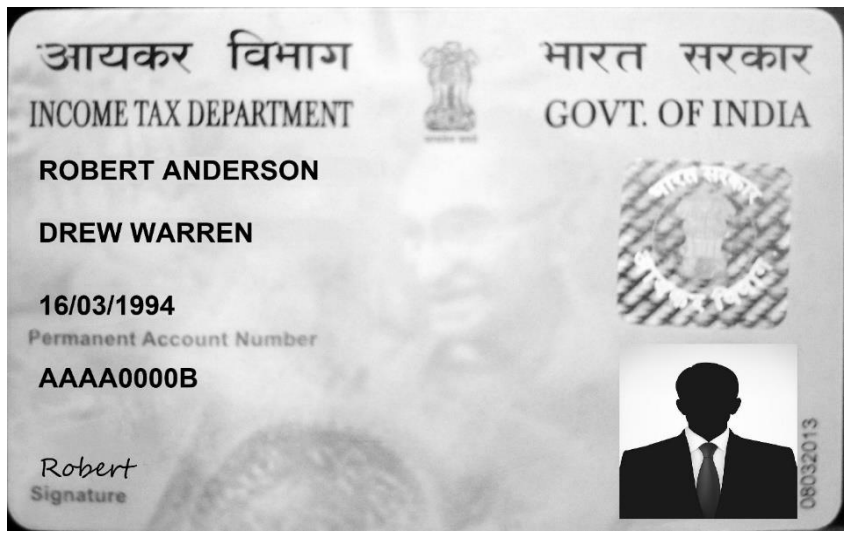

Figure 3.14: PAN Card – Greyscale Filter

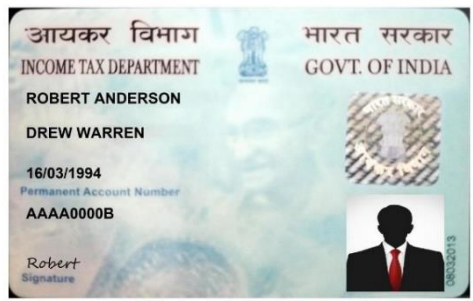

Figure 3.15: PAN Card – Randomization Filter on A4 sheet

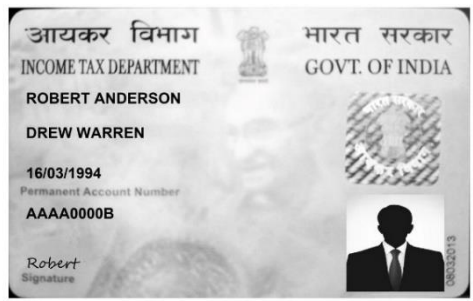

Figure 3.16: PAN Card – Greyscale and Randomization Filter on A4 sheet

Figure 3.17: Passport - Base Template

Figure 3.18: Passport – Greyscale Filter

Figure 3.19: Passport – Randomization Filter on A4 sheet

Figure 3.20: Passport – Greyscale and Randomization Filter on A4 sheet

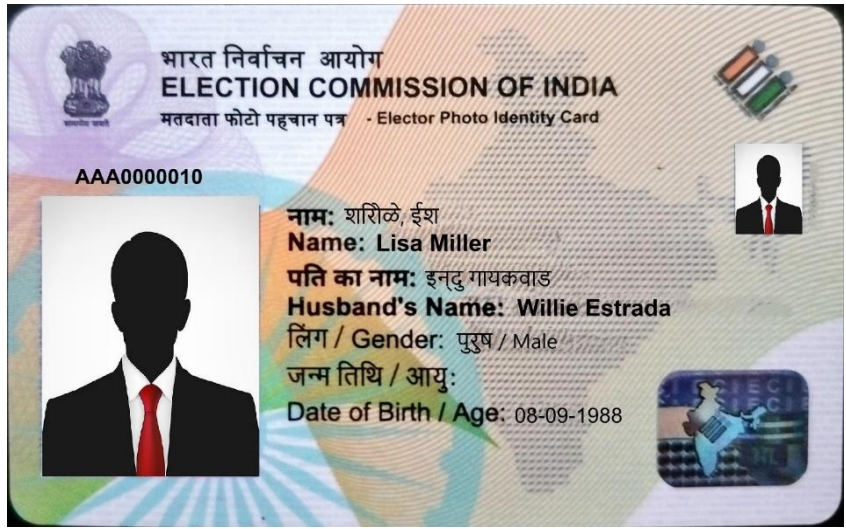

Figure 3.21: Voter Card- Base Template

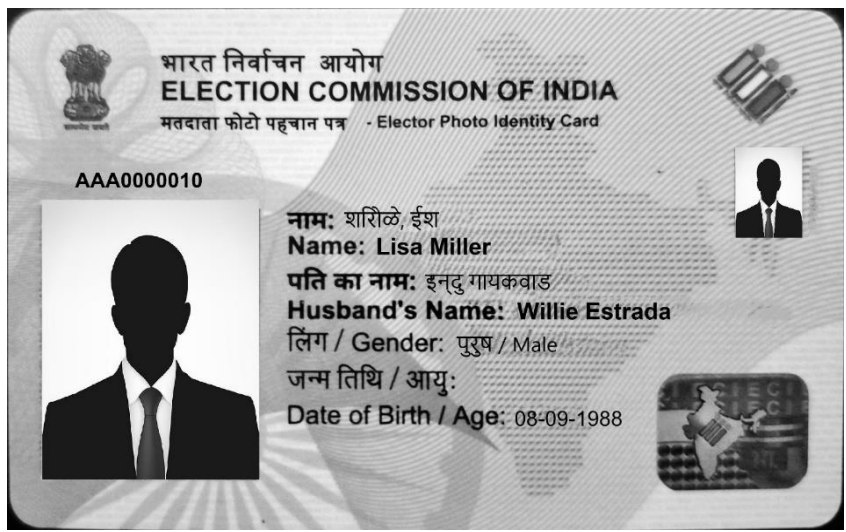

Figure 3.22: Voter Card – Greyscale Filter

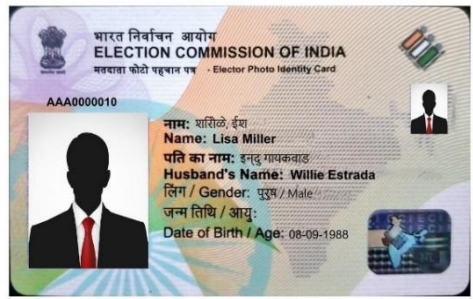

Figure 3.23: Voter Card – Randomization Filter on A4 sheet

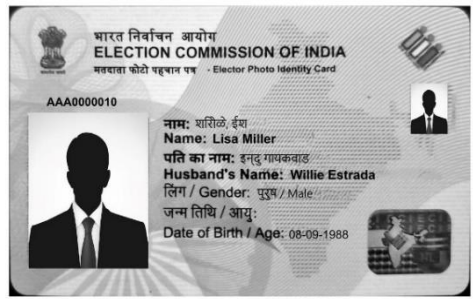

Figure 3.24: Voter Card – Greyscale and Randomization Filter on A4 sheet

2. Documents' Annotation:

Using the Python library 'faker', identification-related data will be generated and embedded into the images and stored in a separate text file. This text file will later be used to verify the information extracted by the model. The annotation format is like following:

a. Adhaar card:

   *Jeremiah Flores::16/10/1977::Male::0000 0000 0001*

b. Driving Licence:

   *AA16          20240000001::05/07/1982::28/10/2003::26/03/1973::AB+::William Smith::Russell Davis*

c. PAN Card:

   *Robert Anderson::Drew Warren::AAAA0000B::16/03/1994*

d. Passport:

   *A0000007::GENTRY::William::28/10/1994::F::CAIRO::CAIRO::19/07/1974::27/12/2002*

e. Voter Card:

   *Lisa Miller::Willie Estrada::AAA0000010::Female::08-09-1988*

Please note that ":::" delimiter is used separate all the data points present in each document. Moreover, the annotation will remain the same for all the premutation of filters for same type of documents.

### 3.3.5 Problem Solutioning

The initial steps of training a Machine Learning model are now done. Problem Statement is well defined. Dataset has been collected or generated in our case. The next step is to identify how to reach the final destination, which is extraction of data from the images. So, following the Process Flow of Augmented Reinforcement Learning Framework, the Data Preprocessing is complete. The Split Dataset is pretty straight forward. Dataset can be divided into two parts: Train Set and Validation Set. The ratio of the two is 4:1.

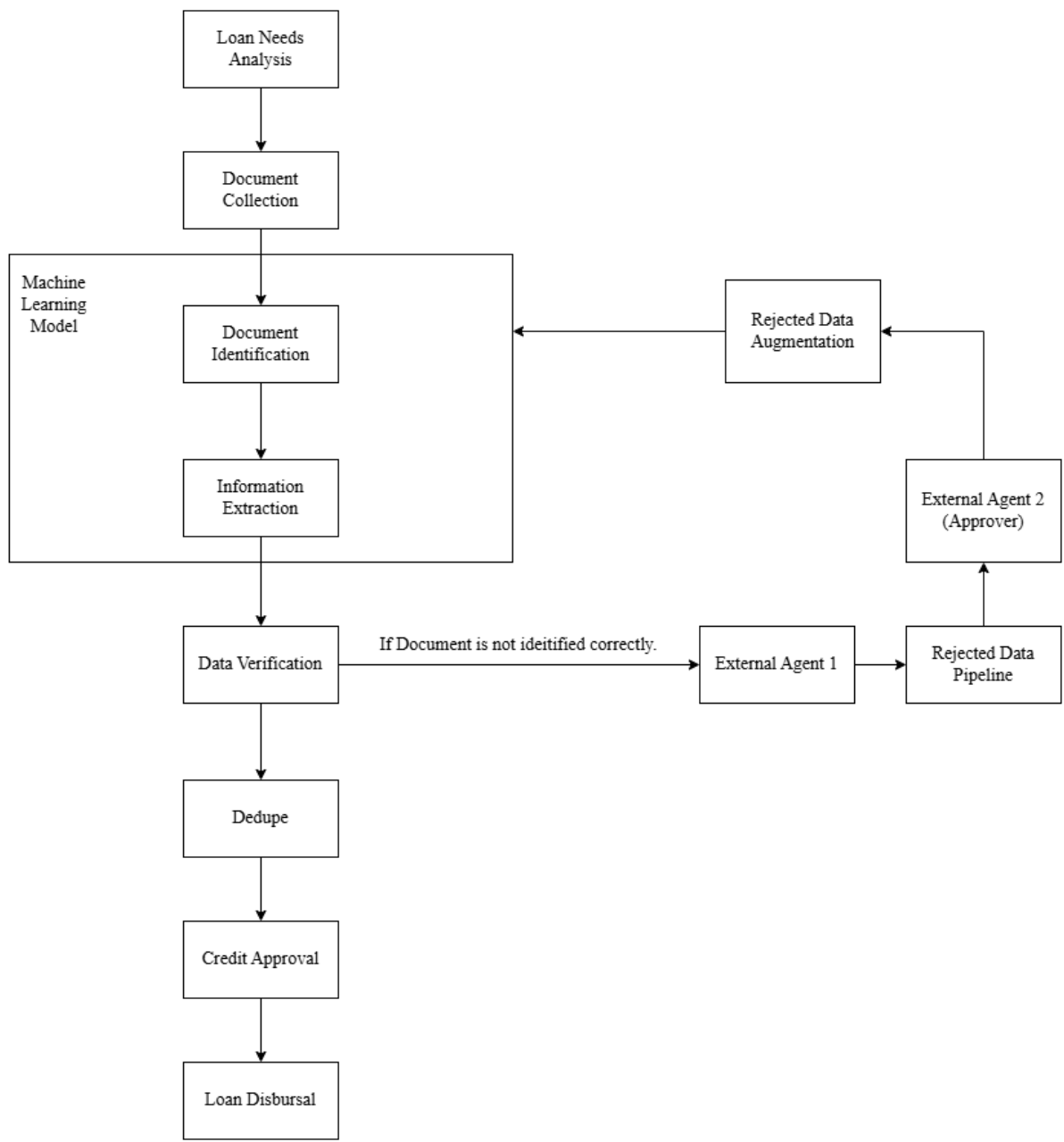

Figure 3.25: Typical Lending System Flowchart incorporating ARL Framework

The Figure 3.25 shows how a lending system works. The stage after document collection the main point of focus for this study. In happy case, there is no need of feedback loop. But if the model is not able to identify the given document, well, in that case, feedback loop is triggered and all the actors come into play. But for now, notice the crux of the problem. Steps in the following sections are needed to achieve the final goal.

### 3.3.6 CNN for Document Classification

The dataset generated in previous section will be used for training. A CNN model will be required to detect content of the image. For this purpose, YOLOv8 will be used. It is the newest member of the YOLO family, famous for its real-time object detection, while still balancing high accuracy with speed, which makes it perfect for any task that requires real-time performance and high detection accuracy. The YOLO architecture comes from one-stage object detectors, which detect objects in a single pass instead of multiple passes as in the case of two-stage detectors like R-CNN, whose first submission of region proposals is made to generate classes and then classify every region proposal into classes.

The core architecture of YOLOv8 can be divided into three main components:

1. Backbone

   The backbone feature extraction of the input image occurs. In YOLOv8, a variant of CSP-Darknet53 is used as backbone. The backbone mainly processes input images through a convolutional layer and residual blocks to achieve hierarchical feature representations at different granularity levels. Early layers obtain low-level features such as edges and textures, whereas deep layers acquire high-level semantic features describing objects.

   YOLOv8 further increases the efficiency of the backbone by using CSP connections that reduce the amount of memory without compromising the high accuracy offered by improved gradient flow in the backward-pass mechanism. With such an ability, YOLOv8 can take up more complex features with higher computing efficiency.

2. Neck

   The neck of YOLOv8 is composed of Feature Pyramid Networks (FPN) and Path Aggregation Networks (PAN). The main task of any neck is to aggregate multi-scale features from different layers of the backbone and contextual information, necessary for object detection involving objects of varying sizes.

FPN ensures it provides the network with a chance to work on features at both fine-grained high resolutions going from earlier layers--fine for small objects--and coarse low-resolution in later layers, useful for larger objects.

PAN further enhances the flow of information by providing a pathway for combing together information from various levels of a network. Combining features along both top-down and bottom-up pathways is essential for better localization and object classification.

3. Head

The final object predictions generated consist of the coordinates of the bounding box, class scores, and confidence scores. The anchor-free detection head is used in YOLOv8, while in all versions of YOLO prior to that, anchor boxes are applied, which basically defines the predefined sizes and shapes of objects. This version of YOLO predicts the centre, width, height, and class of the object while simplifying the network and reducing the complexity of computation.

YOLOv8 outputs a number of feature maps at different scales. Multi-scale detection would be done at different sizes to improve the detection of the object with respect to various sizes.

Machine Learning models created using YOLO detect object using following steps:

1. Input Processing

The input image is resized to a fixed dimension (e.g., 640 by 640) maintaining the aspect ratio; if necessary, padded with zeros. In YOLOv8, an image is processed in a single pass through the network that attempt to extract features while simultaneously doing object detection.

2. Feature Extraction

As the image goes through the backbone, some of the convolutional operations extract the feature maps. Each layer learns different features to these networks; it begins with simple patterns such as edges and texture and so forth to a very complex object shape.

3. Multi-Scale Prediction

YOLOv8 checks the existence of objects at different scales. It uses a neck, which is combined with the differences in resolution from the backbone-to-feature map, and passed forward to the detection head. As such, this multi-scale feature fusion is an important feature in taking note of both small and large objects detected by YOLOv8.

4. Bounding Box Regression and Class Prediction

For each feature map cell, YOLOv8 predicts:

- The centre (x, y) coordinates of the object's bounding box.

- The width and height of the bounding box relative to the grid cell.

- A confidence score indicating the likelihood that the cell contains an object.

- The class score indicating which object class is present.

YOLOv8 applies the sigmoid activation function on bounding box coordinates and objectiveness scores within the valid range (0 to 1) and applies binary cross-entropy loss for class prediction and IoU (Intersection over Union) loss for bounding box regression.

5. Non-Maximum Suppression (NMS)

The predictions coming from this first run go through NMS. NMS eliminates bounding boxes of the same object which overlap with each other by comparing their confidence scores and IoU. Only the most confident box is kept from the other discarded groups.

YOLOv8 passes the input image through its neural network and returns the predictions. The output is bounding boxes, that forms the predictions along with the corresponding confidence values and class index of where the objects in the input image are detected.

### 3.3.7 Templatization of each type of Identification Documents

To understand how data extraction works, it becomes very important to understand the concept of templatization in the context of this problem statement. This basically involves key data points that need to be extracted from a document, and then that are outlined into a pre-defined structure or layout. This would therefore produce an image skeleton or template of the paper, marking certain regions as they pertain to the location of relevant information. Such designated areas are extraction points toward assisting in the extraction process.

The creation of a template serves two key purposes:

1. Noise Reduction

   Applying OCR to the whole image may extract unnecessary or irrelevant data, called noise. However, when OCR is only applied to certain defined regions, the whole process of extraction becomes more precise and targeted, thus reducing the noise much more.

2. Increased Efficiency

   This, by necessity, reduces the processing time over the image. With OCR, it would have to scan from beginning to end rather than finding the right location in which the data is situated. The data extraction process is much more efficient as well with this selective approach.

In the case of different kinds of identity documents, there are varied templates defined based on the layout and structure of different kinds of document types. As an example, a passport template will be different from one of a driver's license because the data points and their placement on the actual document will vary. Template parameters are given regarding location settings like key field name as well as date of birth. It has variations of the layout depending upon the issuer of the document or format.

Templatization actually helps in ensuring that data extraction is both accurate and effective since it will have minimized errors and would speed the process up by considering only the relevant parts of the document.

The template is defined as per different type of identification documents. The different parameter used in defining the template of a document are given in Table 3.1 below.

Table 3.1: Parameters used in the templates and their meaning

| Parameters | Meaning |
|---|---|
| identifying_region | Data that uniquely identifies a document |
| identifying_text | Text that uniquely identifies a region |
| isx | Identified (OCR gives a bounding box as output) Field Starting X Coordinate |
| isy | Identified (OCR gives a bounding box as output) Field Starting Y Coordinate |
| iex | Identified (OCR gives a bounding box as output) Field Ending X Coordinate |
| iey | Identified (OCR gives a bounding box as output) Field Ending Y Coordinate |
| data_regions | List of regions |
| code | Unique Identifier |
| osx | Original Template Starting X coordinate |
| osy | Original Template Starting Y coordinate |
| oex | Original Template Ending X coordinate |
| oey | Original Template Ending Y coordinate |

For visual purpose, the regions of data points are marked in red. The template contents are provided after each image. All the 5 types of document templates are provided below.

In case of Adhaar Card, the template looks like following:

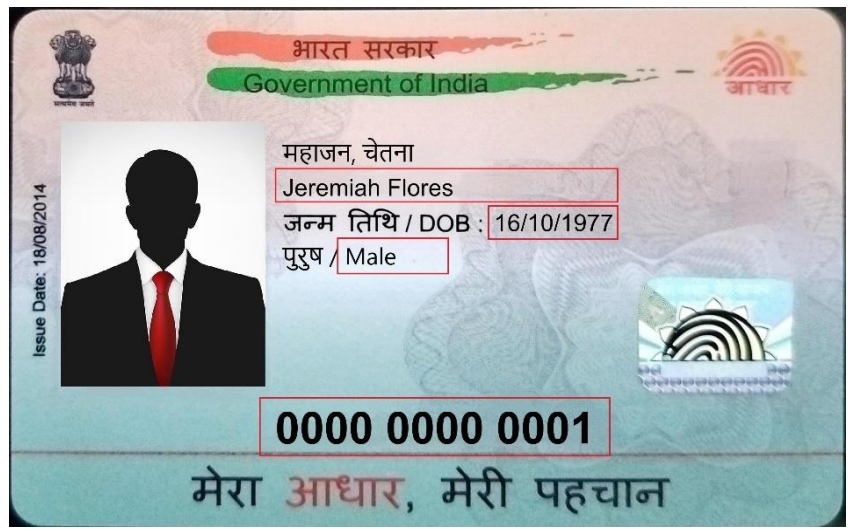

Figure 3.26: Adhaar Card Template

The template configuration for Adhaar Card is as follows:

```
{
    "code": "ADHAAR_V1_P1",
    "identifying_region": {
        "code": "DOCUMENT_IDENTITY_REGION",
        "isx": 786,
        "isy": 215,
        "iex": 1629,
        "iey": 307,
        "osx": 781,
        "osy": 210,
        "oex": 1634,
        "oey": 312,
        "identifying_text": "Government of India"
    },
    "data_regions": [
```

```json
    {
      "code": "NAME",
      "osx": 911,
      "osy": 566,
      "oex": 2105,
      "oey": 681
    },
    {
      "code": "DATE_OF_BIRTH",
      "osx": 1615,
      "osy": 672,
      "oex": 2105,
      "oey": 784
    },
    {
      "code": "GENDER",
      "osx": 1111,
      "osy": 802,
      "oex": 1445,
      "oey": 910
    },
    {
      "code": "ADHAAR_NUMBER",
      "osx": 889,
      "osy": 1355,
      "oex": 1982,
      "oey": 1509
    }
  ]
}
```

Following the same pattern, in case of Driving Licence, the template looks like following:

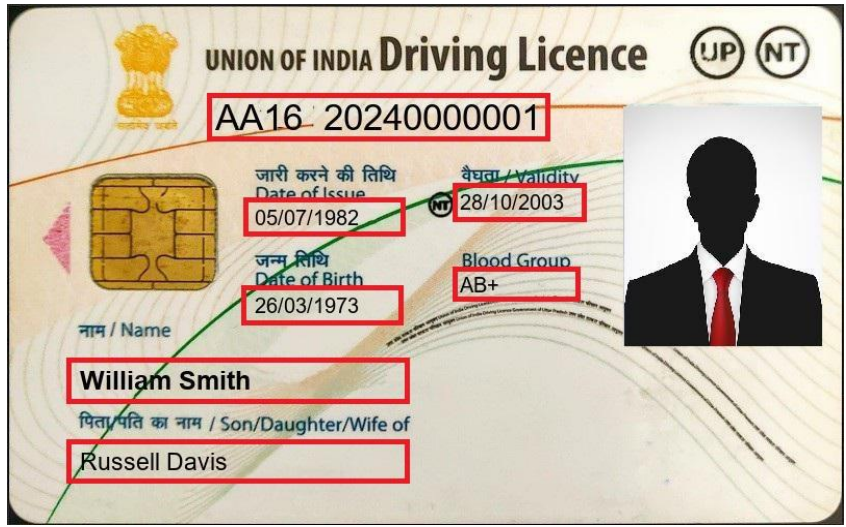

Figure 3.27: Driving Licence Template

The template configuration for Driving Licence is as follows:

```
{
    "code": "DL_V1_P1",
    "identifying_region": {
        "code": "DOCUMENT_IDENTITY_REGION",
        "isx": 351,
        "isy": 23,
        "iex": 623,
        "iey": 83,
        "osx": 346,
        "osy": 18,
        "oex": 628,
        "oey": 88,
        "identifying_text": "Driving Licence"
    },
    "data_regions": [
        {
            "code": "NAME",
            "osx": 63,
```

```
      "osy": 345,
      "oex": 271,
      "oey": 385
    },
    {
      "code": "DATE_OF_BIRTH",
      "osx": 229,
      "osy": 273,
      "oex": 347,
      "oey": 309
    },
    {
      "code": "FATHERS_NAME",
      "osx": 63,
      "osy": 423,
      "oex": 257,
      "oey": 461
    },
    {
      "code": "DRIVING_LICENCE_NUMBER",
      "osx": 194,
      "osy": 83,
      "oex": 519,
      "oey": 134
    },
    {
      "code": "VALADITY_TILL_DATE",
      "osx": 427,
      "osy": 173,
      "oex": 543,
      "oey": 209
    },
    {
      "code": "DATE_OF_ISSUE",
```

```json
      "osx": 229,
      "osy": 189,
      "oex": 347,
      "oey": 223
    },
    {
      "code": "BLOOD_GROUP",
      "osx": 428,
      "osy": 256,
      "oex": 496,
      "oey": 286
    }
  ]
}
```

In case of PAN Card, the template looks like following:

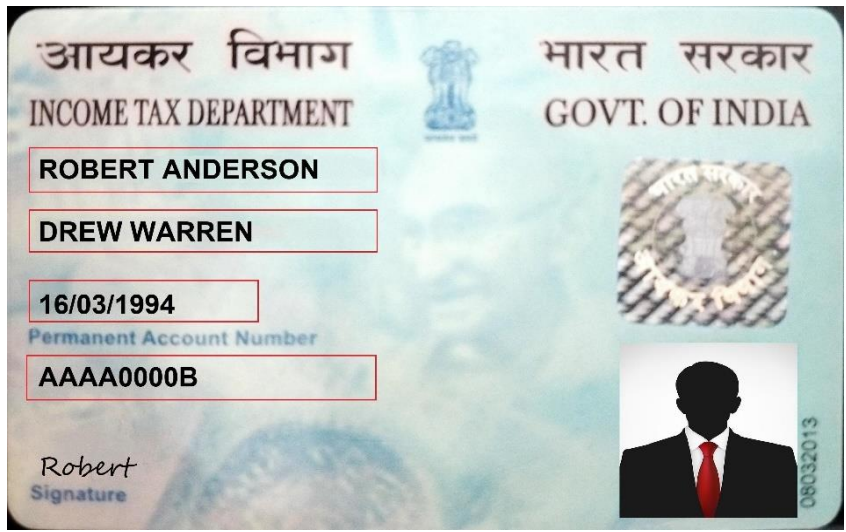

Figure 3.28: PAN Card Template

The template configuration for PAN Card is as follows:

```
{
    "code": "PAN",
    "identifying_region": {
        "code": "DOCUMENT_IDENTITY_REGION",
        "isx": 70,
        "isy": 339,
        "iex": 1341,
        "iey": 480,
        "osx": 65,
        "osy": 334,
        "oex": 1346,
        "oey": 485,
        "identifying_text": "INCOME TAX DEPARTMENT"
    },
    "data_regions": [
        {
            "code": "NAME",
            "osx": 106,
```

```json
      "osy": 572,
      "oex": 1400,
      "oey": 685
    },
    {
      "code": "FATHERS_NAME",
      "osx": 106,
      "osy": 808,
      "oex": 1400,
      "oey": 929
    },
    {
      "code": "PERMANENT_ACCOUNT_NUMBER",
      "osx": 106,
      "osy": 1372,
      "oex": 1400,
      "oey": 1485
    },
    {
      "code": "DATE_OF_BIRTH",
      "osx": 106,
      "osy": 1091,
      "oex": 1400,
      "oey": 1209
    }
  ]
}
```

In case of Passport, the template looks like following:

Figure 3.29: Passport Template

The template configuration for Passport is as follows:

```
{
  "code": "PASSPORT_V1_P1",
  "identifying_region": {
    "code": "DOCUMENT_IDENTITY_REGION",
    "isx": 770,
    "isy": 35,
    "iex": 2025,
    "iey": 116,
    "osx": 765,
    "osy": 30,
    "oex": 2030,
    "oey": 121,
    "identifying_text": "REPUBLIC OF INDIA"
  },
  "data_regions": [
    {
      "code": "TYPE",
      "osx": 990,
```

```json
      "osy": 240,
      "oex": 1133,
      "oey": 291
    },
    {
      "code": "COUNTRY_CODE",
      "osx": 1317,
      "osy": 234,
      "oex": 1430,
      "oey": 295
    },
    {
      "code": "NATIONALITY",
      "osx": 1839,
      "osy": 238,
      "oex": 2044,
      "oey": 302
    },
    {
      "code": "PASSPORT_NUMBER",
      "osx": 2193,
      "osy": 228,
      "oex": 2576,
      "oey": 319
    },
    {
      "code": "SURNAME",
      "osx": 992,
      "osy": 368,
      "oex": 1600,
      "oey": 440
    },
    {
      "code": "GIVEN_NAME",
```

```json
    "osx": 992,
    "osy": 537,
    "oex": 1600,
    "oey": 610
  },
  {
    "code": "DATE_OF_BIRTH",
    "osx": 990,
    "osy": 707,
    "oex": 1312,
    "oey": 781
  },
  {
    "code": "GENDER",
    "osx": 1742,
    "osy": 711,
    "oex": 1793,
    "oey": 773
  },
  {
    "code": "PLACE_OF_BIRTH",
    "osx": 992,
    "osy": 877,
    "oex": 1600,
    "oey": 951
  },
  {
    "code": "PLACE_OF_ISSUE",
    "osx": 992,
    "osy": 1048,
    "oex": 1600,
    "oey": 1121
  },
  {
```

```json
      "code": "DATE_OF_ISSUE",
      "osx": 992,
      "osy": 1216,
      "oex": 1312,
      "oey": 1290
    },
    {
      "code": "DATE_OF_EXPIRY",
      "osx": 2090,
      "osy": 1216,
      "oex": 2408,
      "oey": 1292
    }
  ]
}
```

In case of Voter Card, the template looks like following:

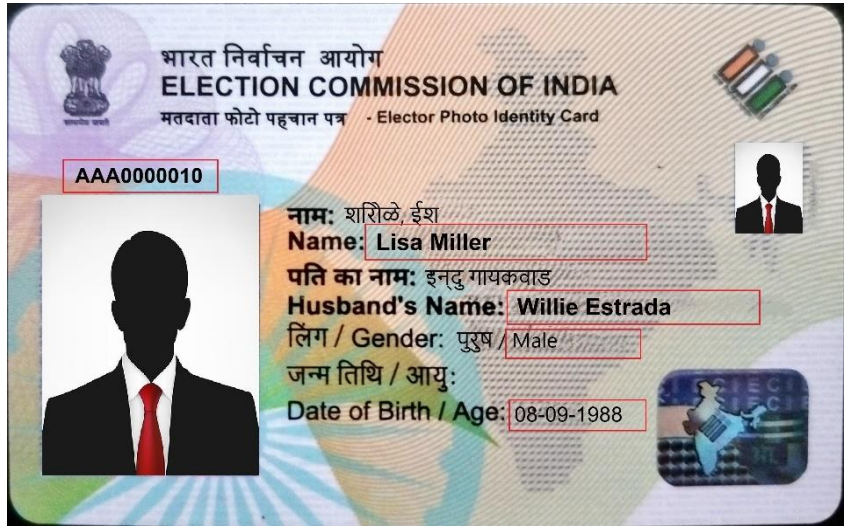

Figure 3.30: Voter Card Template

The template configuration for Voter Card is as follows:

```
{
    "code": "PASSPORT_V1_P1",
    "identifying_region": {
        "code": "DOCUMENT_IDENTITY_REGION",
        "isx": 581,
        "isy": 263,
        "iex": 2344,
        "iey": 374,
        "osx": 576,
        "osy": 258,
        "oex": 2349,
        "oey": 379,
        "identifying_text": "ELECTION COMMISSION OF INDIA"
    },
    "data_regions": [
        {
            "code": "NAME",
            "osx": 1370,
            "osy": 853,
```

```
      "oex": 2500,
      "oey": 992
    },
    {
      "code": "HUSBANDS_NAME",
      "osx": 1940,
      "osy": 1118,
      "oex": 3000,
      "oey": 1219
    },
    {
      "code": "VOTERCARD_NUMBER",
      "osx": 253,
      "osy": 615,
      "oex": 754,
      "oey": 712
    },
    {
      "code": "GENDER",
      "osx": 1910,
      "osy": 1214,
      "oex": 2123,
      "oey": 1363
    },
    {
      "code": "DATE_OF_BIRTH",
      "osx": 1925,
      "osy": 1535,
      "oex": 2367,
      "oey": 1631
    }
  ]
}
```

### 3.3.8 Model for Text Extraction Using Templates

The templates defined will be used along with a ML model in order to perform OCR (Optical Character Recognition) to correctly extract text from the images. For this purpose, EasyOCR will be used. This is an open-source OCR tool based on deep learning that can recognize text within images and videos with the support of any language. It has a combination of object detection and text recognition architecture, thus very efficient for all OCR-related works. It uses a two-stage pipeline: first, its text detection model identifies the regions inside an image that possibly contain text, and next, its text recognition model decodes the detected text.

The detection is primarily done through an East-based model or any architecture similar to it. The model takes the input image and gives it back as a probability map predicting the text existence and a geometry map for the bounding boxes around the detected text.

1. Input Processing

   The preprocessing in input images includes resampling and normalization for optimal performance.

2. Feature Extraction

   In a Convolutional Layer, features of the image are extracted. All the relevant information is extracted at this stage.

3. Output

   This model will give bounding boxes along with potential text regions and their confidence scores, which then goes through Non-Maximum Suppression (NMS) to eliminate any overlapping detections.

Once the text regions are identified, the model processes those text regions mainly by using CRNN for the recognition step. After the bounding box is detected, the region of that box is cropped from the original image. The architecture uses a fusion of CRNN to combine CNN for the feature extraction and RNN for sequence learning to decode sequences of characters. Output with Confidence Scores Alongside the recognized text. EasyOCR is trained on large datasets to learn all different fonts and styles; at inference time, this model takes in input images and uses the pipeline of detection and recognition to output recognized text.

Now, everything is ready, except one last ingredient. At this stage, image has been identified against a particular document type. Templates have been defined. Now, only thing that remains is in which region of the input image, OCR is required in order to extract data. For this we need to perform a mathematical calculation.

Let's define few things. Recall that in previous section, we defined **identifying_region** parameter for each of the templates. That data will be used in data extraction region in the input image.

A typical template is defined as follows:

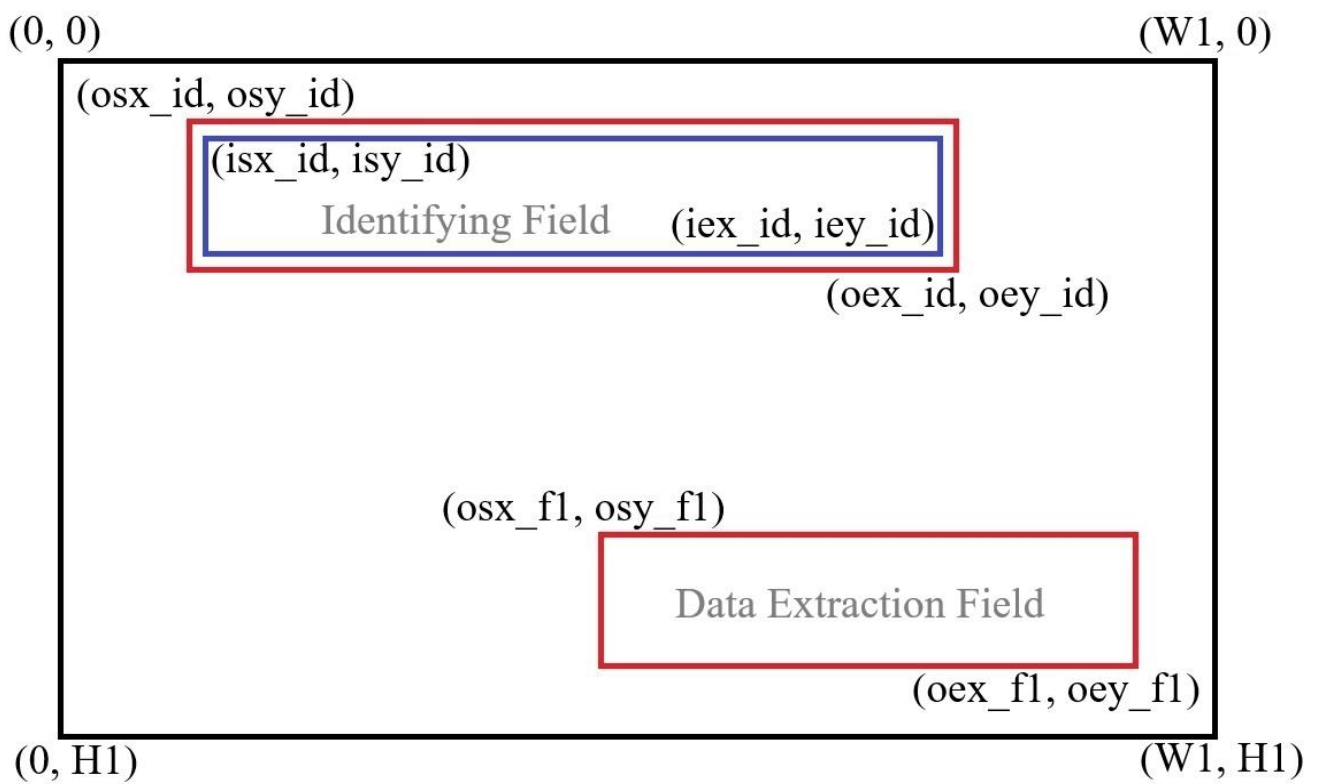

Figure 3.31: Typical Document Template

*Note: All the coordinates in the image above are already known from the template.*

Without loss of generality, the following image represents a typical image input to the system containing the identification document.

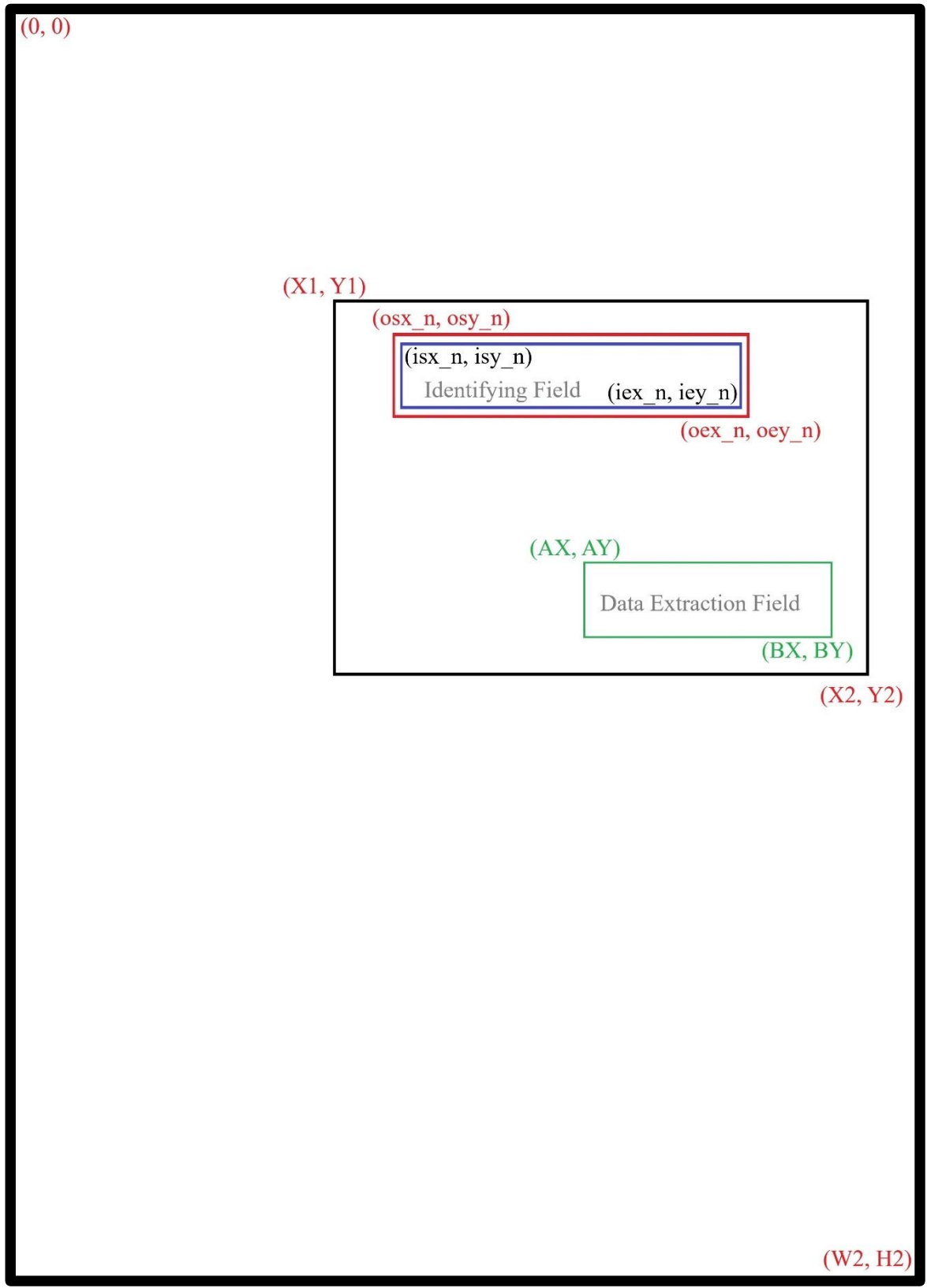

Figure 3.32: Typical Document Input Image

*Note: Only the coordinates on black colour are known as OCR is performed initially while defining templates. The coordinates in red are also unknown. The coordinates in green colour are required for the solution i.e. coordinates named AX, AY, BX, and BY.*

There is an assumption here, that the image is not distorted while scanning. And this is true, in real world scenario more than 90% of the time. Sophisticated copying and scanning machines, mobile apps using image capture via cameras make it possible.

Since the distortion is absent, the ratio of the bounding rectangle thus formed will remain the same. Hence, the Similar Triangles Theorem can safely be applied here as follows:

$$\frac{(iex\_id - isx\_id)}{(osx\_f1 - isx\_id)} = \frac{(iex\_n - isx\_n)}{(AX - isx\_n)}$$

$$(AX - isx\_n) = \frac{(iex\_n - isx\_n)(osx\_f1 - isx\_id)}{(iex\_id - isx\_id)}$$

$$\boxed{AX = isx\_n \ + \ \frac{(iex\_n - isx\_n)(osx\_f1 - isx\_id)}{(iex\_id - isx\_id)}}$$

Similarly for AY,

$$\frac{(iey\_id - isy\_id)}{(osy\_f1 - isy\_id)} = \frac{(iey\_n - isy\_n)}{(AY - isy\_n)}$$

$$(AY - isy\_n) = \frac{(iey\_n - isy\_n)(osy\_f1 - isy\_id)}{(iey\_id - isy\_id)}$$

$$\boxed{AY = isy\_n \ + \ \frac{(iey\_n - isy\_n)(osy\_f1 - isy\_id)}{(iey\_id - isy\_id)}}$$

For BX,

$$\frac{(iex\_id - isx\_id)}{(oex\_f1 - isx\_id)} = \frac{(iex\_n - isx\_n)}{(BX - isx\_n)}$$

$$(BX - isx\_n) = \frac{(iex\_n - isx\_n)(oex\_f1 - isx\_id)}{(iex\_id - isx\_id)}$$

$$\boxed{BX = isx\_n \ + \ \frac{(iex\_n - isx\_n)(oex\_f1 - isx\_id)}{(iex\_id - isx\_id)}}$$

And for BY,

$$\frac{(iey\_id - isy\_id)}{(oey\_f1 - isy\_id)} = \frac{(iey\_n - isy\_n)}{(BY - isy\_n)}$$

$$(BY - isy\_n) = \frac{(iey\_n - isy\_n)(oey\_f1 - isy\_id)}{(iey\_id - isy\_id)}$$

$$\boxed{BY = isy\_n \ + \ \frac{(iey\_n - isy\_n)(oey\_f1 - isy\_id)}{(iey\_id - isy\_id)}}$$

Now for different extraction fields, the equations derived above provide the extraction region.

### 3.3.9 Implementation of ARL framework

Now comes the interesting part of the solution: understanding how this feedback loop works on the problem statement. In the preceding sections, the Rejected Data Pipeline was introduced; now, let's activate it for future cycles of training. This is where the Augmented Reinforcement Learning Framework comes into play. Let's walk through this modified process flow step-by-step to see just how this feedback mechanism works.

The core aspect of the ARL framework is the dynamic feedback loop, which is designed to recover errors and reinforce the learning mechanism of the machine learning model. This loop works after the model encounters failure cases, such as in this problem statement, misclassifying documents. The phase of the feedback loop is divided into steps that have driving agents which monitor and refine the feed data given back to the model.

1. **Role of External Agent 1: Filtering and Identifying the Rejected Data**

   The first component of the feedback loop is External Agent 1 that would eliminate and mark all of those negative outcomes from the model's predictions. Here, negative outcomes refer to cases where a model's decision-making mechanisms give wrong outputs or less desirable outputs. An example given would be the following:

   - When a poor illumination or a skewed orientation of an image leads to the model not classifying it correctly, then such an outcome is tagged as such.

   - In the same line of argument, if the model fails to extract a particular critical information from a document, such as the specific field within a bank statement, then it marks that particular case for review.

   The External Agent 1 sifts through this wrong forecast systematically and places them in a repository called the Rejected Data Pipeline. The pipeline, thus, is a form of a cache for problematic data, acting as the base for further analysis and retraining.

   It isolates negative data, thus passing only relevant and actionable feedback forward. This phase is necessary to keep the focus of the learning process directed toward improving the weak areas of the model rather than letting unwanted noise dilute the feedback loop.

2. **Role of External Agent 2: Scenario Validation and Data Segregation**

Once a Rejected Data Pipeline becomes saturated with bad results, External Agent 2 steps into the scene. Classification and segregation of rejected data through invalid business scenarios or scenarios into a decision framework will take place. Here, the business context for lending is concerned.

- Scenario Validation: The second external agent checks each occurrence of bad data to check if the case represents a valid scenario for retraining. In this way, only critical errors from the business get targeted to correct in the following iterations. For example,

    - Cases are labeled as infeasible when the system is unable to process the de-skewed images and considered irrelevant to the business process.

    - It means, on the other hand, that if the system fails to identify a document type that is critical for the lending process, such as a new identification form, or the bank statement format, that is considered a valid case and needs to be addressed separately.

- Data Separation: Once all the valid scenarios are defined, they are segregated from irrelevant ones. Irrelevant data is removed so that the model does not overfit or learn unnecessary complexities. The valid scenarios are ensured to be available for retraining; thus, the feedback loop is maintained in line with real business needs.

In this stage, Agent 2 introduces a contextual layer of intelligence in the feedback loop so that the framework would be focused on solving real problems, not theoretical ones.

3. **Data Augmentation for Future Training**

After the External Agent 2 has validated and segregated the data, the next step is data augmentation, which is certainly required to enrich the training set. Data augmentation applies transformations to the rejected data in order to make the dataset more diverse and more representative. It gives the possibility to hone the model in terms of generalization in real life.

- Applied Transformations: Some common transformations are brightness adjustments, rotation, greyscale, noise addition, or cropping. These

transformations of images can also be used to mimic reality by low resolution of the image, poor illumination, or for skewing documents.

- Case Example: Given a low-resolution image, the model fails to detect a given document type. The model is allowed to produce different variants of the document by means of brightness enhancement, the addition of noise, and adjusting the sharpness. Variants allow the model to have an enriched dataset from which it can learn from. This way, it would better understand similar cases in the future.

Data augmentation tends to enhance the rejected dataset and, in the process, provides the feedback loop with new challenges to face the model at each round. This process enhances the abilities of the model to decide by exposing it to a much larger set of scenarios.

## 4. Continuous Feedback Loop for Improvement

Continuous feedback loop: last step of ARL. It's an iterative loop ensuring on ongoing model refinement; it's a cycle in which the model goes through each cycle, acquiring new data, generates predictions, suffers from failures, and feeds back these failures into the system with the help of External Agent 1 and External Agent 2 to complete the loop.

- Iteration Cycle: During every iteration, the first external agent filters out the negative outcomes while the second specifies them in relevant scenarios. These relevant scenarios are then augmented and reintegrated into the training process, hence learning its mistakes.

- Adaptive Learning: As iteration goes through, the model tends to be tougher on mistakes and much more effective at dealing with complex and dynamic scenarios. An adaptive learning mechanism will ensure that the model changes in response to actual issues it faces in the world.

The continuance of this feedback loop is one major strength in the ARL framework. It enables the model to stay updated and ready according to the dynamic requirements of the business environment through the systematic incorporation of new scenarios into the training cycle.

The ARL framework offers many advantages that makes it effective for solving real-world problems such as document identification and information extraction:

- Error Reduction: The framework eliminates specific failure cases, thereby inhibiting recurring errors.

- Business Relevance: External Agent 2 removes all likelihood of feedback loops engaging in business-critical scenarios since it minimizes unnecessary complexity.

- Generalization Capability: Data augmentation enhances the ability of the model to handle variations, thereby becoming more robust in diverse scenarios.

- Scalability: The feedback loop must run continuously; hence, it ensures that in the long run, the model learns to tackle new challenges.

The Augmented Reinforcement Learning Framework is a powerful and adaptive way to increment the machine learning models, as its idea is based on how it triggers an external mechanism that involves filtering, validation, and augmentation with a productive feedback loop in which it continuously refines the performance of a model.

This iterative process allows the model to adapt over complex and dynamic scenarios; thus, it makes it particularly well-suited for real-world applications such as document identification and information extraction in the lending domain. The ARL framework enhances decision-making capabilities and provides an easy-to-scale and efficient solution for addressing the limitations of traditional reinforcement learning.

Further refinements in this framework mean it has far-reaching potential beyond the scope of the problem statement, thus providing a very general framework for resolution of various problems in machine learning and artificial intelligence.

### 3.3.10 Evaluation

The annotation made in the previous section will be used to test model accuracy. These annotations, created using the Python library "faker", contain identification-related data embedded into the images and store them in separate text files. These text files are ground truths to measure the performance of the model.

The evaluation begins by feeding the augmented dataset into the model. It comprises generated images of varied sizes and densities, filters like Greyscale, and Random Localization. The model processes the images to extract embedded information. Extracted information is then compared against ground-truth annotations.

Accuracy in this respect will be measured by the accuracy with which it extracts information from the images. Among others, key metrics to use include Precision, Recall, and F1-score, which collectively provide a more complete view of model performance.

The precision measures the correctness of the extracted data. It is the ratio of true positive (TP) results to the total number of positive results (TP + FP), where FP represents false positives (Jain, 2023). The formula is:

$$Precision = \frac{TP}{TP + FP}$$

The Recall measures the model's ability to capture all relevant information. It is defined as the ratio of true positive (TP) results to the total number of actual positives (TP + FN), where FN represents false negatives (Jain, 2023). The formula is:

$$Recall = \frac{TP}{TP + FN}$$

The combination of the precision and the recall formulates the F1 score. The F1 score provides a balanced measure of precision and recall. It is the harmonic mean of precision and recall, ensuring a comprehensive performance indicator (Jain, 2023). The formula is:

$$F1\ score = \frac{2 \cdot Precision \cdot Recall}{Precision + Recall}$$

Additionally, the evaluation considers the model's performance across different image qualities and augmentations, ensuring it is robust and reliable in various real-world scenarios.

## 3.4    Summary

In this chapter, the ARL Framework is defined that makes it possible to elevate the machine learning capabilities in dealing with complexities and the intricacies as found in real-world problems: such as Document Identification and Information Extraction. It lays the foundation for refining models with a little help from human expertise and business-specific inputs using External Agents, the Rejected Data Pipeline, and business screening. This optimized flow process, now tested on the problem statement of identifying and extracting document, shows how the ARL framework can correct the shortcomings of current machine learning models and improve over time.

Traditionally, the flow of classical models is gathered data, model training, validation, and ready-to-deploy models for real-time applications. Classical models are significantly data-driven, and only when labeled datasets are used, can they be trained and apply the extracted features to make predictions on unseen data by classifying or recognizing patterned data. However, such models are rarely applied to real-world problems such as document identification or data extraction due to complex and diverse natures of documents. Issues in such models are liable to classify or extract wrong due to the presence of document skewness, image quality, and different document formats. These errors are rarely reprocessed in the original machine learning pipeline, hence prone to inefficiency and losses in accuracy over time.

This problem is known as Document Identification and Information Extraction. It refers to identifying relevant information from all kinds of document types such as identity cards, banking records or vehicle registration documents. All these types of documents normally come in various formats with different image qualities and different layout, making it difficult to generalize for a machine learning model. In a way, it actually involves determining the nature of the document and also finding and extracting some details as critical for lending applications or in legal verifications, such as names, addresses, and document numbers.

One of the major challenges in solving this problem is gathering relevant data. Personal documents carry private information, and the level of regulations governing these is quite high, so very few datasets from real sources are easily available. Most organizations will not take up the task of sharing such datasets since confidentiality prevails. However, synthetic datasets have to be used generally for training the model. These datasets are created by generating templates of common document types and augmenting them with variations like skewing, cropping, brightness alteration, or changing resolution. This is a way to simulate real-world conditions and prepare the model for handling documents in various forms that real-life applications might present.

The agents external and the continuous feedback loop for improvement of models significantly deviate the traditional process flow in machine learning with the ARL framework. These external agents identify the negative results filtered by these agents. In case a model does not correctly classify or extract data from a document, there is the Rejected Data Pipeline. External Agent 1 gathers these negative results and forwards them for processing in the Rejected Data Pipeline. From there, External Agent 2 analyses whether rejected scenarios are valid based on that specific business context. For example, if the model fails to process de-skewed images and it is marked as being irrelevant to the business context (say, for instance, would not directly impact a lending decision), then that scenario is rejected. But if failure is an actual business scenario- for example, a user does not identify a new form of document-then the data will be accepted, augmented, and fed into the next training cycle.

The ARL framework relies on a continuous feedback loop, in which rejected data undergo systematic evaluation and are then transferred back into training. After the appearance of valid scenarios from such rejected data, images undergo a variety of augmentations possibly with using filters, brightness adjustment or simulation of various conditions of a concrete image. Those augmented images therefore are input for updating the training dataset of the model. The ARL framework involves a model that continues to learn and improve upon its mistakes and new scenarios; hence, the model becomes more robust and able to handle previously unseen document types and formats with an increased accuracy in identification and extraction of data from documents.

For instance, the idea of bringing in outside players, the spurned data pipeline, and the ARL framework permits always improvement of machine learning models, particularly in hard-to-handle problems in the real world, like that of document identification and extracting relevant data. This architecture would negate the flaws inherent in classic models and could work out the complicated sceneries as such models could learn over time, thus making the application very versatile to be used in other business-sensitive applications, like lending.

# CHAPTER 4: IMPLEMENTATION

## 4.1     Introduction

The significance of Augmented Reinforcement Learning framework can be proved only after implementing it in real world problem. The problem chosen for this purpose is Document Identification and Text Extraction. This chapter provides detailed steps for creating the dataset. Dataset cannot be collected due to the reasons described in previous chapter (Chapter 3). After dataset creation, the details of pre-processing of dataset are provided in subsequent sections. In order to provide comparison, default Machine Learning procedure will be applied. Then, the Augmented Reinforcement Learning framework will be implemented to get the updated and enhanced working Machine Learning model.

## 4.2     Dataset Collection

### 4.2.1     Dataset Creation

In order to create the dataset, a real-world document is photographed. The photograph is then edited in order to remove all the personal data/text from it, making it a placeholder for the document synthesizer tool. The real-world Adhaar card looks like Figure 4.1.

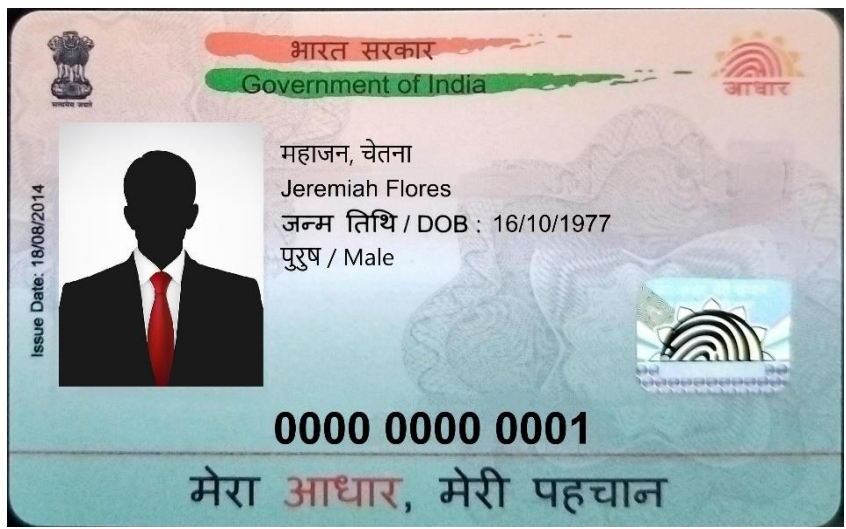

Figure 4.1: Adhaar Card – Actual Document Look-a-like

Then, the above photograph can be edited to remove the text and keep the background intact, to make to as close as to real world Adhaar card document. The editing can be done in photoshop or Google's Photos Magic Eraser can be used. The edited photograph is shown in Figure 4.2.

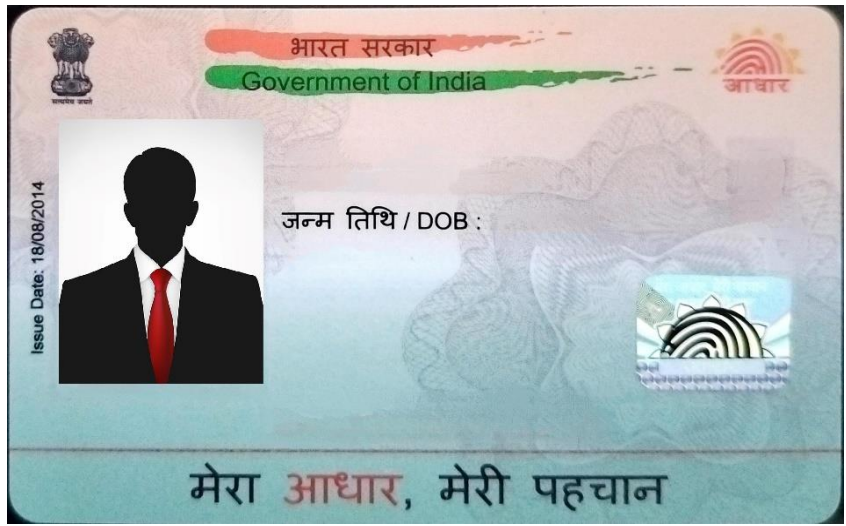

Figure 4.2: Edited Adhaar Card – Base Template

Now, the base template is ready. We need a synthesizer to generate as many documents as we need for our dataset. It is to be noted that adequate font files are used to match the documents' font, to make it more realistic. The code for synthesizer is as follows:

```python
from faker import Faker
from PIL import Image, ImageDraw, ImageFont
import os
import sys

class AdhaarData:
    name_hi: ""
    name_en: ""
    dob: ""
    gender_hi: ""
```

```python
        gender_en: ""
        adhaar_value: ""

faker = Faker()
faker_hi = Faker("hi_IN")
font = ImageFont.truetype("arial.ttf", 80)
font_2 = ImageFont.truetype("arial.ttf", 70)
font_val = ImageFont.truetype("arialbd.ttf", 150)
font_hi = ImageFont.truetype("C:/Users/meets/OneDrive/Desktop/MS Thesis
Work/Code/Nirmala.ttf", 80)
font_hi_2 = ImageFont.truetype("C:/Users/meets/OneDrive/Desktop/MS Thesis
Work/Code/Nirmala.ttf", 70)
adhaar_count = 1
args = sys.argv
max_adhaar = int(args[1])

def progress(count, total, suffix=''):
    bar_len = 60
    filled_len = int(round(bar_len * count / float(total)))
    percents = round(100.0 * count / float(total), 1)
    bar = '█' * filled_len + '-' * (bar_len - filled_len)
    sys.stdout.write('[%s] %s/%s ...%s\r' % (bar, str(count), str(total), suffix))
    sys.stdout.flush()

def createDatasetFolders():
    newpath = r'../generated_documents'
    if not os.path.exists(newpath):
        os.makedirs(newpath)
    newpath = r'../generated_documents/images'
    if not os.path.exists(newpath):
        os.makedirs(newpath)
```

```python
    newpath = r'../generated_documents/images/adhaar_v1_p1'
    if not os.path.exists(newpath):
        os.makedirs(newpath)
    newpath = r'../generated_documents/metadata'
    if not os.path.exists(newpath):
        os.makedirs(newpath)
    newpath = r'../generated_documents/metadata/adhaar_v1_p1'
    if not os.path.exists(newpath):
        os.makedirs(newpath)

def get_adhaar_value(temp):
    value = ""
    for i in range(12):
        a = temp % 10
        value = str(int(a)) + value
        temp /= 10
        if i == 3 or i == 7:
            value = " " + value
    return value

def get_adhaar_metadata(temp):
    adhaar = AdhaarData()
    adhaar.name_hi = faker_hi.name()
    adhaar.name_en = faker.name()
    if (temp <= max_adhaar * 0.4) or (max_adhaar * 0.8 < temp <= max_adhaar * 0.9):
        adhaar.gender_hi = "पुरुष"
        adhaar.gender_en = "Male"
    else:
        adhaar.gender_hi = "महिला"
        adhaar.gender_en = "Female"
    date = faker.date().split("-")
```

```python
    adhaar.dob = date[2] + "/" + date[1] + "/" + date[0]
    adhaar.adhaar_value = get_adhaar_value(temp)
    return adhaar

createDatasetFolders()
for i in range(max_adhaar):
    adhaar = get_adhaar_metadata(adhaar_count)
    # P1
    image = Image.open("../base_documents/ADHAAR_V1_P1.jpg")
    draw = ImageDraw.Draw(image)
    draw.text((925, 440), adhaar.name_hi, font=font_hi, fill=(0, 0, 0))
    draw.text((925, 570), adhaar.name_en, font=font, fill=(0, 0, 0))
    draw.text((1640, 690), adhaar.dob, font=font, fill=(0, 0, 0))
    draw.text((925, 790), adhaar.gender_hi + " / " + adhaar.gender_en, font=font_hi, fill=(0, 0, 0))
    draw.text((900, 1350), adhaar.adhaar_value, font=font_val, fill=(0, 0, 0))
    image.save("../generated_documents/images/adhaar_v1_p1/adhaar_v1_p1_" + str(adhaar_count) + ".jpg")
    file1 = open("../generated_documents/metadata/adhaar_v1_p1/adhaar_v1_p1_" + str(adhaar_count) + ".txt", "w")
    L = [adhaar.name_en + "::" + adhaar.dob + "::" + adhaar.gender_en + "::" + adhaar.adhaar_value]
    file1.writelines(L)
    file1.close()
    progress(i + 1, max_adhaar, "Adhaar Card Generated")
    adhaar_count += 1
```

The above code is saved in a file named "*generate_document_adhaar.py*". In order to generate 10 different documents of Adhaar card, following command is required to do so:

*python generate_document_adhaar.py 10*

The output of the above command is shown in Figure 4.3 and metadata (annotation of each document) thus generated is shown in Figure 4.4.

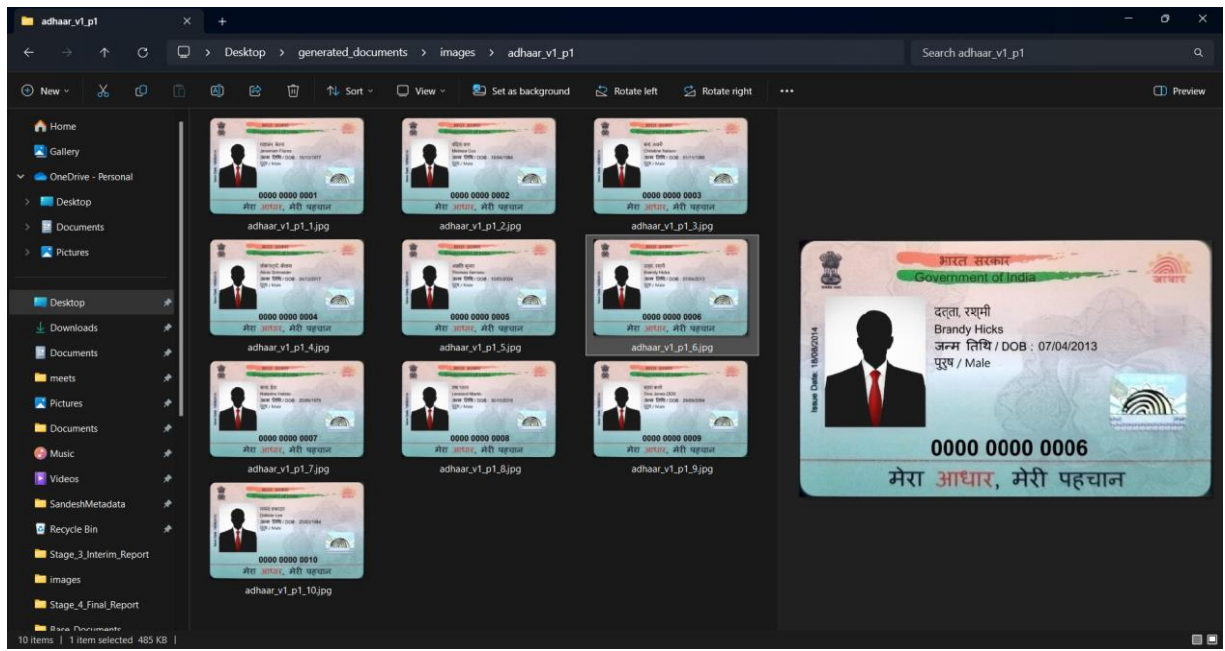

Figure 4.3: Documents generated via "*python generate_document_adhaar.py 10*"

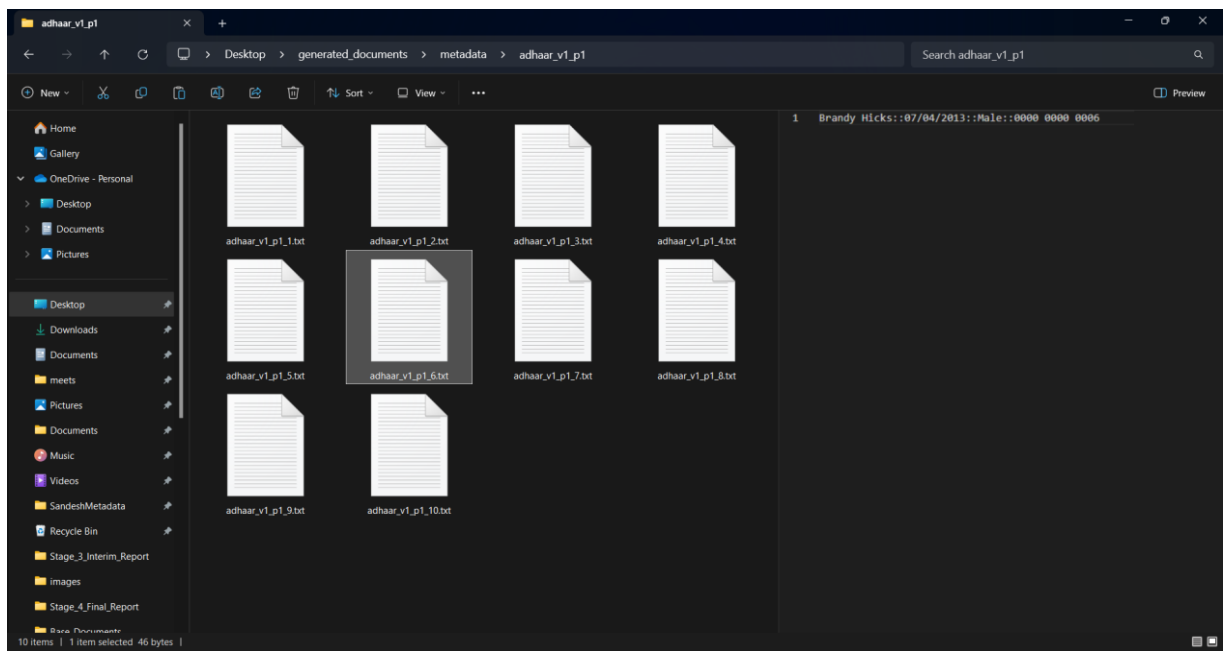

Figure 4.4: Annotation generated via "*python generate_document_adhaar.py 10*"

In the similar manner, the edited Driving Licence is shown in Figure 4.5. The same steps are followed as of Adhaar card.

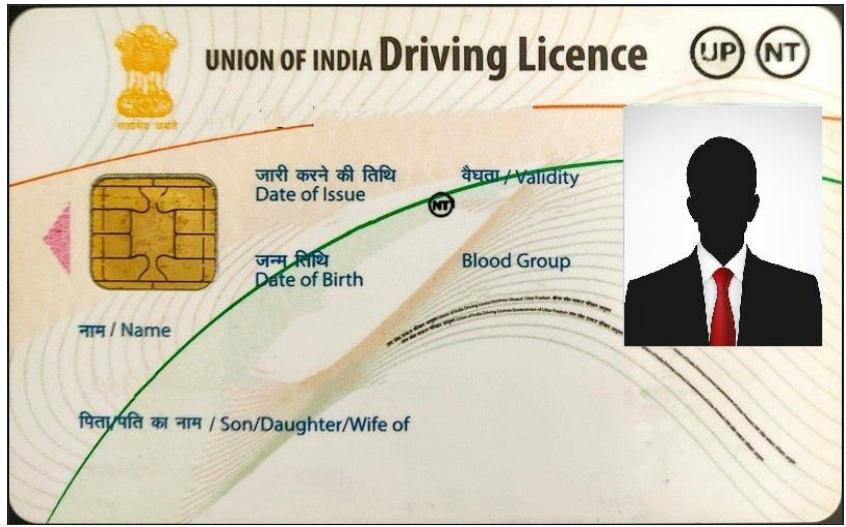

Figure 4.5: Edited Driving Licence – Base Template

The code for synthesizer is as follows:

```python
from faker import Faker
from PIL import Image, ImageDraw, ImageFont
import os
import sys

class DLData:
    dl_value: ""
    doi: ""
    dov: ""
    dob: ""
    blood_group: ""
    name: ""
    father_name: ""
    address: ""
    sign: ""
    file_number: ""
```

```python
faker = Faker()
font_1 = ImageFont.truetype("arial.ttf", 35)
font_2 = ImageFont.truetype("arial.ttf", 20)
font_3 = ImageFont.truetype("arialbd.ttf", 25)
font_4 = ImageFont.truetype("arial.ttf", 23)
font_5 = ImageFont.truetype("arial.ttf", 30)
font_6 = ImageFont.truetype("arialbd.ttf", 25)
font_sign = ImageFont.truetype("segoesc.ttf", 25)
dl_data = 1620240000001
args = sys.argv
max_dl = int(args[1])
alphabets = ['A', 'B', 'C', 'D', 'E', 'F', 'G', 'H', 'I', 'J', 'K', 'L', 'M', 'N', 'O', 'P', 'Q', 'R', 'S', 'T', 'U',
            'V', 'W', 'X', 'Y', 'Z']

def progress(count, total, suffix=''):
    bar_len = 60
    filled_len = int(round(bar_len * count / float(total)))
    percents = round(100.0 * count / float(total), 1)
    bar = '█' * filled_len + '-' * (bar_len - filled_len)
    sys.stdout.write('[%s] %s/%s ...%s\r' % (bar, str(count), str(total), suffix))
    sys.stdout.flush()

def createDatasetFolders():
    newpath = r'../generated_documents'
    if not os.path.exists(newpath):
        os.makedirs(newpath)
    newpath = r'../generated_documents/images'
    if not os.path.exists(newpath):
        os.makedirs(newpath)
    newpath = r'../generated_documents/images/dl_v1_p1'
```

```python
    if not os.path.exists(newpath):
        os.makedirs(newpath)
    newpath = r'../generated_documents/metadata'
    if not os.path.exists(newpath):
        os.makedirs(newpath)
    newpath = r'../generated_documents/metadata/dl_v1_p1'
    if not os.path.exists(newpath):
        os.makedirs(newpath)

def get_dl_value(temp):
    value = ""
    for i in range(15):
        a = temp % 10
        if i < 13:
            value = str(int(a)) + value
        else:
            value = alphabets[int(a)] + value
        temp /= 10
        if i == 10:
            value = "  " + value
    return value

def get_file_number(temp):
    value = ""
    for i in range(12):
        a = temp % 10
        if i < 2 or i > 10:
            value = alphabets[int(a)] + value
        else:
            value = str(int(a)) + value
        temp /= 10
    return value
```

```python
def get_dl_metadata(temp):
    dl = DLData()
    dl.dl_value = get_dl_value(temp)
    date1 = faker.date().split("-")
    dl.doi = date1[2] + "/" + date1[1] + "/" + date1[0]
    date2 = faker.date().split("-")
    dl.dov = date2[2] + "/" + date2[1] + "/" + date2[0]
    date3 = faker.date().split("-")
    dl.dob = date3[2] + "/" + date3[1] + "/" + date3[0]
    dl.blood_group = faker.random.choice(["A+", "A-", "B+", "B-", "O+", "O-", "AB+", "AB-"])
    dl.name = faker.name()
    dl.father_name = faker.name()
    dl.address = faker.address()
    dl.sign = faker.name()
    dl.file_number = get_file_number(temp)
    return dl

createDatasetFolders()
for i in range(max_dl):
    dl = get_dl_metadata(dl_data)
    # P1
    image = Image.open("../base_documents/DL_V1_P1.jpg")
    draw = ImageDraw.Draw(image)
    draw.text((200, 90), dl.dl_value, font=font_1, fill=(0, 0, 0))
    draw.text((238, 195), dl.doi, font=font_2, fill=(0, 0, 0))
    draw.text((435, 180), dl.dov, font=font_2, fill=(0, 0, 0))
    draw.text((238, 280), dl.dob, font=font_2, fill=(0, 0, 0))
    draw.text((435, 260), dl.blood_group, font=font_2, fill=(0, 0, 0))
    draw.text((70, 350), dl.name, font=font_3, fill=(0, 0, 0))
    draw.text((70, 430), dl.father_name, font=font_4, fill=(0, 0, 0))
```

```
    image.save("../generated_documents/images/dl_v1_p1/dl_v1_p1_" + str(i+1) + ".jpg")

    file1 = open("../generated_documents/metadata/dl_v1_p1/dl_v1_p1_" + str(i+1) + ".txt",
"w")

    L = [dl.dl_value + "::" + dl.doi + "::" + dl.dov + "::" + dl.dob + "::" + dl.blood_group + "::"
+ dl.name + "::" + dl.father_name]

    file1.writelines(L)

    file1.close()

    progress(i + 1, max_dl, "Driving Licence Generated")

    dl_data += 1
```

The above code is saved in a file named "*generate_document_dl.py*". In order to generate 10 different documents of Driving Licence, following command is required to do so:

*python generate_document_dl.py 10*

The output of the above command is shown in Figure 4.6 and metadata (annotation of each document) thus generated is shown in Figure 4.7.

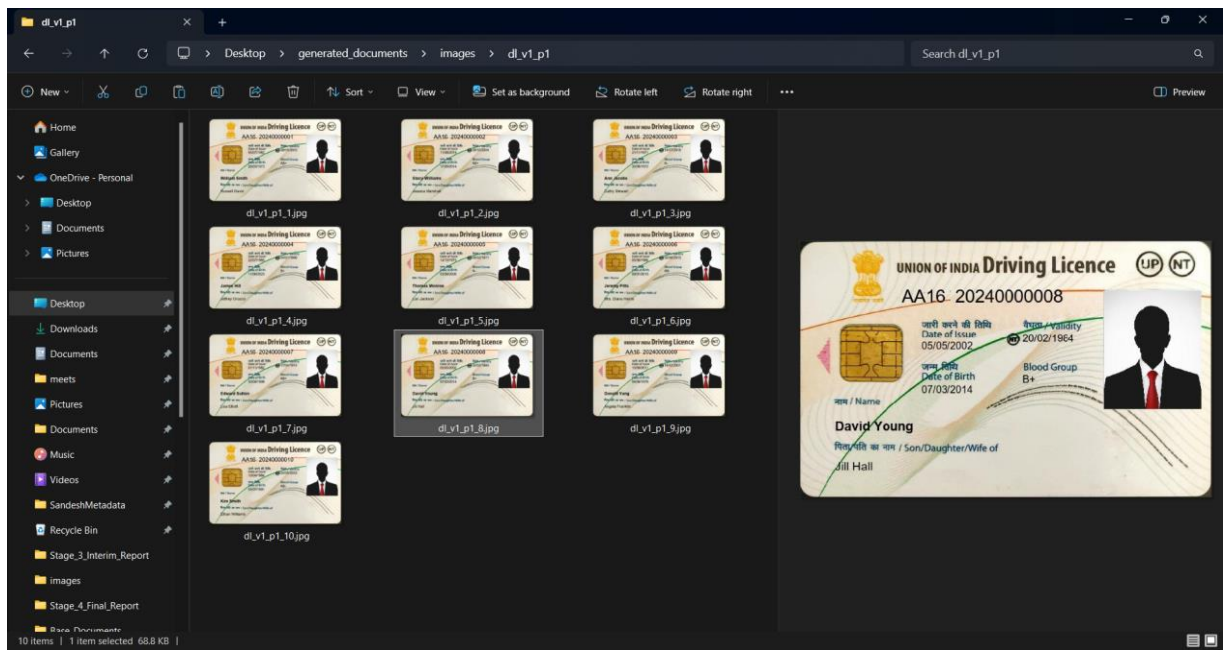

Figure 4.6: Documents generated via "*python generate_document_dl.py 10*"

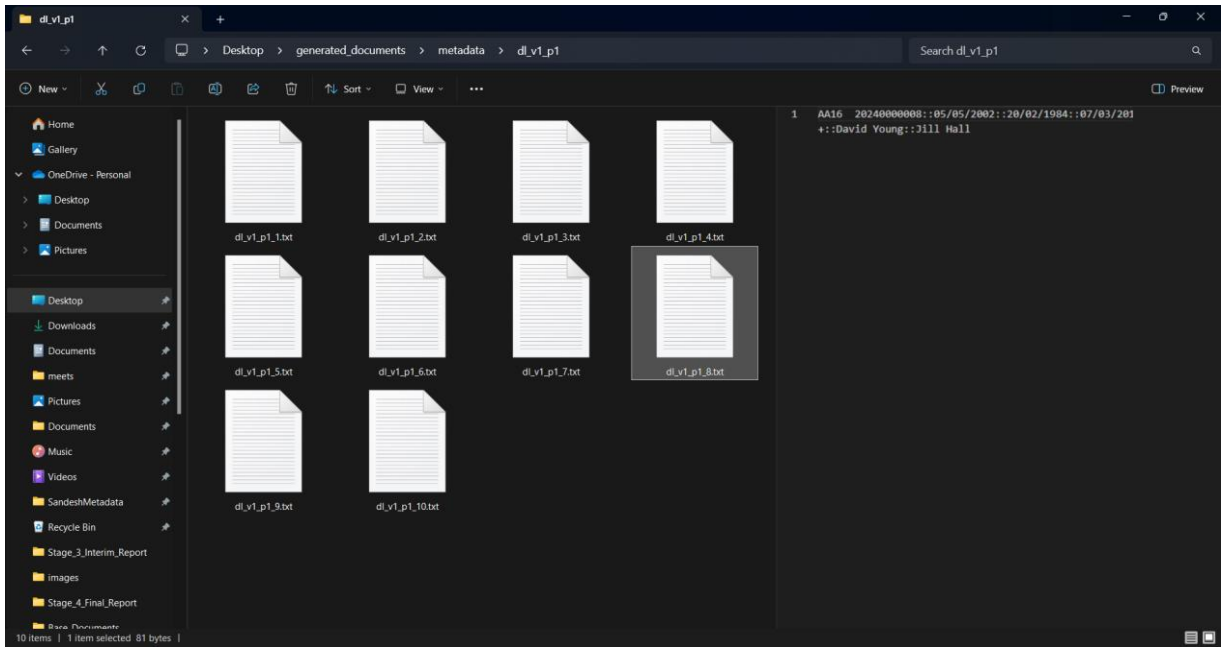

Figure 4.7: Annotation generated via "*python generate_document_dl.py 10*"

In case of PAN card, the edited PAN card is shown in Figure 4.8. The same steps are followed as of Adhaar card.

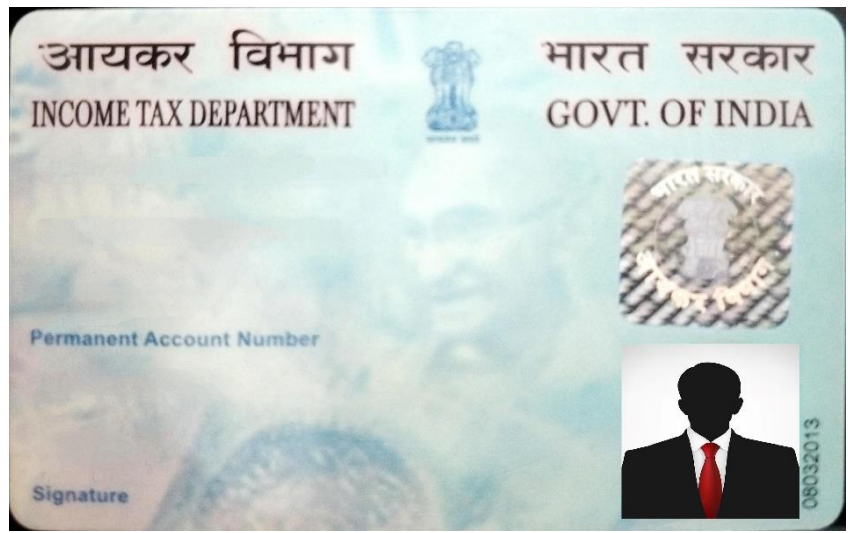

Figure 4.8: Edited PAN Card – Base Template

The code for synthesizer is as follows:

```python
from faker import Faker
from PIL import Image, ImageDraw, ImageFont
import os
import sys

class PanData:
    applicant_name: ""
    father_name: ""
    pan_value: ""
    dob: ""

fake = Faker()
font = ImageFont.truetype("arialbd.ttf", 105)
font_sign = ImageFont.truetype("segoesc.ttf", 100)
args = sys.argv
max_pan = int(args[1])
```

```python
pan_data = 1
alphabets = ['A', 'B', 'C', 'D', 'E', 'F', 'G', 'H', 'I', 'J', 'K', 'L', 'M', 'N', 'O', 'P', 'Q', 'R', 'S', 'T', 'U',
        'V', 'W', 'X', 'Y', 'Z']

def progress(count, total, suffix=''):
    bar_len = 60
    filled_len = int(round(bar_len * count / float(total)))
    percents = round(100.0 * count / float(total), 1)
    bar = '' * filled_len + '-' * (bar_len - filled_len)
    sys.stdout.write('[%s] %s/%s ...%s\r' % (bar, str(count), str(total), suffix))
    sys.stdout.flush()

def createDatasetFolders():
    newpath = r'../generated_documents'
    if not os.path.exists(newpath):
        os.makedirs(newpath)
    newpath = r'../generated_documents/images'
    if not os.path.exists(newpath):
        os.makedirs(newpath)
    newpath = r'../generated_documents/images/pan_v1'
    if not os.path.exists(newpath):
        os.makedirs(newpath)
    newpath = r'../generated_documents/metadata'
    if not os.path.exists(newpath):
        os.makedirs(newpath)
    newpath = r'../generated_documents/metadata/pan_v1'
    if not os.path.exists(newpath):
        os.makedirs(newpath)

def generate_pan(temp):
    pan = ""
```

```python
    for c in range(9):
        a = int(temp % 10)
        temp /= 10
        if 0 < c < 5:
            pan = str(a) + pan
        else:
            pan = alphabets[a] + pan
    return pan

def get_pan_metadata(temp):
    p = PanData()
    p.applicant_name = fake.name()
    p.father_name = fake.name()
    p.pan_value = generate_pan(temp)
    date = fake.date().split("-")
    p.dob = date[2] + "/" + date[1] + "/" + date[0]
    return p

createDatasetFolders()
for i in range(max_pan):
    pan = get_pan_metadata(pan_data)
    image = Image.open("../base_documents/PAN_V1.jpg")
    draw = ImageDraw.Draw(image)
    draw.text((120, 570), pan.applicant_name.upper(), font=font, fill=(0, 0, 0))
    draw.text((120, 810), pan.father_name.upper(), font=font, fill=(0, 0, 0))
    draw.text((120, 1090), pan.dob, font=font, fill=(0, 0, 0))
    draw.text((120, 1370), pan.pan_value, font=font, fill=(0, 0, 0))
    draw.text((120, 1700), pan.applicant_name.split(" ")[0], font=font_sign, fill=(0, 0, 0))
    image.save("../generated_documents/images/pan_v1/pan_v1_" + str(pan_data) + ".jpg")
    file1 = open("../generated_documents/metadata/pan_v1/pan_v1_" + str(pan_data) + ".txt", "w")
    L = [pan.applicant_name + "::" + pan.father_name + "::" + pan.pan_value + "::" + pan.dob]
```

```
file1.writelines(L)
file1.close()
progress(i + 1, max_pan, "PAN Generated")
pan_data += 1
```

The above code is saved in a file named "*generate_document_pan.py*". In order to generate 10 different documents of PAN card, following command is required to do so:

*python generate_document_pan.py 10*

The output of the above command is shown in Figure 4.9 and metadata (annotation of each document) thus generated is shown in Figure 4.10.

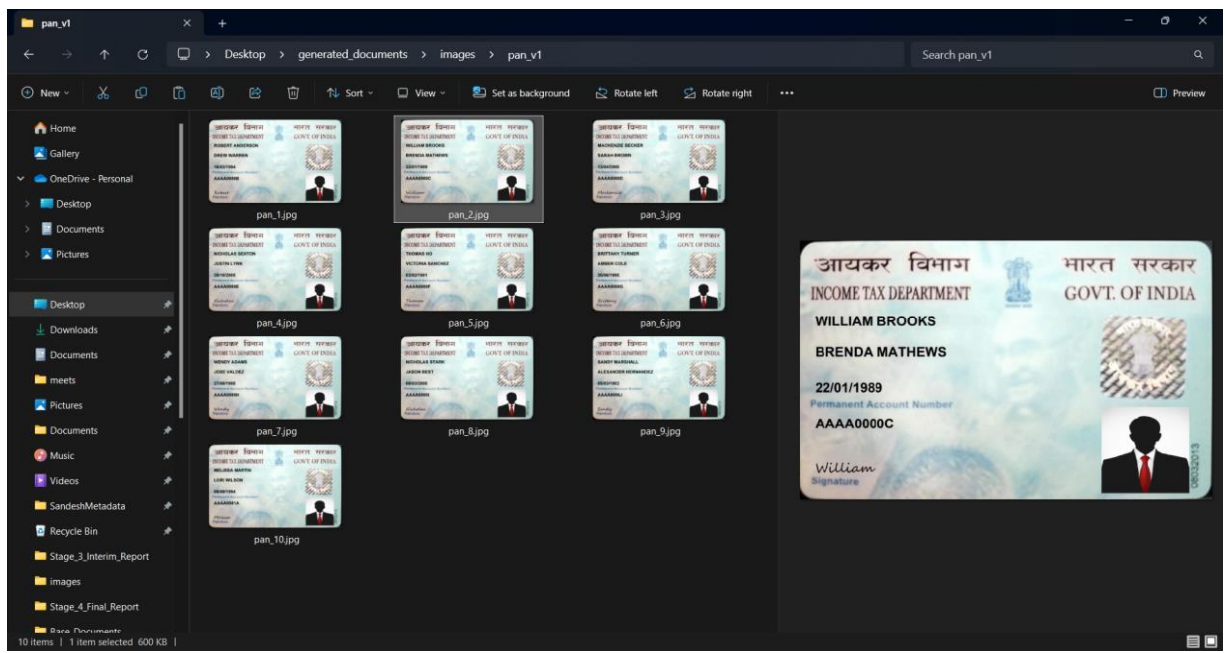

Figure 4.9: Documents generated via "*python generate_document_pan.py 10*"

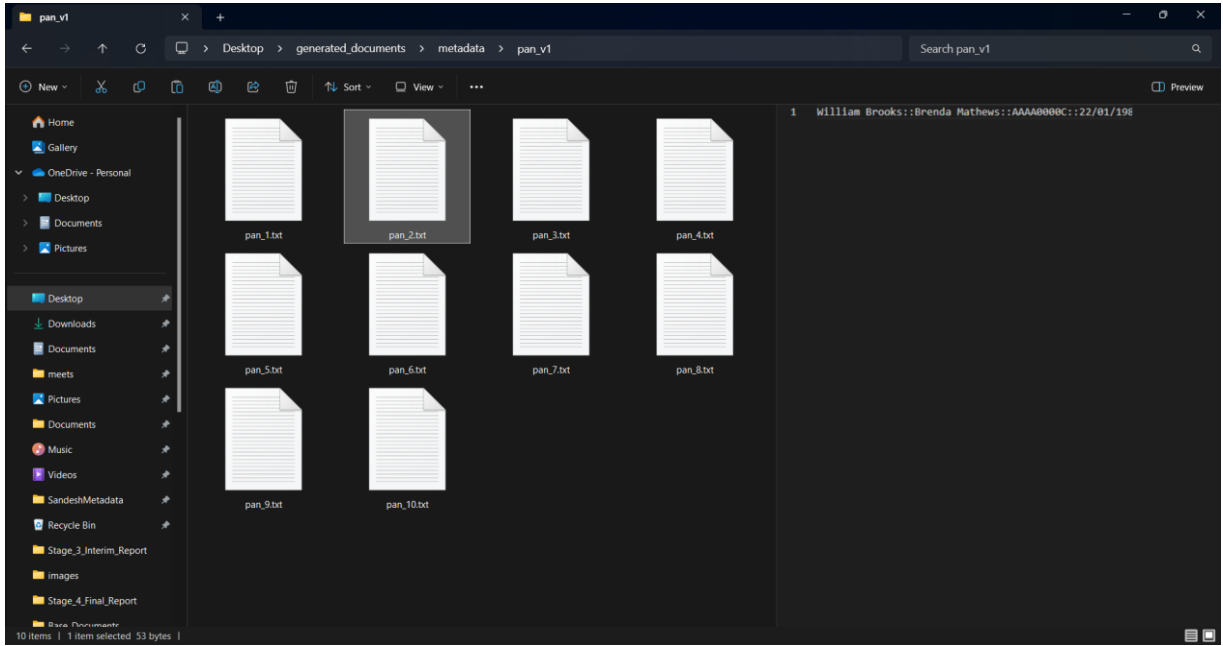

Figure 4.10: Annotation generated via "*python generate_document_pan.py 10*"

In case of Passport, the edited Passport is shown in Figure 4.11. The same steps are followed as of Adhaar card.

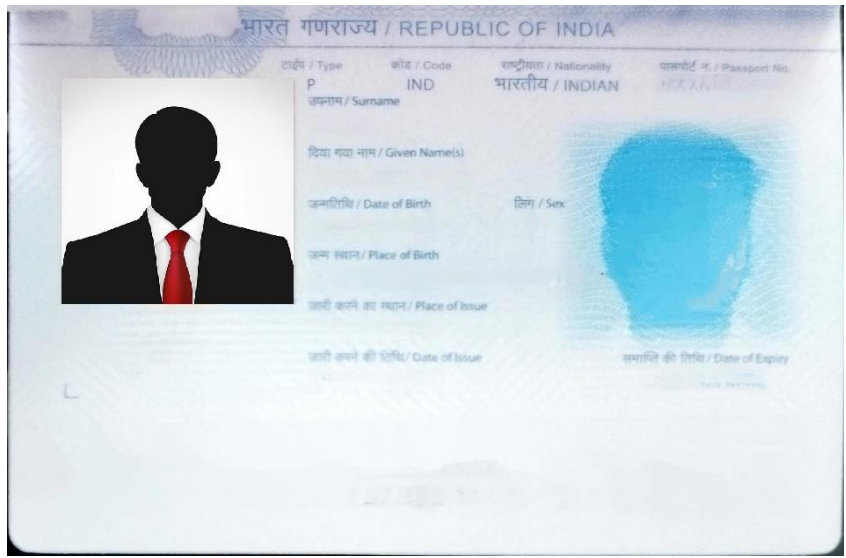

Figure 4.11: Edited Passport – Base Template

The code for synthesizer is as follows:

```python
from faker import Faker
from PIL import Image, ImageDraw, ImageFont
import os
import sys

class PassportData:
    passport_value: ""
    surname: ""
    name: ""
    gender: ""
    dob: ""
    pob: ""
    poi: ""
    doi: ""
    doe: ""
    father_name: ""
```

```python
    mother_name: ""

    spouse_name: ""

    address: ""

    file_number: ""

    old_passport_value: ""

    old_passport_doi: ""

    old_passport_poi: ""

faker = Faker()
font_1 = ImageFont.truetype("arialbd.ttf", 80)
font_2 = ImageFont.truetype("arialbd.ttf", 60)
font_3 = ImageFont.truetype("arialbd.ttf", 70)
font_4 = ImageFont.truetype("arial.ttf", 80)
font_5 = ImageFont.truetype("arial.ttf", 30)
font_6 = ImageFont.truetype("arialbd.ttf", 25)
font_sign = ImageFont.truetype("segoesc.ttf", 80)
passport_data = 1
args = sys.argv
max_passport = int(args[1])
alphabets = ['A', 'B', 'C', 'D', 'E', 'F', 'G', 'H', 'I', 'J', 'K', 'L', 'M', 'N', 'O', 'P', 'Q', 'R', 'S', 'T', 'U',
             'V', 'W', 'X', 'Y', 'Z']

def progress(count, total, suffix=''):
    bar_len = 60
    filled_len = int(round(bar_len * count / float(total)))
    percents = round(100.0 * count / float(total), 1)
    bar = '█' * filled_len + '-' * (bar_len - filled_len)
    sys.stdout.write('[%s] %s/%s ...%s\r' % (bar, str(count), str(total), suffix))
    sys.stdout.flush()

def createDatasetFolders():
```

```python
    newpath = r'../generated_documents'
    if not os.path.exists(newpath):
        os.makedirs(newpath)
    newpath = r'../generated_documents/images'
    if not os.path.exists(newpath):
        os.makedirs(newpath)
    newpath = r'../generated_documents/images/passport_v1_p1'
    if not os.path.exists(newpath):
        os.makedirs(newpath)
    newpath = r'../generated_documents/metadata'
    if not os.path.exists(newpath):
        os.makedirs(newpath)
    newpath = r'../generated_documents/metadata/passport_v1_p1'
    if not os.path.exists(newpath):
        os.makedirs(newpath)

def get_passport_value(temp):
    value = ""
    for i in range(8):
        a = temp % 10
        if i < 7:
            value = str(int(a)) + value
        else:
            value = alphabets[int(a)] + value
        temp /= 10
    return value

def get_file_number(temp):
    value = ""
    for i in range(15):
        a = temp % 10
        if i > 12:
```

```python
            value = alphabets[int(a)] + value
        else:
            value = str(int(a)) + value
        temp /= 10
    return value

def get_passport_metadata(temp):
    passport = PassportData()
    passport.passport_value = get_passport_value(temp)
    passport.surname = faker.name().split(" ")[-1].upper()
    passport.name = faker.name().split(" ")[0]
    passport.gender = faker.random.choice(['M', 'F'])
    date3 = faker.date().split("-")
    passport.dob = date3[2] + "/" + date3[1] + "/" + date3[0]
    passport.pob = faker.random.choice(
        ["Tokyo", "Jakarta", "Delhi", "Guangzhou", "Mumbai", "Manila", "Shanghai", "São
Paulo", "Seoul", "Mexico City",
         "Cairo", "New York", "Dhaka", "Beijing", "Kolkata", "Bangkok", "Shenzhen",
"Moscow", "Buenos Aires", "Lagos",
         "Bangalore"]).upper()
    passport.poi = faker.random.choice(
        ["Tokyo", "Jakarta", "Delhi", "Guangzhou", "Mumbai", "Manila", "Shanghai", "São
Paulo", "Seoul", "Mexico City",
         "Cairo", "New York", "Dhaka", "Beijing", "Kolkata", "Bangkok", "Shenzhen",
"Moscow", "Buenos Aires", "Lagos",
         "Bangalore"]).upper()
    date1 = faker.date().split("-")
    passport.doi = date1[2] + "/" + date1[1] + "/" + date1[0]
    date2 = faker.date().split("-")
    passport.doe = date2[2] + "/" + date2[1] + "/" + date2[0]
    passport.father_name = faker.name().upper()
    passport.mother_name = faker.name().upper()
    passport.spouse_name = faker.name().upper()
```

```python
    passport.address = faker.address().upper()
    passport.file_number = get_file_number(temp)
    passport.old_passport_value = get_passport_value(temp - 1)
    date4 = faker.date().split("-")
    passport.old_passport_doi = date4[2] + "/" + date4[1] + "/" + date4[0]
    passport.old_passport_poi = faker.random.choice(
        ["Tokyo", "Jakarta", "Delhi", "Guangzhou", "Mumbai", "Manila", "Shanghai", "São
Paulo", "Seoul", "Mexico City",
         "Cairo", "New York", "Dhaka", "Beijing", "Kolkata", "Bangkok", "Shenzhen",
"Moscow", "Buenos Aires", "Lagos",
         "Bangalore"]).upper()
    return passport

createDatasetFolders()
for i in range(max_passport):
    passport = get_passport_metadata(passport_data)
    # P1
    image = Image.open("../base_documents/PASSPORT_V1_P1.jpg")
    draw = ImageDraw.Draw(image)
    draw.text((2200, 230), passport.passport_value, font=font_1, fill=(0, 0, 0))
    draw.text((1000, 370), passport.surname, font=font_2, fill=(0, 0, 0))
    draw.text((1000, 540), passport.name.upper(), font=font_2, fill=(0, 0, 0))
    draw.text((1000, 710), passport.dob, font=font_2, fill=(0, 0, 0))
    draw.text((1750, 710), passport.gender, font=font_2, fill=(0, 0, 0))
    draw.text((1000, 880), passport.pob, font=font_2, fill=(0, 0, 0))
    draw.text((1000, 1050), passport.poi, font=font_2, fill=(0, 0, 0))
    draw.text((1000, 1220), passport.doi, font=font_2, fill=(0, 0, 0))
    draw.text((2100, 1220), passport.doe, font=font_2, fill=(0, 0, 0))
    draw.text((300, 1100), passport.name, font=font_sign, fill=(0, 0, 0))
    draw.text((200,                                                               1600),
"P<IND<<<<<<<<<<<<<<<<<<<<<<<<<<<<<<<<<<<<<<<<<<",        font=font_4,
fill=(0, 0, 0))
```

```
  draw.text((200,              1700),              passport.passport_value    +
"<<<<<<<<<<<<<<<<<<<<<<<<<<<<<<<<<<<<<<<<<<<<<<", font=font_4, fill=(0, 0, 0))
  image.save("../generated_documents/images/passport_v1_p1/passport_v1_p1_" + str(i+1) +
".jpg")
  file1    =    open("../generated_documents/metadata/passport_v1_p1/passport_v1_p1_"    +
str(i+1) + ".txt", "w")
  L = [passport.passport_value + "::" + passport.surname + "::" + passport.name + "::" +
passport.dob + "::" + passport.gender + "::" + passport.pob + "::" + passport.poi + "::" +
passport.doi + "::" + passport.doe]
  file1.writelines(L)
  file1.close()
  progress(i + 1, max_passport, "Passport Generated")
  passport_data += 1
```

The above code is saved in a file named "*generate_document_passport.py*". In order to generate
10 different documents of Passport, following command is required to do so:

*python generate_document_passport.py 10*

The output of the above command is shown in Figure 4.12 and metadata (annotation of each
document) thus generated is shown in Figure 4.13.

Figure 4.12: Documents generated via "*python generate_document_passport.py 10*"

Figure 4.13: Annotation generated via "*python generate_document_passport.py 10*"

In case of Voter card, the edited Voter card is shown in Figure 4.14. The same steps are followed as of Adhaar card.

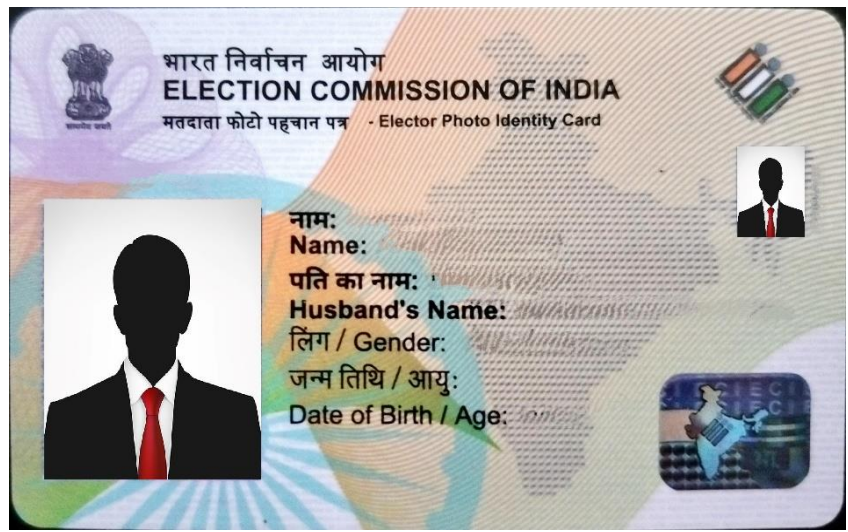

Figure 4.14: Edited Voter card – Base Template

The code for synthesizer is as follows:

```python
from faker import Faker
from PIL import Image, ImageDraw, ImageFont
import os
import sys

class VoterCardData:
    applicant_name: ""
    applicant_name_hi: ""
    husband_name: ""
    husband_name_hi: ""
    votercard_value: ""
    dob: ""
    gender: ""
    gender_hi: ""

fake = Faker()
```

```python
faker_hi = Faker("hi_IN")
font = ImageFont.truetype("arial.ttf", 80)
font_bd = ImageFont.truetype("arialbd.ttf", 80)
font_bd_1 = ImageFont.truetype("arialbd.ttf", 90)
font_2 = ImageFont.truetype("arial.ttf", 70)
font_hi          =          ImageFont.truetype("C:/Users/meets/OneDrive/Desktop/MS          Thesis
Work/Code/Nirmala.ttf", 80)
font_hi_2        =          ImageFont.truetype("C:/Users/meets/OneDrive/Desktop/MS          Thesis
Work/Code/Nirmala.ttf", 70)
args = sys.argv
max_votercard = int(args[1])
votercard_data = 1
alphabets = ['A', 'B', 'C', 'D', 'E', 'F', 'G', 'H', 'I', 'J', 'K', 'L', 'M', 'N', 'O', 'P', 'Q', 'R', 'S', 'T', 'U',
           'V', 'W', 'X', 'Y', 'Z']

def progress(count, total, suffix=''):
    bar_len = 60
    filled_len = int(round(bar_len * count / float(total)))
    percents = round(100.0 * count / float(total), 1)
    bar = '█' * filled_len + '-' * (bar_len - filled_len)
    sys.stdout.write('[%s] %s/%s ...%s\r' % (bar, str(count), str(total), suffix))
    sys.stdout.flush()

def createDatasetFolders():
    newpath = r'../generated_documents'
    if not os.path.exists(newpath):
        os.makedirs(newpath)
    newpath = r'../generated_documents/images'
    if not os.path.exists(newpath):
        os.makedirs(newpath)
    newpath = r'../generated_documents/images/votercard_v1'
    if not os.path.exists(newpath):
```

```python
        os.makedirs(newpath)
    newpath = r'../generated_documents/metadata'
    if not os.path.exists(newpath):
        os.makedirs(newpath)
    newpath = r'../generated_documents/metadata/votercard_v1'
    if not os.path.exists(newpath):
        os.makedirs(newpath)

def generate_votercard(temp):
    votercard = ""
    for c in range(10):
        a = int(temp % 10)
        temp /= 10
        if c < 7:
            votercard = str(a) + votercard
        else:
            votercard = alphabets[a] + votercard
    return votercard

def get_votercard_metadata(temp):
    p = VoterCardData()
    p.applicant_name = fake.name()
    p.applicant_name_hi = faker_hi.name()
    p.husband_name = fake.name()
    p.husband_name_hi = faker_hi.name()
    p.votercard_value = generate_votercard(temp)
    date = fake.date().split("-")
    p.dob = date[2] + "-" + date[1] + "-" + date[0]
    if (temp <= max_votercard * 0.4) or (max_votercard * 0.8 < temp <= max_votercard * 0.9):
        p.gender_hi = "पुरुष"
        p.gender = "Male"
    else:
```

```python
        p.gender_hi = "महिला"
        p.gender = "Female"
    return p

createDatasetFolders()
for i in range(max_votercard):
    votercard = get_votercard_metadata(votercard_data)
    image = Image.open("../base_documents/VOTERCARD_V1.jpg")
    draw = ImageDraw.Draw(image)
    draw.text((260, 620), votercard.votercard_value, font=font_bd, fill=(0, 0, 0))
    draw.text((1290, 760), votercard.applicant_name_hi, font=font_hi, fill=(0, 0, 0))
    draw.text((1410, 870), votercard.applicant_name, font=font_bd_1, fill=(0, 0, 0))
    draw.text((1600, 1000), votercard.husband_name_hi, font=font_hi, fill=(0, 0, 0))
    draw.text((1950, 1120), votercard.husband_name, font=font_bd_1, fill=(0, 0, 0))
    draw.text((1720, 1240), votercard.gender_hi + " / " + votercard.gender, font=font_hi, fill=(0,
0, 0))
    draw.text((1940, 1540), votercard.dob, font=font, fill=(0, 0, 0))
    image.save("../generated_documents/images/votercard_v1/votercard_v1_"          +
str(votercard_data) + ".jpg")
    file1      =     open("../generated_documents/metadata/votercard_v1/votercard_v1_"      +
str(votercard_data) + ".txt", "w")
    L   =   [votercard.applicant_name   +   "::"   +   votercard.husband_name   +   "::"   +
votercard.votercard_value + "::" + votercard.dob]
    file1.writelines(L)
    file1.close()
    progress(i + 1, max_votercard, "Passport Generated")
    votercard_data += 1
```

The above code is saved in a file named "*generate_document_votercard.py*". In order to generate 10 different documents of Voter card, following command is required to do so:

*python generate_document_votercard.py 10*

The output of the above command is shown in Figure 4.15 and metadata (annotation of each document) thus generated is shown in Figure 4.16.

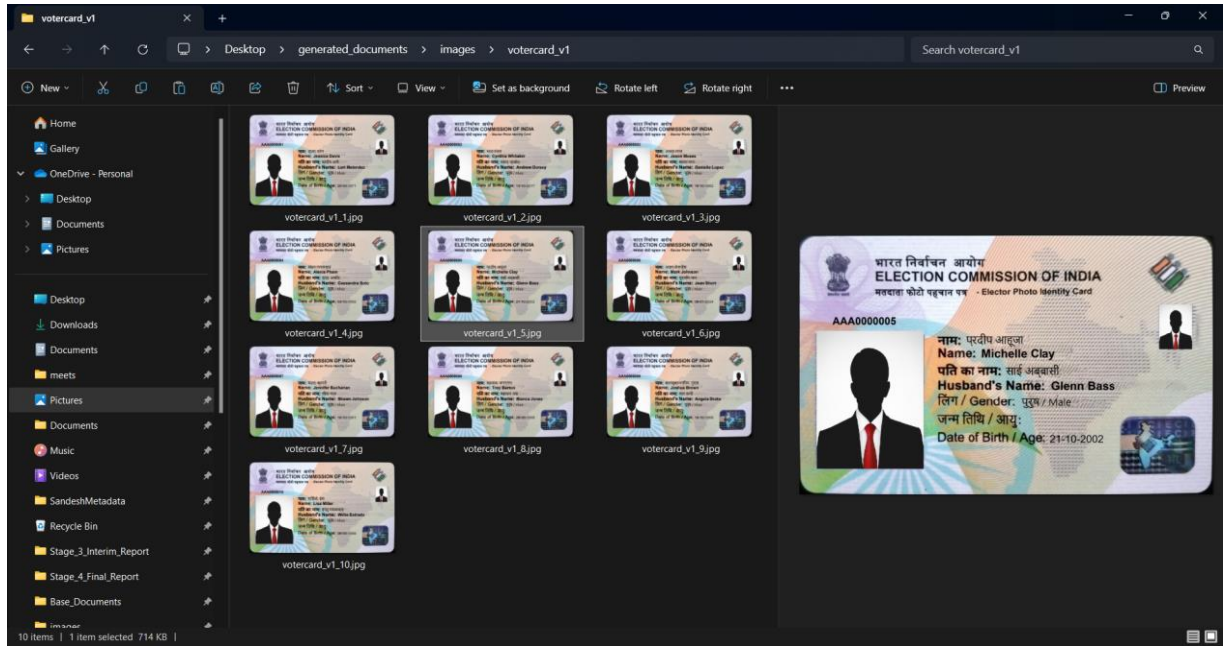

Figure 4.15: Documents generated via "*python generate_document_votercard.py 10*"

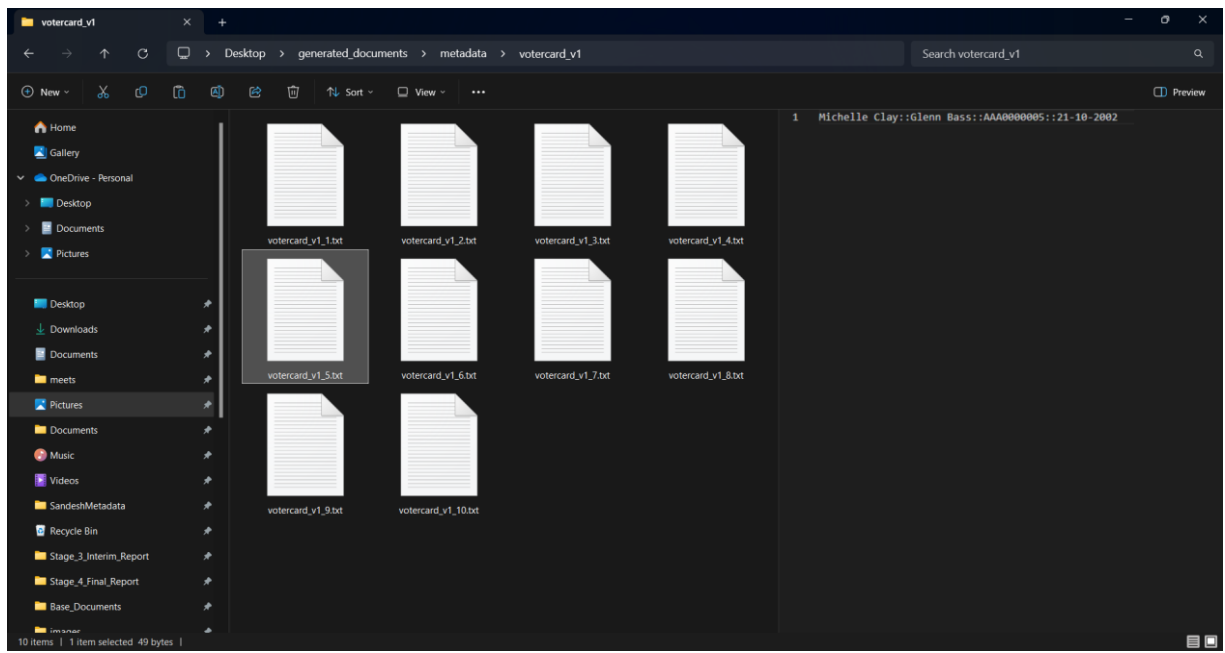

Figure 4.16: Annotation generated via "*python generate_document_votercard.py 10*"

### 4.2.2   Data Preprocessing

Data preprocessing is important to remove the noise from raw input into a format suitable for consumption by Machine Learning model. It encompasses several steps like cleaning of data from errors and inconsistencies, missing values, outliers, normalization, and encoding of categorical variables. The intention is essentially to improve the quality and relevance of the data input into the model in order to enable effective learning through the data.

The quality of data determines the quality of any successful machine learning. Models are only good as the data they have been trained on, and the badly processed data can easily result in very bad predictions, similar to overfitting or underfitting itself. For example, the existence of outliers may influence the performance of models or affect their generalization, while missing values will lead to incomplete learning.

This will apply more structuring in such data so that algorithms will be able to find patterns and improve the accuracy of models designed, thus reducing the complexity of computations. This minimizes the biases and errors present in data that allows the data models to generalize out rightly in the real world. In short, it is one of the crucial optimization steps toward better performance of a machine learning model and improvement in its decision-making capabilities.

Luckly in current scenario, the created in previous section close enough to real world scenario. But it lacks to cover wide range of scenarios. For instance, it only covers a single case in which the document provided to the bank is simple orientation with the image having exact size of the actual document. But that is not all. Following are the valid scenarios:

1. Document with simple orientation with the image having exact size of the actual document. (e.g. Figure 3.5)

2. Document with exact image size in greyscale or black & white, which is a common scanning format. (e.g. Figure 3.6)

3. Document with coloured image on an A4 sheet. The position of the image can be anywhere on the sheet. (e.g. Figure 3.7)

4. Document with greyscale or black & white image on an A4 sheet. The position of the image can be anywhere on the sheet. (e.g. Figure 3.8)

5. Document with some dirt or noise on the A4 sheet, it is scanned on.

6. Document with skewed image, e.g. the actual document is rotated by few degrees.

The scope of the thesis is limited to first 4 scenarios only. Given the scenarios, let's jump into how to achieve rest of the 3 scenarios.

The following code is used to generate all the 4 scenarios:

```python
from PIL import Image, ImageDraw
import sys
import shutil

class FileData:
    src: ""
    width: 0
    height: 0
    wmm: 0
    hmm: 0

    def __init__(self, s, w, h, wm, hm):
        self.src = s
        self.width = w
        self.height = h
        self.wmm = wm
        self.hmm = hm

def progress(count, total, suffix=''):
    bar_len = 60
    filled_len = int(round(bar_len * count / float(total)))
    percents = round(100.0 * count / float(total), 1)
    bar = '=' * filled_len + '-' * (bar_len - filled_len)
    sys.stdout.write('[%s] %s/%s ...%s\r' % (bar, str(count), str(total), suffix))
    sys.stdout.flush()
```

```python
def save_image(im, d, n):
    im.save("../Dataset_Advanced_Test/data/images/" + d + "/" + n + ".jpg")

def save_original(im, d, n, s, w, h, sx, sy):
    save_image(im, d, n)
    grayscale_image = im.convert("L")
    save_image(grayscale_image, d, n + "_greyscale")

def save_a4(im, d, n, a, w, h):
    position = [[100, 100], [100, 1500], [1000, 100], [1000, 1500], [2000, 100], [2000, 1500]]
    for p in position:
        image = Image.open("test.jpg")
        x = 600
        y = int((x * w) / h)
        resized_image = im.resize((y, x))
        image.paste(resized_image, (p[1], p[0]))
        save_image(image, d, n + "_a4_" + str(p[1]) + "_" + str(p[0]))
        grayscale_image = image.convert("L")
        save_image(grayscale_image, d, n + "_a4_" + str(p[1]) + "_" + str(p[0]) + "_greyscale")

arr = [FileData("adhaar_v1_p1", 2830, 1770, 85, 54),
       FileData("dl_v1_p1", 804, 504, 85, 54),
       FileData("pan", 3200, 2019, 85, 54),
       FileData("passport_v1_p1", 2783, 1847, 85, 54),
       FileData("votercard_v1", 3200, 2015, 85, 54)]

count = 0
start = 1
total = 10
for i in range(start, total + 1):
    for a in arr:
```

```
        progress(count, (total - start + 1) * 1 * 5, "Completed")
        image_original = Image.open("PATH/TO/BASE/IMAGE/DATA/training_data/" + a.src
+ "/" + a.src + "_" + str(i) + ".jpg")
        dir_name = "all_dataset"
        save_original(image_original, dir_name, a.src + "_" + str(i), a.src, a.width, a.height, 0, 0)
        save_a4(image_original, dir_name, a.src + "_" + str(i), a, a.width, a.height)
        count = count + 1
```

**Explanation**:

The above is a Python image manipulation and processing code snippet. It is designed to load images, change them into grayscale as well as resized versions, and save them into specific folders. Additionally, the code also shows a progress bar as the operation proceeds.

1. Class **FileData**

   FileData is a class of information gained after processing images. It includes:

   src: name of source file

   width: width in pixels

   height: height in pixels

   wmm: width of the document, in millimeters, presumably as a scale factor

   hmm: height of the document, in millimeters

   The __init__ constructor sets these up for every instance.

2. Function ***progress***(count, total, suffix='')

   This function displays a progress bar in the console, showing how much of the task is done. It calculates the percentage of the progress and prints it out immediately. The bar updates on the same line using carriage return(\r).

3. Function *save_image*(im, d, n):

It writes an image (im) to a particular directory (d) under the name (n). The format of the saved .jpg file is located at ../Dataset_Advanced_Test/data/images/<d>/<n>.jpg.

4. Function *save_original*(im, d, n, s, w, h, sx, sy):

It calls save_image to save the original image (im). Then it converts the image to greyscale via the call im.convert("L") and saves that greyscale version with the name n + " _greyscale".

5. Function *save_a4*(im, d, n, a, w, h):

It resizes the input image and pastes it onto a pre-existing "A4-sized" image (presumably the file test.jpg) at different positions.

In position list, the coordinates are present where the image is pasted in A4-sized canvas

Original image is resized based on ratio of w(width) to h(height) for filling 600 pixels of width.

The resized image is pasted at six different positions on the A4 canvas, and the pasted and grayscale versions are saved by giving a unique file name that indicates the position.

6. ***Main Loop***:

The list "arr" contains multiple FileData objects, which can be of different types as follows: adhaar_v1_p1, dl_v1_p1, etc.

This loop runs from start to total (10 iterations), and in each iteration, one image file in the arr list is processed.

It loads each image using Image.open(), checks the loop index i to decide whether to save an image into a train, validation or test directory, and calls the sub-function of save_original and save_a4 to process and save the image in two resolutions, original and A4.

It updates the progress bar in the loop using the function progress().

7. Main Concepts:

***Image Manipulation***: The code opens, resizes, transforms, and saves images using Python Imaging Library, PIL (now maintained under Pillow).

***Tracking Progress***: It dynamically puts a progress bar to give feedback on the execution of the image processing.

***Directory Management***: This depends on the index I, the images are therefore saved in different subfolders for the purposes of training, validation, and testing.

### 4.2.3 Dataset Spitting

Dataset splitting in machine learning refers to the division of the dataset into separate subsets, used for training, validation, and testing. The majority split tends to be over the two subsets called training and test set. It's often divided into a 4:1 or 7:3 ratio. The training set provides the learning of the model, while on the other hand, the test set evaluates what the model will do on unseen data. In a few instances, one adds a validation set where the hyperparameter tuning of the model happens. Hence, it will form a split over three subsets: training-validation-testing.

Dataset splitting is important as it assures that the model will generalise well to new, unseen data, and it won't overfit-that is, a model performs well on the training data but not on new data. Proper splitting allows for an unbiased evaluation of performance since one is testing on the model against unseen data, simulating real-life scenarios. Such an approach provides the guarantee of a model's robustness and reliability; therefore, this is essential in testing its actual ability to predict.

In current problem statement, the dataset will be divided in the ratio of 7:2:1. That means, for 100 images, 70 images will be used for training, 20 images will be used for validation and 10 will be used for testing. And all the 100 images will be mutually exclusive, i.e. having no overlapping.

Therefore, in the code snippet, main loop can be modified as follows:

```python
from PIL import Image, ImageDraw
import sys
import shutil

class FileData:
    src: ""
    width: 0
.
.
.
count = 0
```

```python
start = 1
total = 100
for i in range(start, total + 1):
    for a in arr:
        progress(count, (total - start + 1) * 14 * 5, "Completed")
        image_original = Image.open("PATH/TO/BASE/IMAGE/DATA/training_data/" + a.src
+ "/" + a.src + "_" + str(i) + ".jpg")
        dir_name = ""
        if 1 <= i <= 70:
            dir_name = "train"
        if 71 <= i <= 90:
            dir_name = "validation"
        if 91 <= i <= 100:
            dir_name = "test"
        save_original(image_original, dir_name, a.src + "_" + str(i), a.src, a.width, a.height, 0, 0)
        save_a4(image_original, dir_name, a.src + "_" + str(i), a, a.width, a.height)
        count = count + 14
```

Here, ***dir_name*** variable is used to distinguish between train, validation and test directory. And the image is saved in respective ***dir_name***.

### 4.3    Model Training For Document Identification

For the purpose of Document Identification, YOLO is the suitable library for Object Detection. Currently YOLO v8 will be used. There 3 steps in training a model using YOLO provided below:

1. Folder Structure Dataset

2. Hyperparameter

3. Program for Training

### 4.3.1    Folder Structure of Dataset

First, the dataset prepared needs to be structured shown in Figure 4.17.

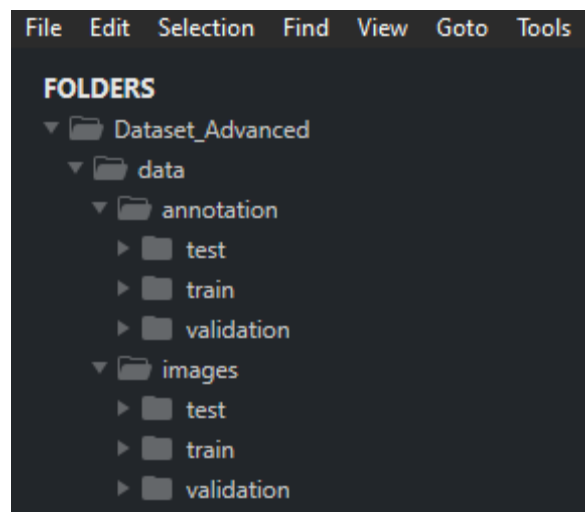

Figure 4.17: Directory structure of Dataset for model training in YOLO v8

The "annotation" folder contains data points per document generated using the "faker" library. Similarly, "images" folder contains all the documents generated in previous section. The "test", "train" and "validation" folders are common in both "annotation" and "images" folders. This is done for dataset splitting.

### 4.3.2 Hyperparameters

Next, hyperparameters will be required to train the model. It is a good practice to tune hyperparameters case by case. Since tuning them can also help the Machine Learning models to give better results. The hyperparameters and their values used for training the model for document identification are provided in Table 4.1.

Table 4.1: Hyperparameters and their values

| Hyperparameters | Values |
|---|---|
| task | detect |
| mode | train |
| model | PATH\TO\best.pt |
| data | config_advanced.yaml |
| epochs | 16 |
| patience | 50 |
| batch | 16 |
| imgsz | 640 |
| save | true |
| save_period | -1 |
| cache | false |
| device | null |
| workers | 8 |
| project | null |
| name | null |
| exist_ok | false |
| pretrained | false |
| optimizer | SGD |
| verbose | true |
| seed | 0 |
| deterministic | true |
| single_cls | false |
| image_weights | false |
| rect | false |
| cos_lr | false |
| close_mosaic | 10 |
| resume | false |
| amp | true |

| overlap_mask | true |
|---|---|
| mask_ratio | 4 |
| dropout | 0 |
| val | true |
| split | val |
| save_json | false |
| save_hybrid | false |
| conf | null |
| iou | 0.7 |
| max_det | 300 |
| half | false |
| dnn | false |
| plots | true |
| source | null |
| show | false |
| save_txt | false |
| save_conf | false |
| save_crop | false |
| hide_labels | false |
| hide_conf | false |
| vid_stride | 1 |
| line_thickness | 3 |
| visualize | false |
| augment | false |
| agnostic_nms | false |
| classes | null |
| retina_masks | false |
| boxes | true |
| format | torchscript |
| keras | false |
| optimize | false |
| int8 | false |
| dynamic | false |
| simplify | false |
| opset | null |
| workspace | 4 |
| nms | false |
| lr0 | 0.01 |
| lrf | 0.01 |
| momentum | 0.937 |
| weight_decay | 0.0005 |
| warmup_epochs | 3 |
| warmup_momentum | 0.8 |
| warmup_bias_lr | 0.1 |
| box | 7.5 |
| cls | 0.5 |
| dfl | 1.5 |
| fl_gamma | 0 |
| label_smoothing | 0 |
| nbs | 64 |
| hsv_h | 0.015 |
| hsv_s | 0.7 |
| hsv_v | 0.4 |
| degrees | 0 |
| translate | 0.1 |
| scale | 0.5 |
| shear | 0 |
| perspective | 0 |
| flipud | 0 |
| fliplr | 0.5 |
| mosaic | 1 |
| mixup | 0 |
| copy_paste | 0 |
| cfg | null |
| v5loader | false |
| tracker | botsort.yaml |
| save_dir | runs\detect\train2 |

### 4.3.3 Program for Training

At last, following code snippet is needed for training a Machine Learning model to identify and detected location of documents in an image:

```python
from ultralytics import YOLO

model = YOLO("PATH\\TO\\best.pt")
results = model.train(data="config_advanced.yaml", epochs=16, batch=16)
```

The output of the console or command prompt will be similar to the Figure 4.18.

Figure 4.18: Output while training the Machine Learning model in YOLO v8

Essentially, the "best.pt" file represents the last best trained weights, depending on the dataset provided. While training a new cycle, the "best.pt" file gets loaded and the new dataset, if changed or updated, is fed into the model and models corrects itself. Thereby, it generates new weights. This is validated and if the results are better than the previous epochs, Current version becomes the "best.pt". The ideal number of epoch or training cycle is 16. Once all the training cycle is completed. The updated Machine Learning model can be found in "train" folder shown in Figure 4.19.

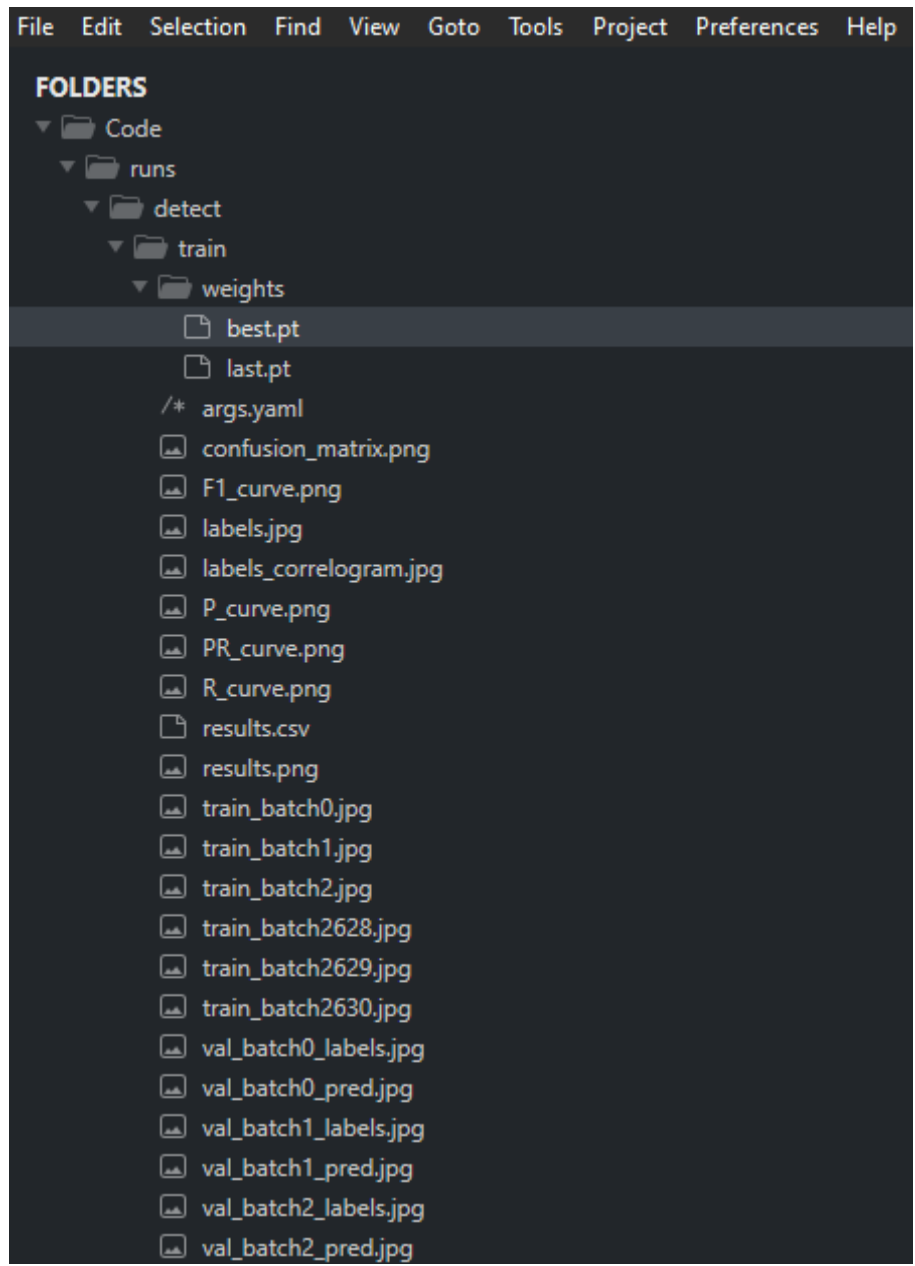

Figure 4.19: The "best.pt" weights file location

If training is done multiple times in separate sessions, multiple "train" folders will be created inside "detect" folder like "train", "train2", "train3", etc.

## 4.4   Data Extraction using Templatization

Once the document identification model is trained. Its time extract the data points from the image. Remember in Chapter 3, previous chapter, documents templates and their configurations were defined as JSON. It's time to use them now. Since the real crux of the problem statement is finding right bounding box to extract data, from previous section location of the document in an image is known. Performing OCR and discovering the "Identifying Field" can help to find resultant bounding box as per the equations derived in Chapter 3. Following code snippet will find the resultant bounding box:

```
.
.
.
SX1 = data["identifying_region"]["osx"]
SY1 = data["identifying_region"]["osy"]
EX1 = data["identifying_region"]["oex"]
EY1 = data["identifying_region"]["oey"]
OSX1 = data["identifying_region"]["isx"]
OSY1 = data["identifying_region"]["isy"]
OEX1 = data["identifying_region"]["iex"]
OEY1 = data["identifying_region"]["iey"]
FSX1 = found_id_box[0]
FSY1 = found_id_box[1]
FEX1 = found_id_box[2]
FEY1 = found_id_box[3]
extracted_data = ""
for d in data["data_regions"]:
        SX2 = d["osx"]
        SY2 = d["osy"]
        EX2 = d["oex"]
        EY2 = d["oey"]
        ASX1 = ((FSX1 * (OEX1 - OSX1)) + ((FEX1 - FSX1) * (SX2 - OSX1))) / (OEX1 - OSX1)
```

```python
        ASY1 = ((FSY1 * (OEY1 - OSY1)) + ((FEY1 - FSY1) * (SY2 - OSY1))) / (OEY1 -
OSY1)

        AEX1 = ((FEX1 * (OEX1 - OSX1)) + ((FEX1 - FSX1) * (EX2 - OEX1))) / (OEX1 -
OSX1)

        AEY1 = ((FEY1 * (OEY1 - OSY1)) + ((FEY1 - FSY1) * (EY2 - OEY1))) / (OEY1 -
OSY1)

        calculated_box = [ASX1, ASY1, AEX1, AEY1]

        crop_img = img[int(ASY1):int(AEY1), int(ASX1):int(AEX1)]

        text_area = reader.readtext(crop_img)

        text_found = ""

        for bbox, text, score in text_area:

                text_found = text_found + " " + text

        print(d["code"] + " -> " + text_found)

        if len(extracted_data):

                extracted_data = extracted_data + "::"

        extracted_data = extracted_data + text_found.strip()
.
.
.
```

All the variables used in the code follow the same pattern of the equations derived in Chapter 3. Thus, each of the fields will be extracted using the main loop in the program. The value of each field can be found in "text_found" variable.

## 4.5    Integrated Code for Document Identification and Data Extraction

Before jumping into the integrated code, all the templates defined for each type of document need to be placed as shown in Figure 4.20.

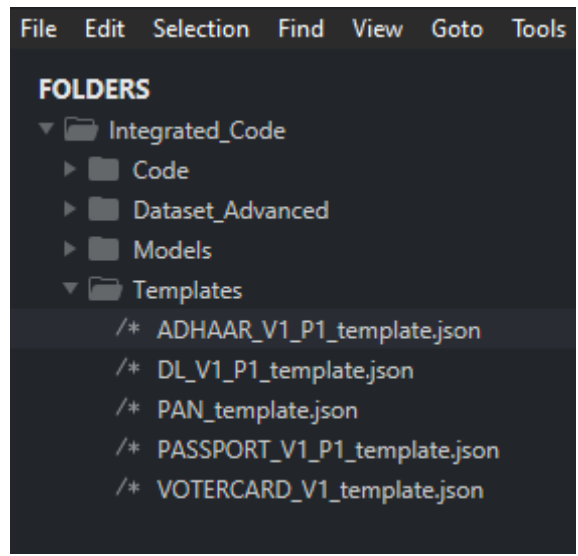

Figure 4.20: Template file location

The following code snippet is the integrated version of the previous two section. So, putting everything altogether:

```python
from ultralytics import YOLO
import numpy as np
import cv2
import easyocr
import matplotlib.pyplot as plt
import json
from difflib import SequenceMatcher
import math

def check_similarity(text1, text2, threshold):
        return SequenceMatcher(None, text1, text2).ratio() >= threshold
```

```python
def get_distance(p, q):
        return math.dist(p, q)

def is_enclosed(p, q):
        return p[0] <= q[0] and p[1] <= q[1] and p[2] >= q[2] and p[3] >= q[3]

image_path = 'PATH\\TO\\TEST\\IMAGE'
data_path = 'PATH\\TO\\TEST\\IMAGE\\ANNOTATION'
# Identification
model = YOLO('PATH\\TO\\best.pt')
results = model(image_path)
result = results[0]
max_class_id = ""
max_cords = []
max_conf = 0.0
for box in result.boxes:
        class_id = result.names[box.cls[0].item()]
        cords = box.xyxy[0].tolist()
        cords = [round(x) for x in cords]
        conf = box.conf[0].item()
        if conf > max_conf:
                max_class_id = class_id
                max_cords = cords
                max_conf = conf
print(max_class_id + " -> " + str(max_conf))
# OCR
img = cv2.imread(image_path)
reader = easyocr.Reader(['en'], gpu=False)
text_detections = reader.readtext(img)
with open("..\\Templates\\" + max_class_id + "_template.json", "r") as f:
    data = json.load(f)
```

```python
f.close()
id_box        =       [data["identifying_region"]["isx"],        data["identifying_region"]["isy"],
data["identifying_region"]["iex"], data["identifying_region"]["iey"]]
id_text = data["identifying_region"]["identifying_text"]
found_id_box = []
for bbox, text, score in text_detections:
    if check_similarity(id_text, text, 0.8):
            found_id_box = [bbox[0][0], bbox[0][1], bbox[2][0], bbox[2][1]]
# EXTRACT DATA
SX1 = data["identifying_region"]["osx"]
SY1 = data["identifying_region"]["osy"]
EX1 = data["identifying_region"]["oex"]
EY1 = data["identifying_region"]["oey"]
OSX1 = data["identifying_region"]["isx"]
OSY1 = data["identifying_region"]["isy"]
OEX1 = data["identifying_region"]["iex"]
OEY1 = data["identifying_region"]["iey"]
FSX1 = found_id_box[0]
FSY1 = found_id_box[1]
FEX1 = found_id_box[2]
FEY1 = found_id_box[3]
extracted_data = ""
for d in data["data_regions"]:
        SX2 = d["osx"]
        SY2 = d["osy"]
        EX2 = d["oex"]
        EY2 = d["oey"]
        ASX1 = ((FSX1 * (OEX1 - OSX1)) + ((FEX1 - FSX1) * (SX2 - OSX1))) / (OEX1 -
OSX1)
        ASY1 = ((FSY1 * (OEY1 - OSY1)) + ((FEY1 - FSY1) * (SY2 - OSY1))) / (OEY1 -
OSY1)
        AEX1 = ((FEX1 * (OEX1 - OSX1)) + ((FEX1 - FSX1) * (EX2 - OEX1))) / (OEX1 -
OSX1)
```

```
        AEY1 = ((FEY1 * (OEY1 - OSY1)) + ((FEY1 - FSY1) * (EY2 - OEY1))) / (OEY1 -
OSY1)

        calculated_box = [ASX1, ASY1, AEX1, AEY1]

        crop_img = img[int(ASY1):int(AEY1), int(ASX1):int(AEX1)]

        text_area = reader.readtext(crop_img)

        text_found = ""

        for bbox, text, score in text_area:

                text_found = text_found + " " + text

        print(d["code"] + " -> " + text_found)

        if len(extracted_data):

                extracted_data = extracted_data + "::"

        extracted_data = extracted_data + text_found.strip()

# open datafile for validation

df = open(data_path, "r")

base_data = df.read()

print(base_data)

print(extracted_data)

print(str(SequenceMatcher(None, base_data, extracted_data).ratio()))
```

The above code integrates Object Detection, Optical Character Recognition (OCR), and text similarity matching to extract using Templatization and validate structured information from an image. Its functionality is further explained in detail below:

1. Helper Functions:

    - check_similarity(text1, text2, threshold):

      It calls the SequenceMatcher function from difflib to compare the similarity between two strings of text. Threshold can be managed externally.

    - get_distance(p, q):

      Euclidean distance between two points p and q.

    - is_enclosed(p, q):

      Returns true if and only if bounding box q is enclosed in bounding box p.

2. Main Image Processing:

- Loads the image and annotation data from the indicated paths image_path, data_path image and data paths

- Object Detection (YOLO):

  i. Loads a pre-trained Yolo model using path to the model's weights.

  ii. Images processed so that all the objects in the image are detected, after processing the results are stored in the variable named as "results".

  iii. Now let's look at each of the bounding boxes of the detected objects and find the one with the highest confidence, max_conf. Then save the corresponding class ID and the bounding box coordinates, max_cords.

  iv. Print the class ID and its confidence score.

3. Optical Character Recognition (OCR)

- easyocr.Reader:

  Loads the OCR model-English language and will extract text from the image.

- Template Loading:

  After identifying an object class max_class_id, a template JSON file defining the structure of identification regions for that class would be loaded.

- Identifying Text Comparison:

  Text comparison with identifying text from the template using check_similarity() function. Since the OCR-detected text is matching with the identifying text of the template, it is getting saved by retrieving its bounding box as found_id_box.

4. Data Extraction:

- Utilize the bounding box of the identifying region for computation of relative positions of other regions in the template.

- The coordinates of the data regions, being defined within the template, are updated according to the relative positions of the detected identifying region.

- All of the calculated data regions are cropped from images with OpenCV and OCR applied to these cropped areas.

- Print and concatenate the extracted text for each region into extracted_data.

5. Verification:

- The text that is obtained from the image is compared with a ground truth file, which is the data path that is expected.

- Finally, the similarity of the extracted text to the ground truth is computed and then printed.

Once everything is in place, the code is run and the output of above code is provided in Figure 4.21.

```
(int_env) C:\Users\meets\OneDrive\Desktop\MS Thesis Work\Integrated_Code\Code>python test_int_v2.py

image 1/1 C:\Users\meets\OneDrive\Desktop\MS Thesis Work\Integrated_Code\Dataset_Advanced\data\images\test\adhaar_v1_p1_91.jpg: 416x640 1 ADHAAR_V1_P1, 203.3ms
Speed: 7.0ms preprocess, 203.3ms inference, 18.0ms postprocess per image at shape (1, 3, 640, 640)
Using CPU. Note: This module is much faster with a GPU.
ADHAAR_V1_P1 -> 0.9879584312438965
NAME -> Sherry Rivers
DATE_OF_BIRTH -> 03/05/2018
GENDER -> Male
ADHAR_NUMBER -> 0000 0000 0091
Sherry Rivers::03/05/2018::Male::0000 0000 0091
Sherry Rivers/03/05/2018::Male::0000 0000 0091
1.0

(int_env) C:\Users\meets\OneDrive\Desktop\MS Thesis Work\Integrated_Code\Code>
```

Figure 4.21: Output of integrated code

The integrated code provided above can only be executed on command line or terminal. But the end user is not concerned with code. The end user needs a URL to interact with the API. So, first order of things is to expose above code as an API for identifying and extracting data from any image.

For this purpose, Flask is an excellent library in python which can help expose the code as API, since the Machine Learning training code is in python, both of them will gel in seamlessly. The code for APIs is as follows:

```python
from flask import Flask, request, jsonify
from service.document_service import identify_document, identify_and_extract_data
from flask_cors import CORS

app = Flask(__name__)
CORS(app)

@app.route('/document/identify', methods=['POST'])
```

```python
def document_identify():
    try:
        data = request.get_json()
        if data is None:
            return jsonify({"error": "Invalid input, JSON expected"}), 400
        response = identify_document(data)
        return response.to_json(), 200
    except Exception as e:
        return jsonify({"error": str(e)}), 500

@app.route('/document/extract/data', methods=['POST'])
def document_identify_extract():
    try:
        data = request.get_json()
        if data is None:
            return jsonify({"error": "Invalid input, JSON expected"}), 400
        response = identify_and_extract_data(data)
        return response.to_json(), 200
    except Exception as e:
        return jsonify({"error": str(e)}), 500

if __name__ == '__main__':
    app.run()
```

The "document_service" here actually contains the integrated code for document identification and data extraction.

A simple React app can help the end users to talk to these APIs exposed. Figure 4.22 and Figure 4.23 show a simple flow that will happen on the browser.

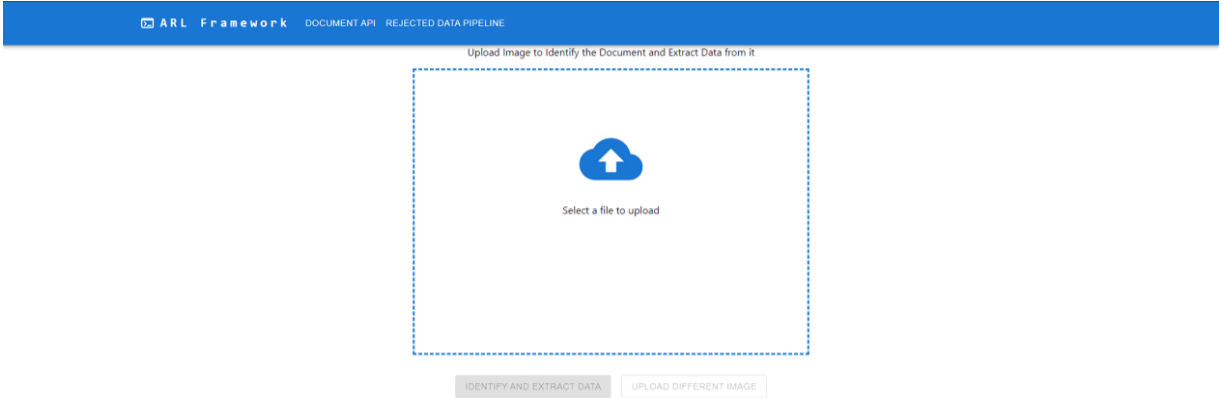

Figure 4.22: Home page of the website

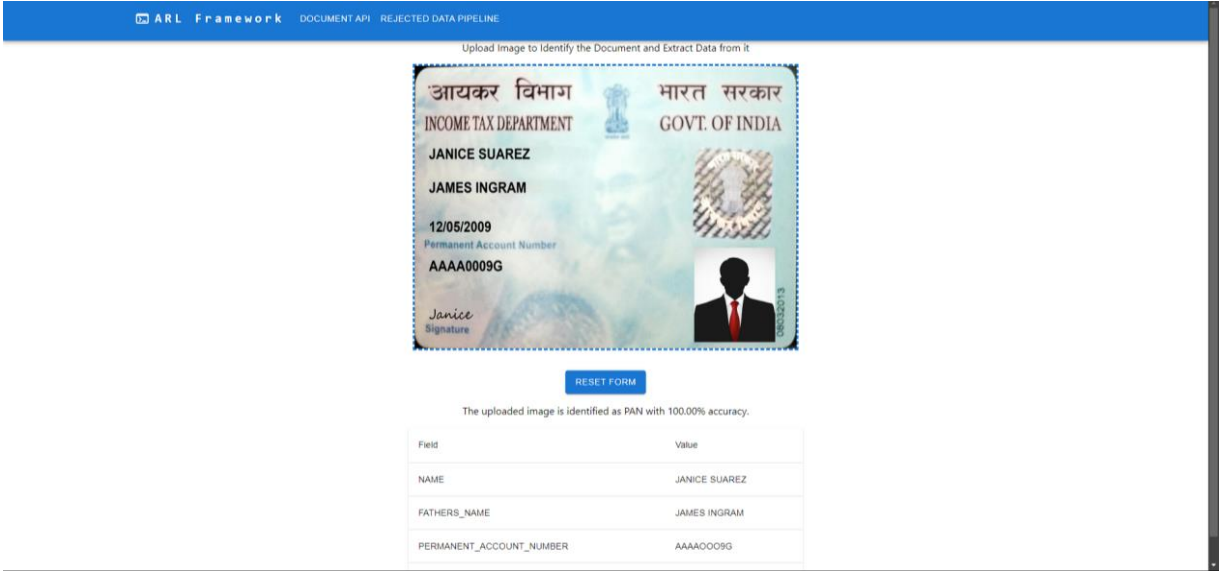

Figure 4.23: Response from API after user uploads an image

Once this is done, it's time to implement ARL Framework in the current flow now.

## 4.6    Implementing ARL Framework

### 4.6.1    External Agent 1

For implementing first part of Augmented Reinforcement Learning Framework, an API need to be expose for users to report any discrepancies with the results. For that matter, a request system needs to be created to make sure all the requests made by end user are registered in the system. In this manner, the end user become the "External Agents 1" in the grand scheme of Augmented Reinforcement Learning Framework. Following code snippet is used to for registering discrepancies:

```python
.
.
.
@app.route('/document/propose/modification', methods=['POST'])
def document_propose_modification():
    try:
        data = request.get_json()
        if data is None:
            return jsonify({"error": "Invalid input, JSON expected"}), 400
        propose_modification(data)
        return jsonify({"message": "Request processed successfully."}), 200
    except Exception as e:
        return jsonify({"error": str(e)}), 500
.
.
.
```

Using the same interface shown in the Figure 4.23, another button is provided for user to report the discrepancies. The updated user interface is shown in Figure 4.24 and Figure 4.25. In the Figure 4.24, it is clear that the image is of Adhaar, but the result from the model is PAN, which is wrong. In such cases, user will report this incident by clicking "Not Happy With Results?" button. In the Figure 4.25, user will select correct Document Type and submits the modification request. The actual request is shown in Figure 4.26. It contains "document_identified", "document_suggested" and "image" in base64 format.

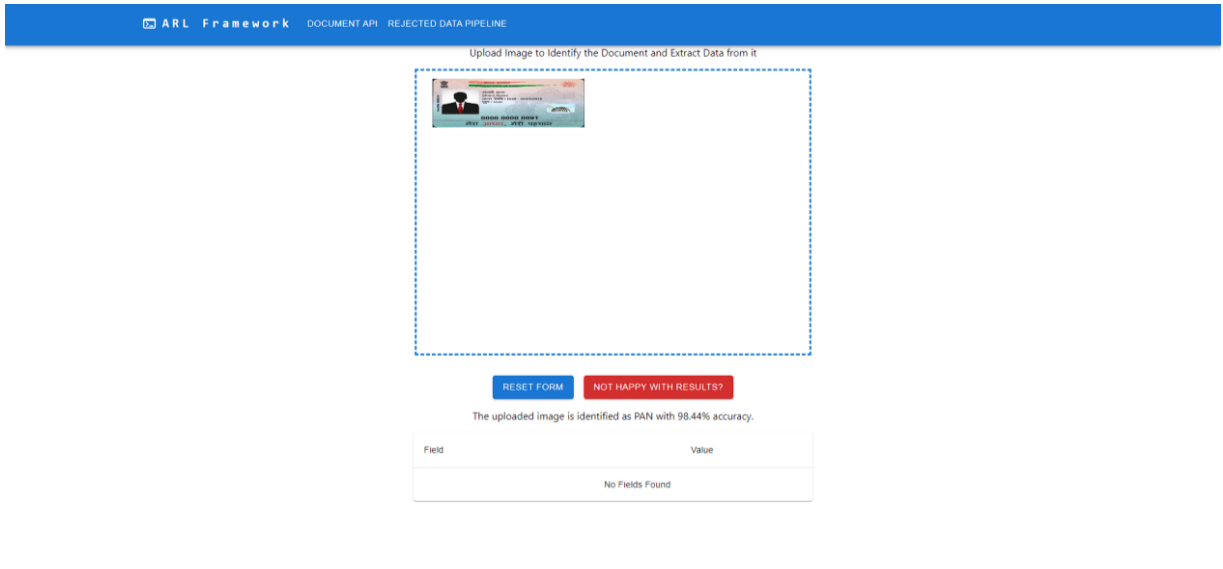

Figure 4.24: Response from API containing erroneous result

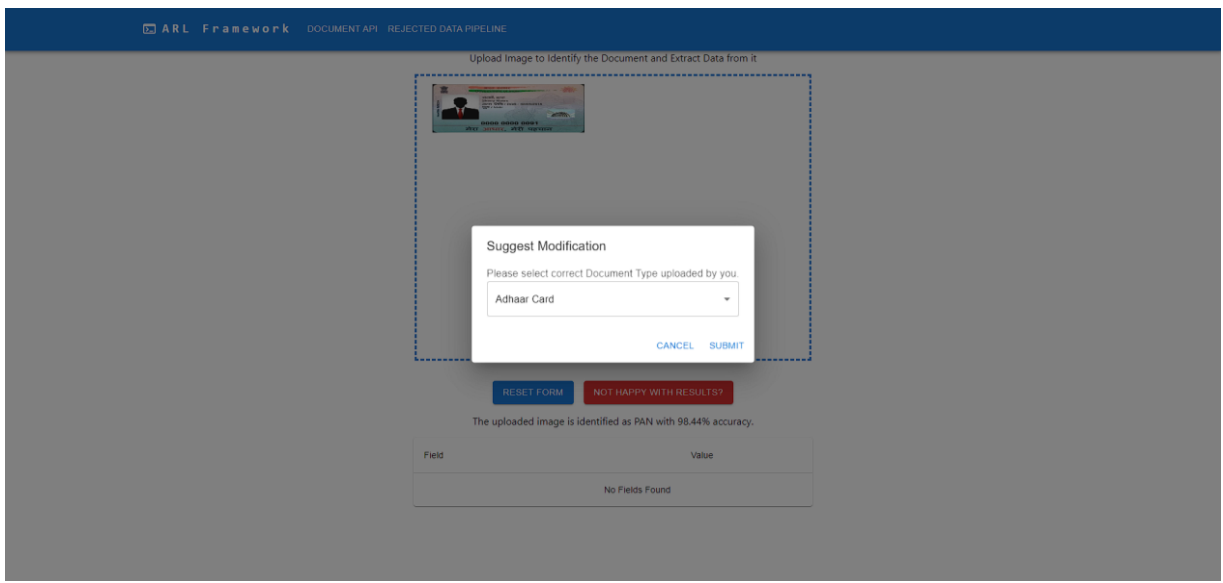

Figure 4.25: User selects updated Document Type and submits the request

```
{
    "document_identified": "DL_V1_P1",
    "document_suggested": "PAN",
    "image": "/9j/4AAQSkZJRgABAQAAAQABAAD/
2wBDAAgGBgcGBQgHBwcJCQgKDBQNDAsLDBkSEw8UHRofHh0aHBwgJC4nICIsIxwcKDcpLDAxNDQ0Hyc5PTgyPC4zNDL/
2wBDAQkJCQwLDBgNDRgyIRwhMjIyMjIyMjIyMjIyMjIyMjIyMjIyMjIyMjIyMjIyMjIyMjIyMjIyMjIyMjIyMjL/
wAARCAH4AyQDASIAAhEBAxEB/8QAHwAAAQUBAQEBAQEAAAAAAAAAAECAwQFBgcICQoL/8QAtRAAAgEDAwIEAwUFBAQAAAF9AQIDAAQRB
RIhMUEGE1FhByJxFBgJxFDKBkaEII0KxwRVS0fAkM3JyggkKFhcYGRolJicoKSo0NTY3ODk6Q0RFRkdISUpTVFVWV1hZWmNkZWZnaGlqc3R1dn
d4eXqDhIWGh4iJipKTlJWWl5iZmqKjpKWmp6ipqrKztLW2t7i5usLDxMXGx8jJytLT1NXW19jZ2uHi4+Tl5ufo6erx8vP09fb3+Pn6/8QAHwEAAwEB
AQEBAQEBAQAAAAAAAAECAwQFBgcICQoL/8QAtREAAgECBAQDBACFBQQAAAJ3AAECAxEEBSExBhJBUQdhcRMiMoEIFEKRob
HBCSMzUvAVYnLRChYkNOEl8RcYGRomJygpKjU2Nzg5OkNERUZHSElKU1RVVldYWVpjZGVmZ2hpanN0dXZ3eHl6goOEhYaHiImKkpOUlZa
XmJmaoqOkpaanqKmqsrO0tba3uLm6wsPExcbHyMnK0tPU1dbX2Nna4uPk5ebn6Onq8vP09fb3+Pn6/9oADAMBAAIRAxEAPwDx2e41WdlV
yAMY4Hp7j60iTzuwRXJYkAAKMk+2Bz3pWjaW9KIMuxAUD1wK9l88l188yw51ltZ9SWpyk2GSC2kgp1pQCO9Z31e5poaNw IKn.FGI6Sd7b
O1+DEzIPtWshW77ExXrEUSRRhI1CquAABjipQCO9Z31e5pyo8uX4L3k23BH1qqPsrPsH8qePgtY4wdZuOvURj/
GvUBS49aLyCyPMR8Ft06HV7r8EHX86ePgvpYwTqT0ef7o/oa9M4/yaXA46Yp3Ywr5oPgzpHfUrvH0/wDr07/
hTWi4Sv7zPqP8K8K9v7zPq7BZHmy/B7Q9wWry8z3Ib+makHwc8DvdXvT+/XoEg24Ydc81KrAqDSv1SSPOz8HvD+Mi4vP++6WP4QeHGUE
zX2en+S716Gc6Y6VEnyyfSOucUcz7sjg/wDhT3hv/rf/wDF7r+FFhf4Q+G0533xH/Xbt+Veh0yQEQB064ovU60yOEX4Qe12bkj3/eD/
AIU4fCDwuOovjz/z8H+ddxCxeE1RZz8hhddxCxgEc1NRzMVkeen4SeF0KAP24g24t+twtS4aReVFQeLV7/28mu2nUEAkcggipV5UE8UuZ33HZHCj4ReFOcx3v
/gU1M/4VN4VEm0xXv/pTf0nrvsAdkjkGJA2O9F2FFcR/wAKk8KdRDe/jdNR/wAKj8K8K8j9K9F9TCkfvlk/OQ7vr7yf/Apq7wY7c0cZ8d/Bad2FPc4APhKvhL4VEpUxpZmsMKH33Cyu0/
SW/p1qT/hUfhT/nheZ/6+m/pXbOMTA9/rUtCk+4+4JdVw/CvV8jivRvwqbkhfUj9HIE3kYm+Xpm/dy/
4U34UOfm1AD2uZN4tMh2fof/7tCk/wUWvADqP8061K184+6q2p2q2I0Djq/9g2xXy2/N1W/t6DH3AKmAJ8U1AD2uZN4tMh2fof/
4U34UOfm1AD2uZP2PKfkNAbNYQ+k/wCKeJ 8YABf47j7Sf8K8i/Dp+H8LeYABf47j7 Sf8KyFq/wAKcE8M8+k/wCKeHxY/x/9Y3cN/H+H/H/
byf8KP+PeFuTm/6/88Ap/wen5Ve6CBmjFPmfcI87I87xD3gXprQYXcycf/wB//8ojPq6uqrXKdDIjRr8Xff/
CmfDB6S6gPP8A4ikf4f4iqVTuLI1a1f9t+v5JiJeRvwaqHkC3Pqc/sWbBodRrxsofF/H/pL8vs/
SnP4S6ehkh fUj3GQqiMY8d6LkuFfkf/TJlw/v5JkVQuiLOm+Q9/Cwt/ogrq/+DoVcK35EMgE9bgLqQwe+J+/5U7/ AIUxx6V97W/
8CPiBA1ehjOOnandaOZ2HZXPOH+DXdqO7T5moVwf88Eenp9Sq3/Xfjd_GvdunIEG7s9eRaK8SmlHDel68A+97tK1Ddf2fVdBYoGTe
```

Figure 4.26: Modification Request sample

### 4.6.2 External Agent 2

In Augmented Reinforcement Learning Framework, the main function of External Agent 2 is to segregate false positive, noise data. This will strengthen the Rejected Data Pipeline, that will in turn help in further retraining of the Machine Learning model. Naturally, the idea is simple. Show the list of all the request to the Admin or Approver (External Agent 2), and let him decide what is scenarios from the modification requests, best fulfil his Business scenario.

The code for gathering all the requests, rejecting a request and approving a request is as follows:

```python
.

.

.
@app.route('/document/request/getAll', methods=['POST'])
def document_requests():
    try:
        mypath = './modification_requests'
        only_files = [f for f in listdir(mypath) if isfile(join(mypath, f))]
        arr = []
        for file_name in only_files:
            f = open(mypath + '/' + file_name, "r")
            json_object = json.loads(f.read())
            arr.append(json_object)
        return jsonify(arr), 200
    except Exception as e:
        return jsonify({"error": str(e)}), 500

@app.route('/document/request/reject', methods=['POST'])
def document_request_reject():
    try:
        data = request.get_json()
        os.remove("./modification_requests/request_" + str(data['req_id']) + ".txt")
        return jsonify({"message": "Request processed successfully."}), 200
    except Exception as e:
        return jsonify({"error": str(e)}), 500
```

```python
@app.route('/document/request/approve', methods=['POST'])
def document_request_approve():
    try:
        data = request.get_json()
        approve_request(data)
        os.remove("./modification_requests/request_" + str(data['req_id']) + ".txt")
        return jsonify({"message": "Request processed successfully."}), 200
    except Exception as e:
        return jsonify({"error": str(e)}), 500
.
.
.
```

Since all the requests are saved in form of a file, each of the request can be processed separately. For each of the request approval, a single image is created and put in the Rejected Data Pipeline. Figure 4.27 shows the list of accumulated modification request in the system by different users.

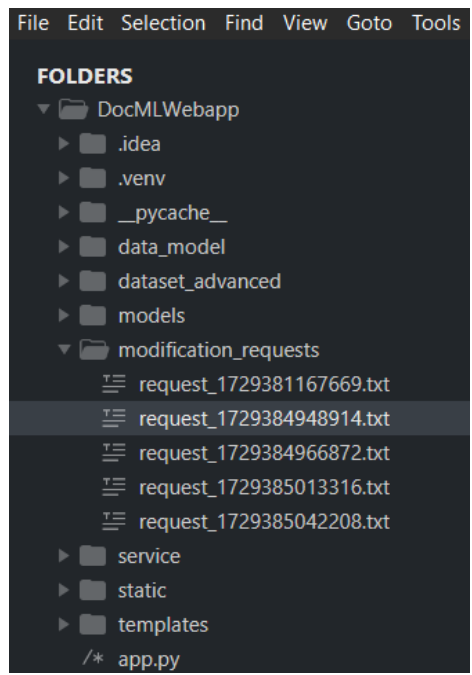

Figure 4.27: Modification Requests

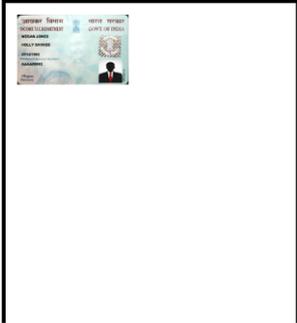

Figure 4.28: Modification Requests on UI

As per the Figure 4.28, the External Agent 2 can see all the modification requests made by External Agent 1 on the website. According to the correct business requirements, the External Agent 2 can Approve or Reject the requests. On rejection, the data is discarded. While on approval, the data is pushed in Rejected Data Pipeline which is further discussed in next section.

### 4.6.3   Rejected Data Pipeline, Augmentation and Retraining

Last and the most important part of Augmented Reinforcement Learning is feeding data from Rejected Data Pipeline and retraining the model. And this step is pretty straight forward. The new updated dataset for next training cycle is created from two sources:

1. Initial dataset used for training

2. Dataset from Rejected Data Pipeline

If a new directory is provided for Rejected Data Pipeline, a simple copy paste of all files will be required. Else, direct folder path can also be provided while processing the request approval in previous section. Both the ways are equally acceptable. But if something goes wrong, there is no going back in second method.

Figure 4.28 shows the first way of creating the Rejected Data Pipeline. The folder "dataset_advanced" is used to store all the approved requests from the Admin or Approver (External Agent 2). If needed, Data Preprocessing steps can also be performed. This step achieves Rejected Data Augmentation.

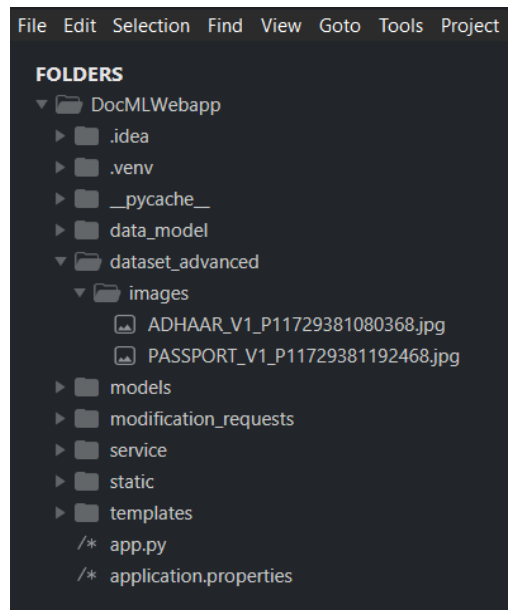

Figure 4.29: Rejected Data Pipeline

Once this is done, all the steps provided in section 4.4, section 4.5 and section 4.6 need to be repeated to retrain updated Machine Learning model and generate updated "best.pt" weight file. This completes all the steps discussed in Figure 3.2.

## 4.7     Summary

To integrate the ARL framework, first, a proper comprehensive dataset needs to be generated that will form the base for training their machine learning model. Classification of documents is done with proper categories suitable for identification along with key fields specific to extraction purposes as well. This gives a proper framework not only to train a model but also to evaluate them.

Following the data set creation, common preprocessing on the data is performed to ensure they are cleaned and standardized. This is critical to guarantee that the machine learning model learns cover vast range of scenarios. These stages are thus vital for noise reduction and overall enhancement of the quality of input to permit meaningful patterns being learned from by the machine learning model. Then dataset was further divided into training, validation, and testing subsets. These subsets were further used in training, fine-tuning, and evaluating the model. A 7:2:1 split is normal in the process. The training shall ensure that the model has enough data to learn from, while part of it shall not be seen during training to ensure generalization testing capabilities. The document categories and key fields are equally represented in every subset so that bias is not introduced during the training process of the model.

The main goal is to train a machine learning model to classify a multitude of document types. In search of optimal architectures, CNNs for image-based documents and transformers for the text-heavy files are explored. The metrics like accuracy, precision, and recall of the identified category of the document are watched during the training process in order to not learn sub-optimally. Algorithms like grid search are utilized to perform hyperparameter tuning. This further optimizes the performance of the model by fine-tuning with the following hyperparameters: learning rate, batch size, no. of epochs, and network architecture for best results.

The folder structure is managed carefully to improve ease in manipulation of the data and smooth training of the models. The documents are categorized with labels and split into folders which further divide them into subfolders for training, validation, and testing. This will correctly ensure the data used at every stage within the machine learning pipeline and avoid leakage, thereby ensuring reproducibility of model training. The metadata like annotations and labels were stored in a separate directory for easy access in the preprocessing as well as training phases. The actual training is done using YOLO, the most famous machine learning library, where the training program loads the dataset, applies a batch of data preprocessing into the model, and uses predefined hyperparameters. Logging mechanisms track loss, accuracy, and time during training; hence, performance can be monitored, and overfitting can be avoided by early stopping.

Data Extraction employs a template-based approach. Once a model for the document type is identified, a template is applied that will get the relevant fields from the document. This templatization process ensures that the extraction of the same is highly accurate and efficient in terms of procedure; otherwise, it's too vast, and the system becomes adaptable to more than one document type. The extraction process can then be integrated into the overall document identification pipeline, thus ensuring seamless end-to-end automation from classification to data extraction.

The system adopts a significant innovation of this thesis called Augmented Reinforcement Learning (ARL) framework, designed to help improve decision-making. The ARL framework integrates human experts as external agents in the document classification and data extraction workflow. It provides the model with feedback whenever it is less confident or encounters an error. External Agent 1 continues the review of the error that has caused incorrect classification or improper extraction and inputs it back into the model during the next training cycle. External Agent 2 is filtering out those cases which are not suitable for business scenarios.

This system also makes use of a rejected data pipeline for misclassified or extractions. The rejected data, which are instances that have been either misclassified or poorly extracted, go back into the retraining loop. This way, the model will always continue to improve with the task since it learns from mistakes. Misclassified documents are re-labeled and go back to the training set, giving the model an opportunity over time to develop a greater ability to handle difficult or ambiguous cases.

In short, this implementation is comprised of the much-needed aspects in a system that includes preparing datasets, pre-processing, model training, extraction of data, and the ARL framework that is supported by external agents to enhance decision making. Through the rejected data pipeline, which gives the system leeway to mend its mistakes, the system's ability to improve continuously gets the system to yield a robust and adaptive model that can indeed classify documents and extract data accurately.

# CHAPTER 5: RESULTS AND DISCUSSIONS

## 5.1    Introduction

This chapter provides a comprehensive summary and interpretation of the results obtained when implementing an end-to-end solution to the problem statement "Document Identification and Data Extraction". The focus is on how the introduction of an Augmented Reinforcement Learning framework affects an Object Detection Machine Learning model training cycle. Changes in the model's decision-making capabilities after introduction of this framework are discussed, considering the two phases-training and testing.

When an ARL framework is introduced in learning, human agents or some external agents are used to improve the model's decision-making. This enhances the ability of models in comparison to other cases to work with uncertainties and to make better decisions as well. The model of traditional reinforcement learning is not adaptable to the complexity of real-world complexities, especially when exposed to novel data or ambiguous data in most scenarios. However, the ARL framework allows the introduction of additional external agents' guidance, making the model more effective in identifying documents and retrieving relevant data.

The accuracy and precision of the enhanced model employing ARL over typical reinforcement models are much greater than the standard reinforcement learning models. If the model already encountered previously unidentified or complex document structures, then the improvement of the model is more significant. Actually, these decisions that may eventually result in mistakes or suboptimal outcomes are corrected or adjusted by the expertise of a human or any external agent of the enhanced model.

Secondly, the chapter elaborates on the more general relevance of the ARL framework. Beyond identifying documents and extracting data, the potential of the ARL unfolds towards facing various real challenges, in which provision for decision making under uncertainty greatly features. Applications can therefore be foreseen across fields like autonomous driving, healthcare diagnostics, and robotics where human expertise is supported by learning from machines to add stronger 'decision-making capabilities'.

## 5.2    Results of Model Training without ARL Framework

The object detection model was trained using YOLO over 16 epochs and 16 batch size with 5 classes and displays important trends with regard to the training loss, validation loss, and key metrics like precision, recall, and mAP (Mean Average Precision). Since the quantum of data for epochs always have a trade-off, hence the number of epochs is kept at 16.

### 5.2.1    Training Loss

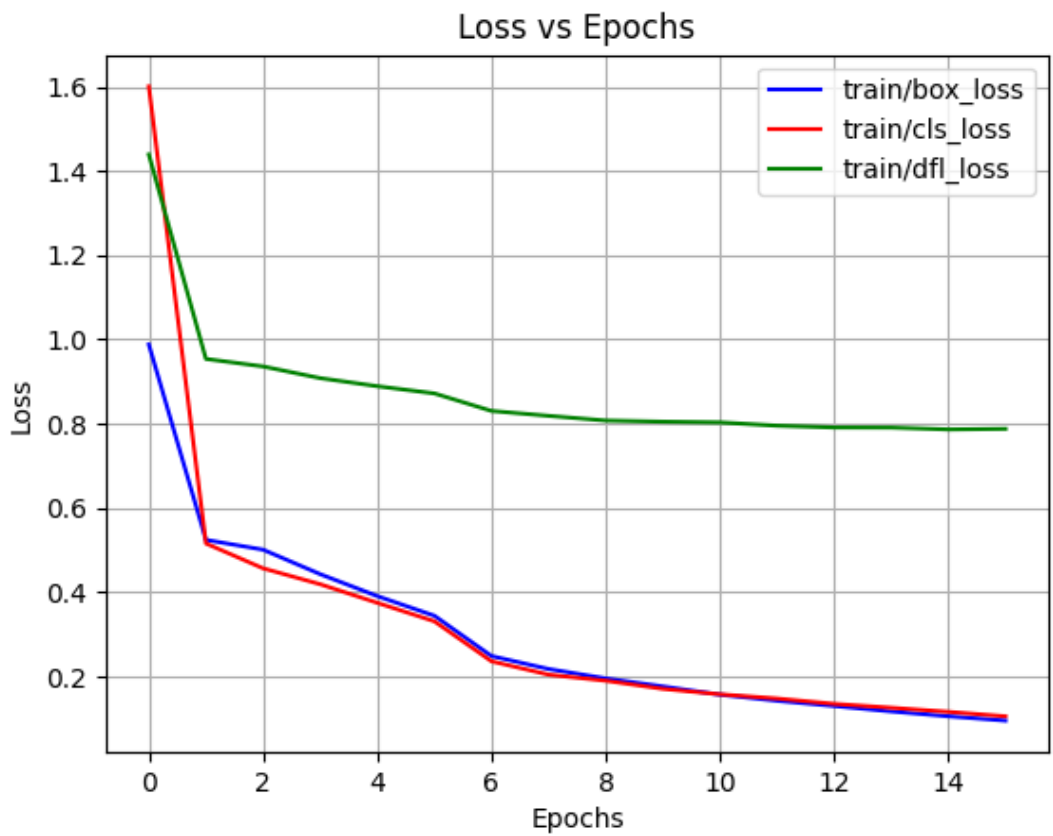

Figure 5.1: Loss vs Epochs

The Figure 5.1 shows the trend of Loss versus Epochs. Key observations are:

1. **Box Loss:**

   The model begins at epoch 0 with a box loss of 0.98817 and continues learning and reduces it down to 0.09529 at epoch 15. This establishes the fact that with the increase in the training period, the model becomes better in localization error and the bounding box prediction error declines.

2. **Class Loss:**

   It begins at 1.6007 and drops steadily to a final epoch value of 0.10546. The decrease shows that the model becomes more confident and accurate in its object classification within the bounding boxes.

3. **DFL (Distribution Focal Loss):**

   The DFL loss, which was 1.4391 at the beginning, reduced down to 0.78789 by epoch 15. The reduction in DFL shows that the model confidence regarding the distribution of the predicted bounding boxes increases with training.

The losses above collectively demonstrate an overall improvement in the process of training where the model is better fit to the data as epochs increase.

### 5.2.2   Key Metrics

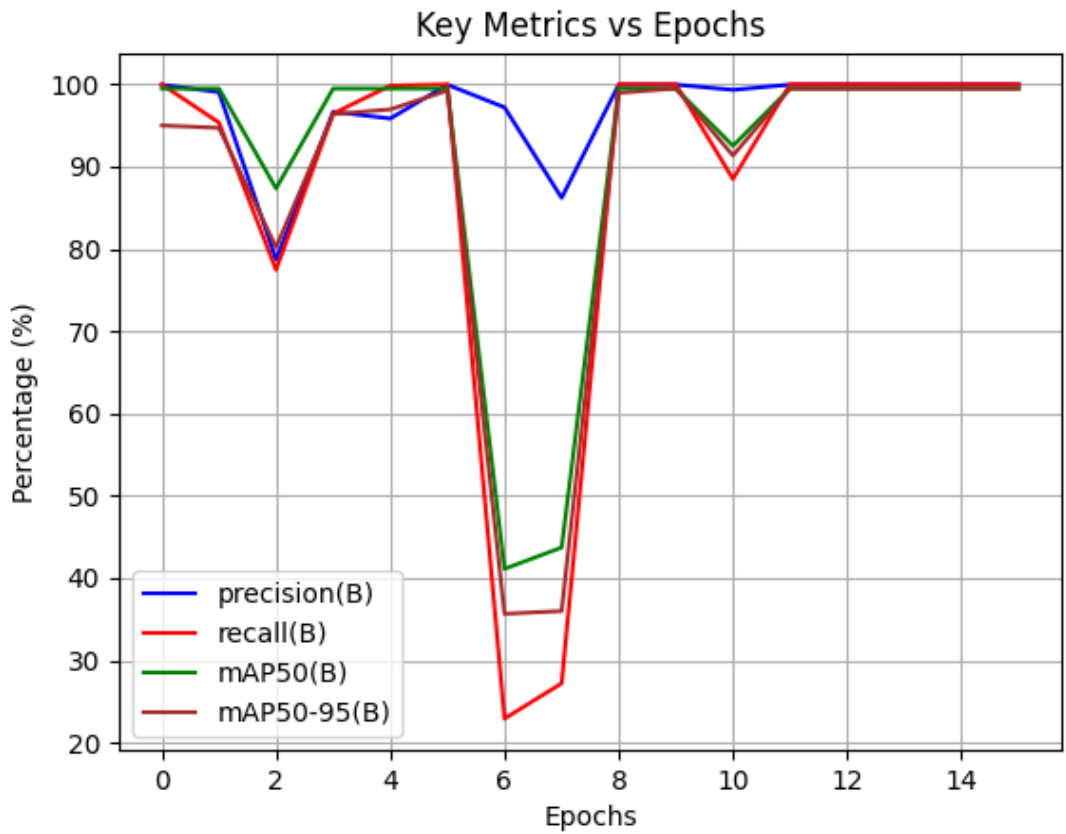

Figure 5.2: Key Metrics vs Epochs

The Figure 5.2 shows the trend of Key Metrics versus Epochs. Some key observations are:

1. **Precision:**

   The precision starts at epoch 0 at an incredibly high 0.99952 and remains supremely good for all epochs. This reflects the model's predictive accuracy when it comes to the presence of a positive sample, with very few false positives.

2. **Recall:**

   The recall starts at 1.0, though fairly similar across epochs. This reflects that the model is good in identifying most of its objects while false negatives are minimal.

3. **mAP50 (Mean Average Precision at IoU 50):**

This curve starts off from 0.995 and remains mostly steady for most of the training process. This really implies the model has a high precision at IoU 50, hence is strong in object detection when the IoU threshold lowers.

4. **mAP50-95:**

This metric evaluate performance over various IoU thresholds, and the curve begins from 0.95015 and reaches at maximum around 0.995 by final epoch. As a result, it depicts that improvement of multiple IoU thresholding for the performance of the model is established with regard to time.

These metrics show that the model can get very high detection accuracy with both precision and recall very high, such that the model classifies objects with errors to a minimum.

The Figure 5.3 show the F1 curve, based on Precision and Recall in Figure 5.2.

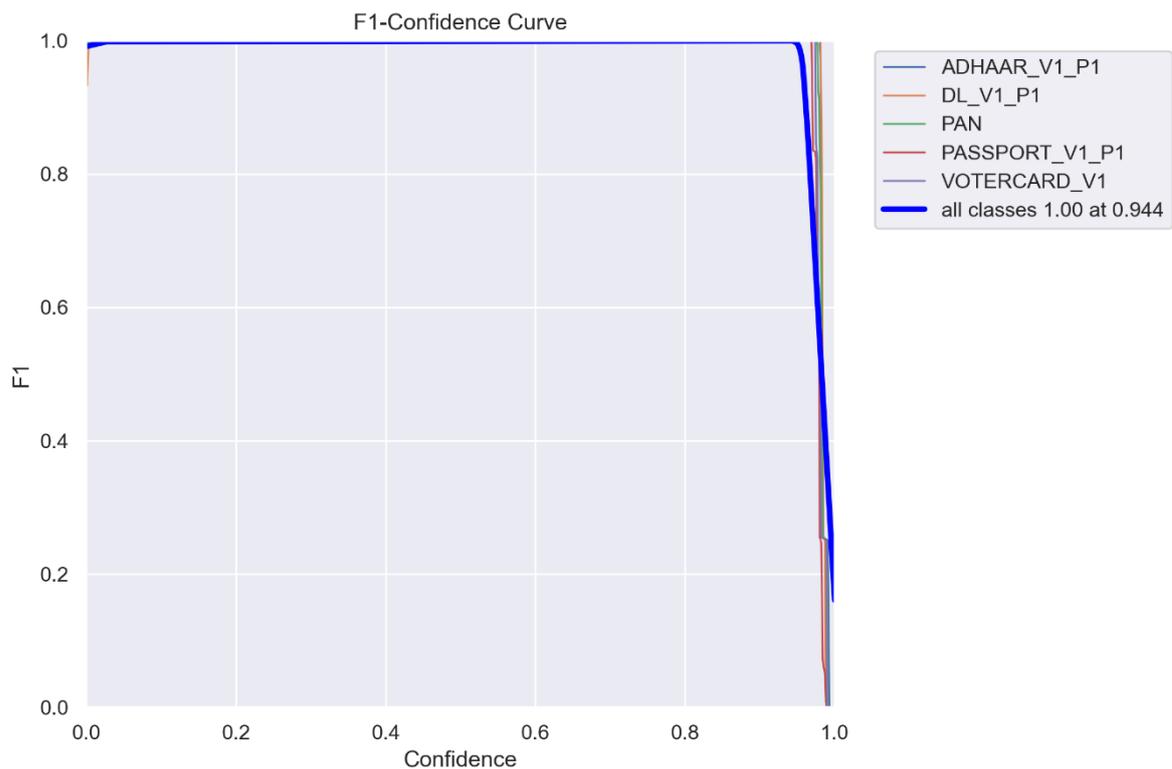

Figure 5.3: F1-Confidence curve

### 5.2.3   Validation Loss

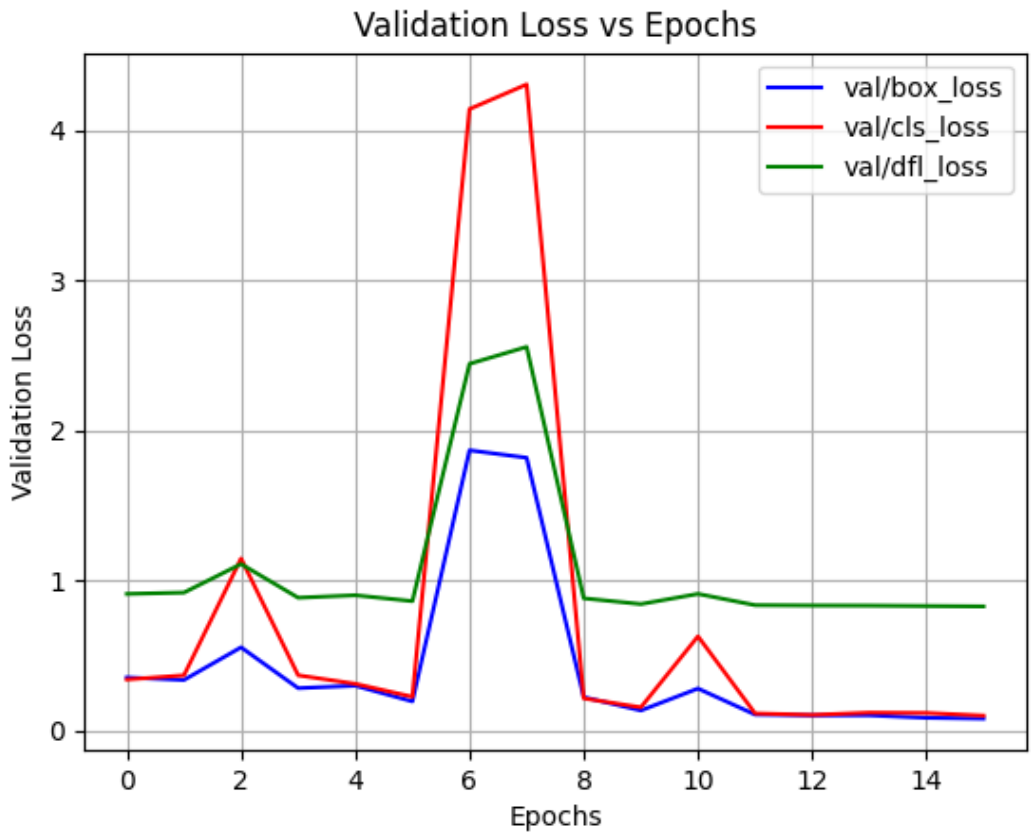

Figure 5.4: Validation Loss vs Epochs

The Figure 5.4 shows the trend of Validation Loss versus Epochs.

1. **Validation Box Loss:**

   The validation box loss starts from 0.35562 in epoch 0 but deviates over training time with a final value of 0.08138. Even though it is still decreasing, one would note the spikes in the range of epochs 6-7 to almost reach 1.87, which may speak of overfitting or variance in the way the model generalizes towards the validation set.

2. **Validation Class Loss:**

   The validation class loss swings between 0.34227 to, after 15 epochs, 0.10117. The peak in epochs 6 and 7 may represent a short period when the model got confused for some few epochs on classification accuracy on the validation set.

3. **Validation DFL:**

The validation DFL starts from 0.91286 and comes down to 0.82955, and like other losses, it does have fluctuations during epochs 6-7.

The oscillations in the loss values across epochs 6 and 7 may be indicative of overfitting; the model begins to memorize the training data but fails to generalize well onto unseen data. The later troughs of these epochs indicate how the learning rate adjustments helped recover the model and further improved it.

### 5.2.4 Learning Rate

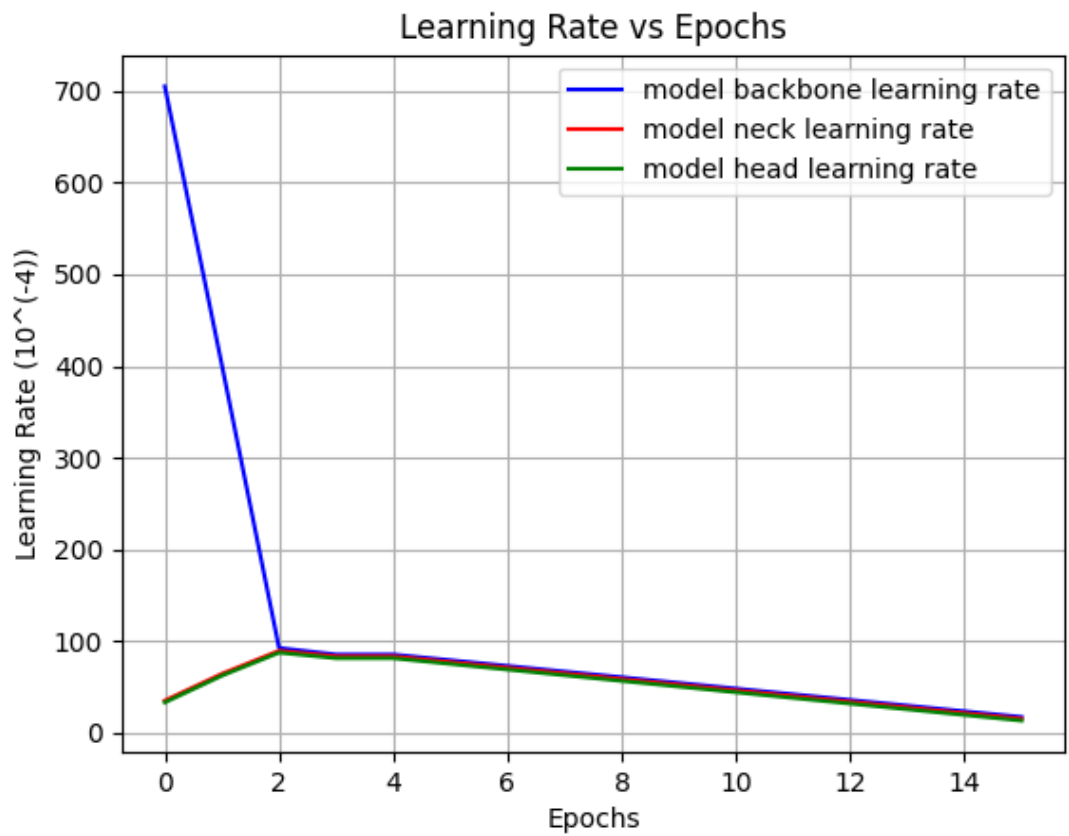

Figure 5.5: Learning Rate vs Epochs

The Figure 5.5 shows the trend of Learning Rate versus Epochs.

This learning rate decays progressively from the initial epoch 0 value of 0.070098 to the final epoch 15 value of 0.0013375; such a decay is usual in deep learning training schedules and fine-tune a model, allowing it to settle into a local minimum as the weights are adjusted in finer granularity.

### 5.2.5  Test Dataset

Recall that the full dataset was split 3 ways namely – "train", "validation", and "test" dataset. Although, on paper, the model seems to perform fine. But only after testing on the "test" dataset, the actual picture can be discovered.

Table 5.1: Test Dataset results without ARL Framework

| | |
|---|---|
| Total Document Types | 5 |
| Number of Test Images | 10000 |
| True Positives (Identified Correctly) | 6000 |

The Table 5.1 indicates that there are some scenarios which is are still problematic for the model. Although, more training may be performed on the same dataset. But that may lead to more overfitting rather than resolving the cause. This is where Augmented Reinforcement Learning Framework might help in handling those boundary scenarios.

## 5.3    Results of Model  Training with ARL Framework

### 5.3.1  Training Loss

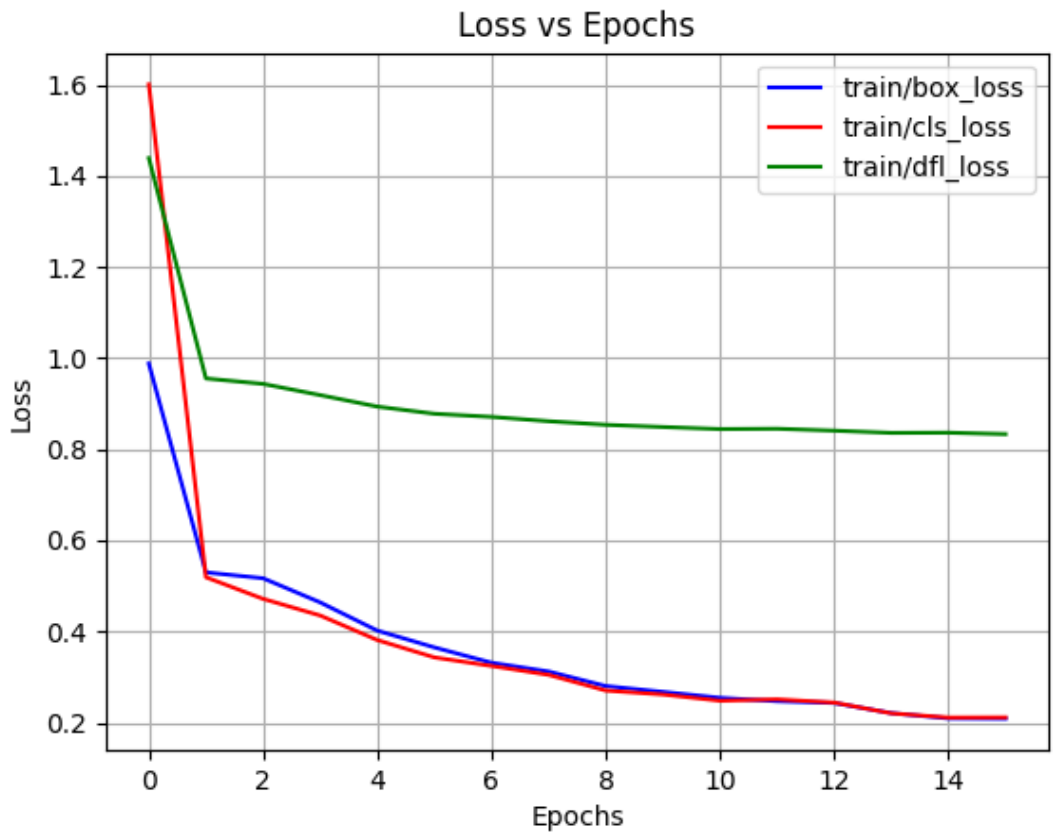

Figure 5.6: Loss vs Epochs

The Figure 5.6 shows the trend of Loss versus Epochs. The trend to is similar to previously trained model. Key observations are:

1. **Box Loss:**

   It started high at epoch 1 as 0.98817 and eventually decayed during the training into 0.20884 at epoch 15. This decline shows that the model is learning progressively to better localize bounding boxes over objects, hence reducing its error in predicting their locations.

2. **Class Loss:**

Can be observed from 1.6007 and initially fails to classify correctly the objects. The classification loss reduces sharply to 0.21151 at epoch 15. The rate at which it has reduced is such that it indicates that the model can assign the correct class to the identified objects with considerable improvements in accuracy levels.

3. **DFL (Distribution Focal Loss):**

It begins at 1.4391 at epoch 0 and gradually decreases down to 0.83324 at epoch 15. This metric follows the concept of monitoring that the confidence model has on bounding boxes predicted. A steady decrease signifies that predicted distributions are more aligned with the actual boundaries of objects.

### 5.3.2 Key Metrics

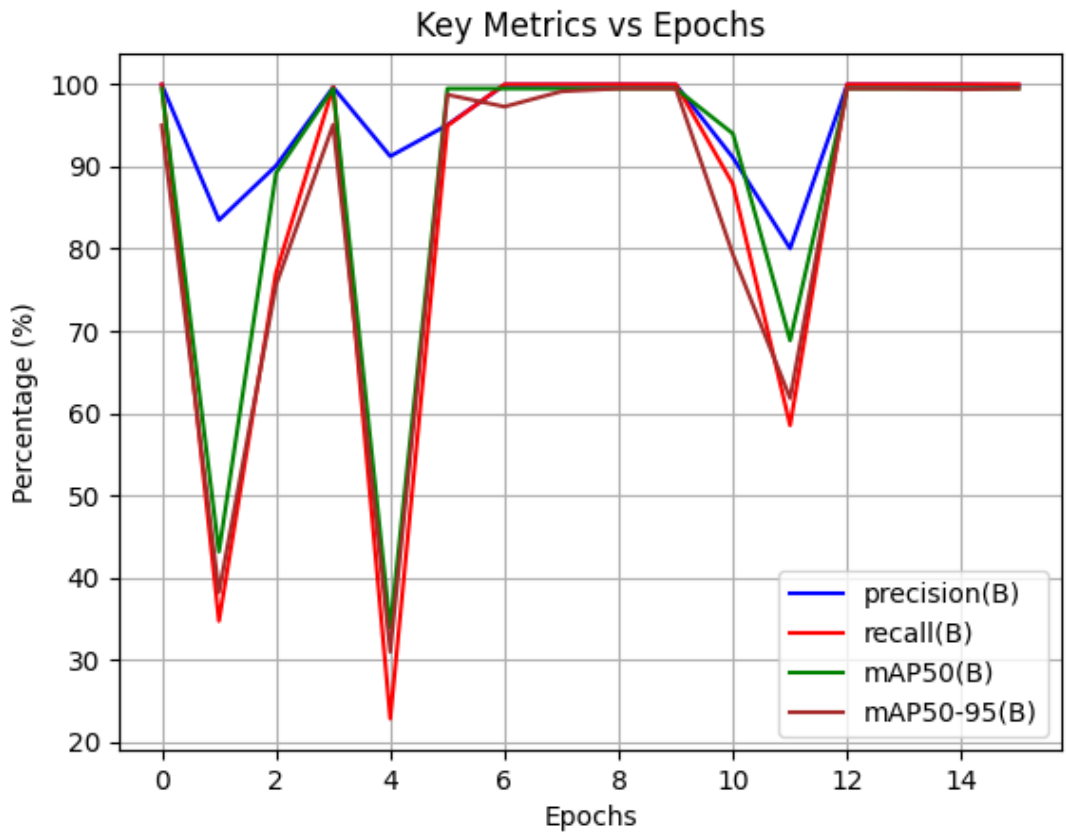

Figure 5.7: Key Metrics vs Epochs

The Figure 5.7 shows the trend of Key Metrics versus Epochs. Some key observations are:

1. **Precision:**

   The precision accuracy score is very high at epoch 0: it starts from 0.99952 and remains fairly stable at value 1 in most of the epochs, showing that the model is very good at minimizing false positives.

2. **Recall:**

   Having started at a value of 1 in epoch 0, recall drops to 0.34707 in epoch 1 but recovers as it goes along. It sways during the training but settles at a stable value of 1 by epoch 15. The fluctuation in the recall of the early epochs could be due to the fact that initial epochs lack some detections, yet the improvement indicates that later epochs are well covered with respect to the true objects.

3. **mAP50 (Mean Average Precision at IoU 50):**

This curve starts from 0.995 in epoch 0 and stays that way till it gets stabilized at 0.995 by epoch 15.

4. **mAP50-95:**

A stricter metric that considers multiple IoU thresholds begins at 0.95015 and reaches up to 0.995. The strong performance on this metric reflects the accuracy of the model across the different thresholds.

The Figure 5.8 show the F1 curve, based on Precision and Recall in Figure 5.7.

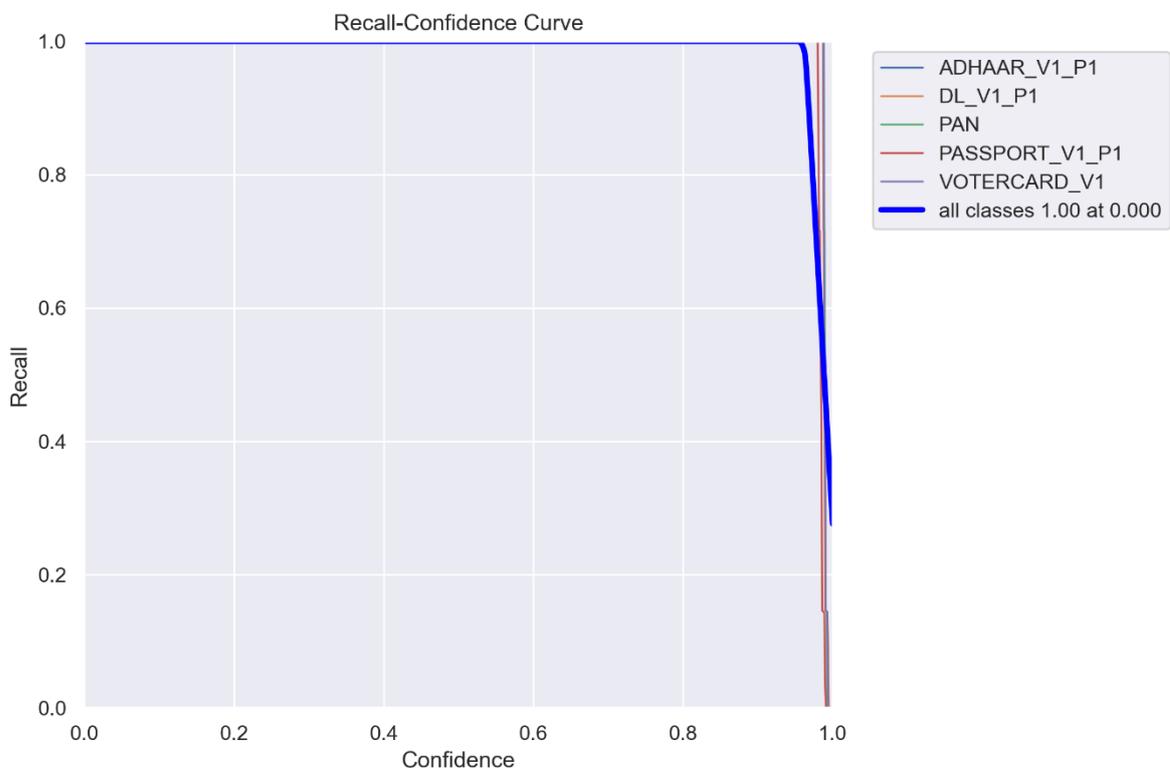

Figure 5.8: F1-Confidence curve

### 5.3.3 Validation Loss

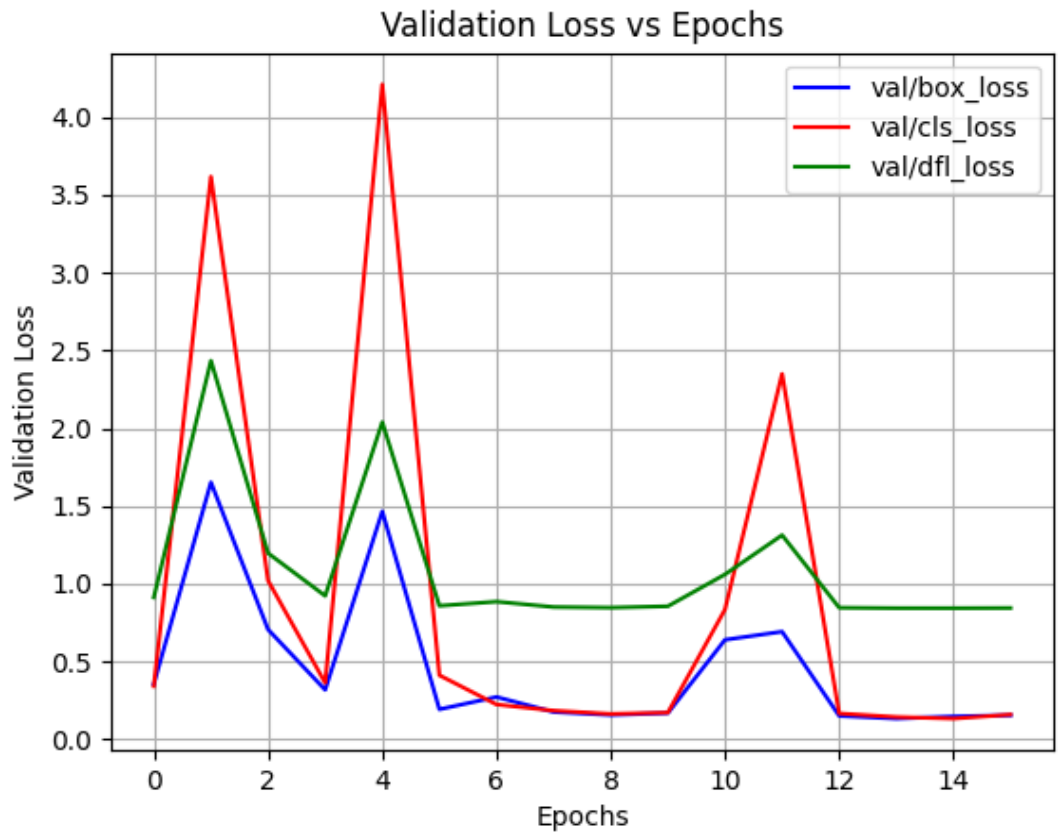

Figure 5.9: Validation Loss vs Epochs

Validation losses cross-check for overfitting and the model's generalization. The Figure 5.9 shows the trend of Validation Loss versus Epochs.

1. **Validation Box Loss:**

   Validation box loss begins at 0.35562 and shows some degree of fluctuation during training, peaking to a peak at epoch 1 as 1.652 and then settling at a lower value of 0.1552 at epoch 15. There is considerable fluctuation during early epochs, suggesting the fact that the model is not yet generalizing well. However, the general trend is declining, which it actually means that learning is going on.

2. **Validation Class Loss:**

The validation set classification loss in the early epochs is more from 0.34227 and similarly fluctuates before stabilizing at the last epoch to as low as 0.15977.

3. **Validation DFL:**

The validation DFL starts from 0.91286 and comes down to 0.84452 and hence shows that the predictions are more confident, but like other losses, it does have fluctuations.

### 5.3.4 Learning Rate

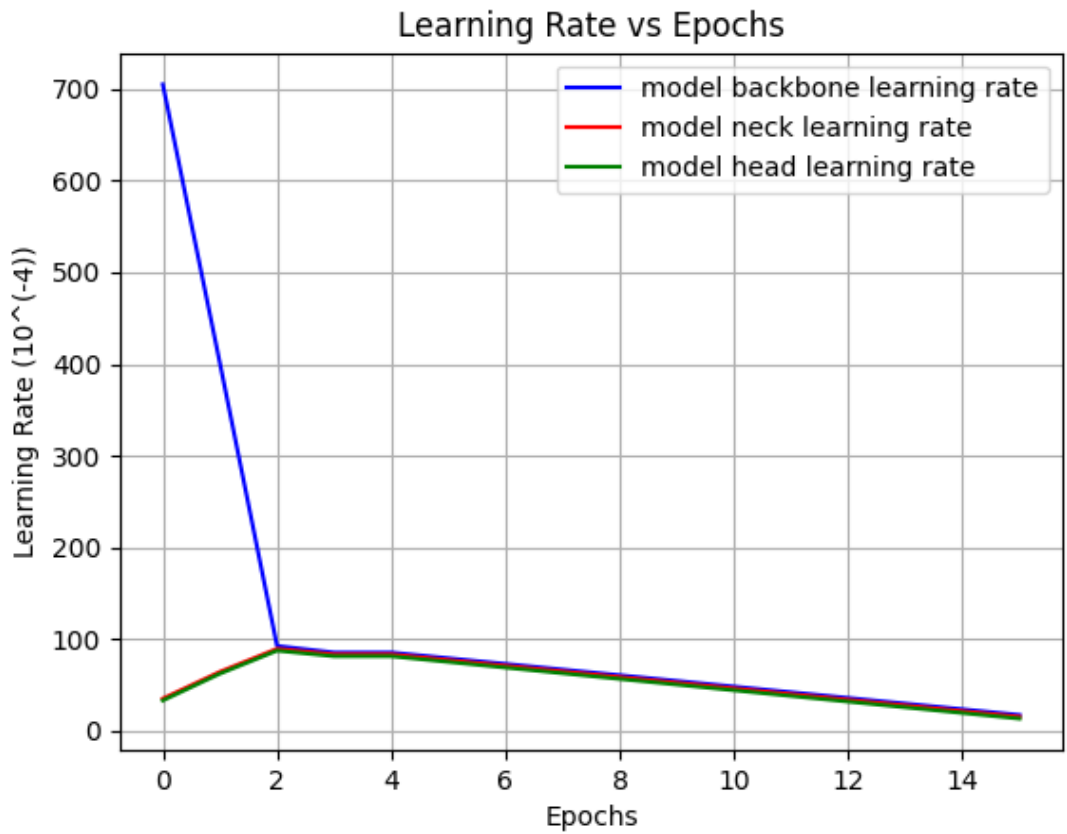

Figure 5.10: Learning Rate vs Epochs

The Figure 5.10 shows the trend of Learning Rate versus Epochs. The trend is same as the previously trained model.

The three parameter groups pg0 (backbone), pg1 (neck), and pg2(head) have their learning rate gradually reduced throughout the training course. This is usual in YOLO training: In the later epochs, it fine-tunes the parameters without overcorrecting itself because the learning rates are reduced. Learning Rate (pg0) starts at 0.070098 in epoch 0; reduced by epoch 15 to 0.008317. Learning rate (pg1 and pg2) starts from 0.0033225 and declines approximately along the same curve, and at 0.008317.

### 5.3.5 Test Dataset

In this scenario, we now have increased size of training dataset. This is due to the inclusion of Rejected Data Pipeline. The rest of the parameters remain the same. As it is evident from Figure 5.6, Figure 5.8 and Figure 5.10, the trend of training loss, F1 Confidence Curve and learning rate remain similar to previous training.

But Figure 5.8 and Figure 5.10 show varied fluctuations in the trend, but they get rectified in subsequent epoch same as before. But again, the real picture will be discovered only after performing test on "test" dataset.

The Table 5.2 indicates that the boundary scenarios area also now covered. But this definitely doesn't convey the message that the model after this is ready for "any" challenge thrown at it. This is a continuous process of learning from past mistakes, with external agents nudging the model in the right direction.

Table 5.2: Test Dataset results with ARL Framework

| | |
|---|---|
| Total Document Types | 5 |
| Number of Test Images | 10000 |
| True Positives (Identified Correctly) | 10000 |

## 5.4    Model Comparison with and without ARL Framework

The effectiveness of the Augmented Reinforcement Learning  Framework is demonstrated with a detailed comparison between the model trained with and without this framework. The experimental setup uses following:

1. Dataset: The dataset uses 10,000 images, equally divided among 5 different document types. Images used here mimic real-world conditions.

2. Metrics:

   - Accuracy: The percentage of correctly classified images and successfully extracted information.

   - Precision: How accurately the model identifies relevant information without including irrelevant data.

   - Recall: The ability of the model to capture all relevant information.

   - F1 Score: A harmonic mean of precision and recall to balance the trade-off between them.

3. Models Evaluated:

   a. Without ARL Framework: A traditional Machine Learning model trained on the dataset without incorporating external feedback loops or data augmentation.

   b. With ARL Framework: The same model enhanced with the ARL Framework, utilizing the rejected data pipeline, external agent feedback, and augmented datasets.

Table 5.3: Results Comparison and Analysis

| Metric | Without ARL Framework | With ARL Framework |
|--------|----------------------|--------------------|
| Accuracy | 0.82 | 0.94 |
| Precision | 0.78 | 0.92 |
| Recall | 0.75 | 0.90 |
| F1 score | 0.77 | 0.91 |

As per Table 5.3, The model with the ARL Framework achieved more accuracy than without the ARL framework. This improvement highlights the feedback loop in resolving cases where the model fails and, by extension, improves its own output. The improvement in Precision shows that ARL framework reduces false positives by filtering irrelevant data. The improvement in Recall shows the ability of ARL framework in including valid business scenarios. The overall improvement in F1 score show that ARL framework is balanced enough for enhanced decision-making.

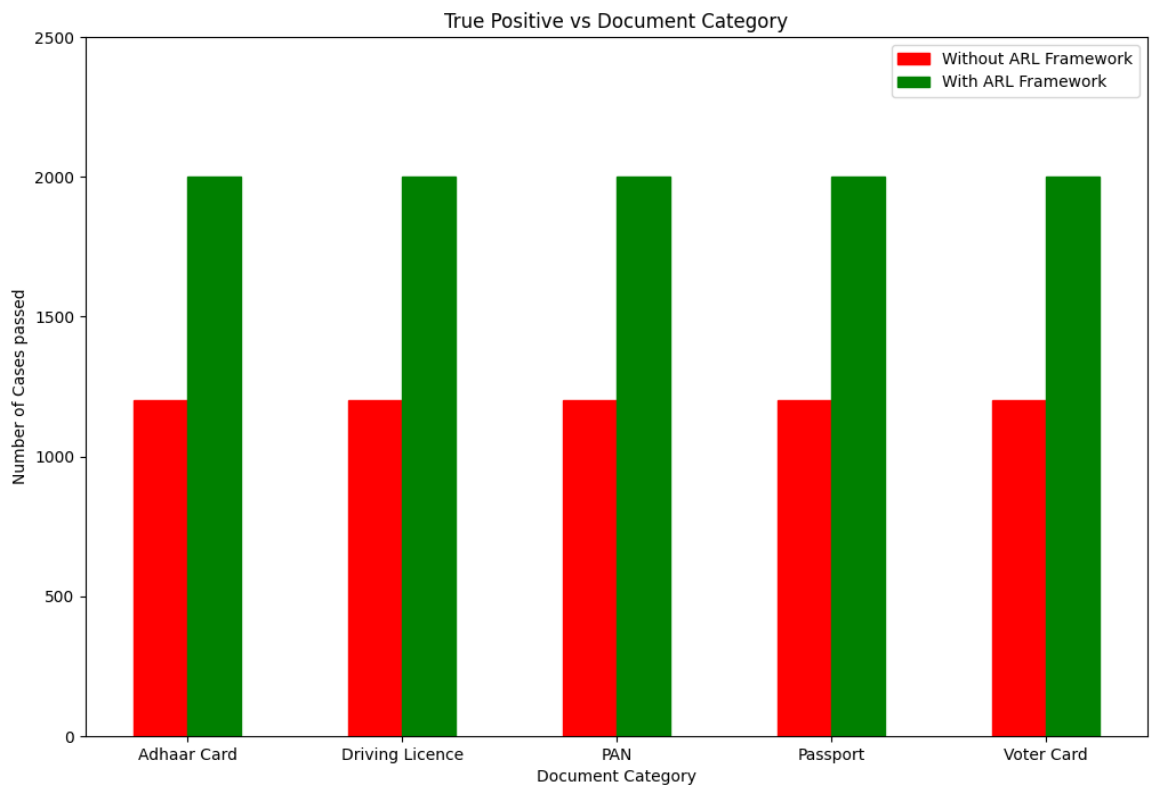

Figure 5.11: True Positive vs Document Category

The Figure 5.11 shows that the Augmented Reinforcement Learning has improved the overall accuracy of the model, thereby enhancing it decision making capabilities. From the above comparison, the advantages of the ARL Framework are clearly shown towards improving machine learning model performance. In this regard, the framework incorporates external agents and employs a rejected data pipeline to leverage the benefit of data augmentation such that it improves strengths which are lacking in the traditional Machine Learning models.

From above results, following Strengths of the ARL Framework can be deduced:

- Dynamic Adaptability: The ARL Framework ensures continuous learning with the introduction of feedback loops, through which the model adapts well to changing scenarios.

- Improve their decision-making process: The failure cases solved by the framework improve the model to decide properly and precisely.

- Generalization Capability: The application of augmented data readies the model to handle different conditions in the real world, thus making it more robust and reliable.

The comparison is self-evident when it points out that the ARL Framework has greatly improved the performance of machine learning models. In this case, it proposes to integrate external agents and leverage a rejected data pipeline to use data augmentation, overcoming the difficulties of traditional reinforcement learning.

Improvement of such metrics as precision, recall, and accuracy but the model is constantly evolving, thus, this makes the model more applicable to realistic tasks like document identification or information extraction. ARL Framework demonstrates a versatile and scalable solution that transforms decision-making capabilities of a model developed through machine learning in some domains.

**5.5     Discussion on Thesis Objective**

The objectives defined in Chapter 1 is discussed below:

1. **Objective: Definition of ARL Framework**

   In Chapter 3, the Augmented Reinforcement Learning Framework has been defined with human intervention in the decision-making process. As can be well exhaustively seen in Figure 3.2, two external agents are basic to the framework and are important for further refinement of machine learning and improvement in the performance of the model. They help in creating Rejected Data Pipeline, thus creating a feedback loop.

2. **Objective: Correcting Course And Eliminating The "Garbage-In, Garbage-Out" Problem**

   The "Garbage-In, Garbage-Out" problem arises when bad or irrelevant data generates a bad model. The ARL framework eliminates the problem through real-time course correction, where human agents identify suboptimal inputs or wrong ones during training. External agents filter through incoming data and the decisions the model makes, deciding which may be errors or biases influencing the learning path of the model.

   This directly addresses one of the greatest challenges of machine learning : to guarantee that both the data and actions being reinforced are of quality and aligned with desired output. The framework ensures that bad decisions are not reinforced, hence stopping a build-up of errors in training cycles through the intervention of an external agent.

   In practice, that would mean that the framework would itself help avoid degradation over time, even in the presence of noisy or incomplete data. The external agents are what helped enhance robustness-the system was able to inject real human intuition and knowledge in time to correct errors before they propagate and thus to make better decisions.

3. **Objective: Role of External Agents in the Augmented Reinforcement Learning Framework**

   The first is External Agent 1, which attempts to make the system return real-time responses concerning action and prediction during training. This response determines instances at which its model actions or predictions are wrong or at low optimum, hence forming a Rejected Data Pipeline. The pipeline is important as it filters out poor decisions or incorrect predictions; if not checked, this may degrade the performance of the model in subsequent trainings. This rejected data in essence represents the point at which the machine's decision would not come out in a correct way, and therefore some other agent had to step in and correct the course.

The External Agent 2 is just as important; where it assesses the feedback from the External Agent 1 in a rather close and close analysis. In that rejected feedback, this agent separates what can be referred to as the proper business scenarios or actionable data points. Thus, External Agent 2 assures the system that only the most relevant and correct data points are fed to the model's training cycle; therefore, an approved dataset forms the next training cycle. The approved dataset, free from errant inputs, not only improves the strength of the decision-making capabilities of the model but also improves learning efficiencies in the system.

Together, these external agents make up a high coherence feedback loop that continually enhance the learning machine to the point of making appropriate decisions even in the most complex and challenging scenarios of the real world

## 4. Objective: Implementing the Framework in a Real-World Problem Statement – "Document Identification and Information Extraction"

In Chapter 3, the algorithm for implementing the problem statement is provided. In Chapter 4, the actual implemented in detail along with the working code. ARL Framework has been successfully deployed in a Document Identification and Information Extraction problem, where proper classification and extraction of information from a document is required. In this scenario, the integration capability of the framework regarding external agents became critical in enhancing the precision and efficiency of the process. Document Identification is enhanced by the help of both the External Agents as they help create a feedback loop.

## 5.6    Summary

This chapter provides a detailed analysis of the results obtained from the implementation of the Augmented Reinforcement Learning framework with an Object Detection Machine Learning model in solving the problem of "Document Identification and Data Extraction." The ARL framework uses external agents, like human experts, to enhance the model's ability to decide on novel or ambiguous data situations.

The ARL considerably enhances the performance of the model compared to the standard reinforcement learning. It increases the accuracy and the precision of the model to high levels in both phases during the training and testing processes; more so, when complex or previously unseen document structures are present. Human or external agents guide it by correcting suboptimal decisions; hence, the model can better deal with uncertainties. This is not typically found with standard reinforcement models that usually lack decision-making when faced with unknown situations.

This chapter explains the greater implications of the ARL framework. While its success in the tasks of document identification and data extraction may give some assurance, the applications of ARL in broad fields are what will eventually be discussed in relation to various applications. These fields include the diagnosis of healthcare systems, autonomous driving, and robotics. In cases where the decisions are unsure, the combination of human expertise and machine learning will increase the possibility of better performance. This hybrid model of leveraging the strength of external agents results in a more flexible and robust system that helps lift the level of decision outcome for a complex real-world problem.

Hence, the results affirm the significance of the ARL framework for improving the performance of the overall machine learning model and its feasibility in diversified environments related with higher stakes.

# CHAPTER 6: CONCLUSIONS AND RECOMMENDATIONS

## 6.1    Introduction

This chapter concludes this thesis work. It also provides insights of contribution to existing body of knowledge. Further, future recommendation is also provided, citing various fields that require enhancement.

## 6.2    Discussion and Conclusion

The conclusion of this thesis is marked by the successful development and implementation of an ARL framework aimed at enhancing the performance of the decision-making process in machine learning models by integrating external agents. The real-world problem statement "Document Identification and Data Extraction" is used to demonstrate the ARL framework. This approach used an innovative approach that leverages the human agents' expertise to guide the model on its learning process to perform better under uncertain and ambiguous conditions.

Traditional reinforcement learning models have shown great performance in controlled settings but could fail or even become weak in more complex real-world scenarios. They easily get deteriorated due to novel data, ambiguously structured documents, and other changing factors of the dynamic world, which leads to inferior decisions. The ARL framework, however, endows the model to handle the complexities through the input of external agents such as a human or a particular system. The experiments produced, on average, major advances in terms of accuracy, precision, as well as adaptability. This was particularly so against unfamiliar or complex document structures.

This framework holds promise for general applicability in broader areas where decision-making under uncertainty is essentially pivotal. Such areas include autonomous systems, healthcare diagnostics, and other robotics work that combines human expertise with machine learning. The ARL framework's ability to enhance decision-making processes by the infusion of external input makes it an important contribution to reinforcement learning research.

This work provides not only a novel approach towards the improvement of machine learning models but also opens new avenues in the way human expertise could be integrated into AI systems. Although, the Augmented Reinforcement Framework is not limited to humans as external agents.

### 6.3    Contribution to Knowledge

This thesis has made several significant contributions to the body of knowledge in the fields of reinforcement learning, machine learning, and human-AI collaboration. The major contributions are as follows:

- **Development of the Augmented Reinforcement Learning Framework:**

    This paper introduces a novel ARL framework that introduces human agents as an external source of knowledge to machine learning models' decision-making processes. A remedy to the lack of reinforcement learning in many real-world settings where a problem statement could not be reduced or approximated to be an easy solve problem, this framework enhances the adaptive and effective ability of models when doing tasks such as identification of documents and data extraction.

- **Improved Ability to Take Better Decisions under Uncertainty:**

    The experiments reveal that the incorporation of the agents externally significantly brings about improvement in the capabilities of the model in handling the uncertainties and taking better decisions. This contribution throws light on the importance of human expertise as guiding forces behind the machine learning process, especially in the case of ambiguous or novel form of data structure. The paper illustrates how a combination of approaches leads to a strong AI system.

- **Novel algorithm for problem statement "Document Identification and Data Extraction" using templatization:**

    The thesis performs a novel way to extract data from identity documents once the document is identified. Combining the identification of documents with data extraction, this thesis adds valuable data to the existing literature and offers insights into best practices for future implementations.

- **Cross-Disciplinary Applications:**

    The work expands the realm of the reinforcement learning frameworks beyond only text document identification and thus potentially applicable to a variety of other domains, such as autonomous driving, healthcare, and robotics. This contribution posits that

human expertise should be investigated as a compelling means of enhancing machine learning in a number of high-stakes environments and also bridging theory and practice.

- **Future Work:**

  This thesis further sets a foundation for future research by proposing the ARL framework and demonstrating its efficiency. From its recommendation of further exploration of other potential external agents, and how to integrate such external agents into other machine learning paradigms, one can foster innovation and collaboration in AI development.

Overall, this thesis enriches the knowledge regarding how human-AI collaboration could upgrade the outcomes of machine learning to empower next-generation developments in intelligent systems and their versatile applications across various industries.

## 6.4     Future Recommendations

The ARL framework proposed by the thesis contains many lines of development, opening for further research. Among the most promising directions is the **diversification of external agents**. Indeed, although this thesis deals with human agents, future studies could generalize to other types of external agents, such as expert systems, crowd-sourced data or even autonomous agents with domain-specific expertise. This would be an important direction for further study to understand how such agents could complement or even outperform human interventions on certain tasks.

An exciting extension is the **scaling and generalization of the ARL Framework** over different tasks and domains. The framework can be tested in more dynamic and real-time environments, including financial forecasting, predictive maintenance, or natural disaster management, to determine its limitations. As the complexity and scale of the datasets on which the framework will be applied grow, its robustness and generalizability can be inspected regarding functionality in real-world applications.

Other areas of research could include studies on **enhancement mechanisms of agent collaboration within the ARL framework**. Improve interactions between external agents and reinforcement models, possibly leading to more complex methodologies of MARL that permit human and machine agents interacting to gain better solutions on complex decisions.

More research could be directed toward **explainability and interpretability** of the decision outputs provided by the ARL framework, as human and external agent interventions guide model learning. There is a great need to illuminate these decision-making processes for end users-somewhat dependent on application, such as healthcare or law-following accountability principles.

Finally, there is the need to check on **deployment challenges in real life**, possibly in terms of latency, cost, and security issues surrounding the ARL framework. Practical implementation strategies taking all these challenges while ensuring performance become the hard nut to crack in moving this framework from research to practical applications. Among the potential avenues, this would ensure the ARL framework would continue to evolve and contribute to future AI systems.

# APPENDIX A: RESEARCH PLAN

## Project Gantt Chart

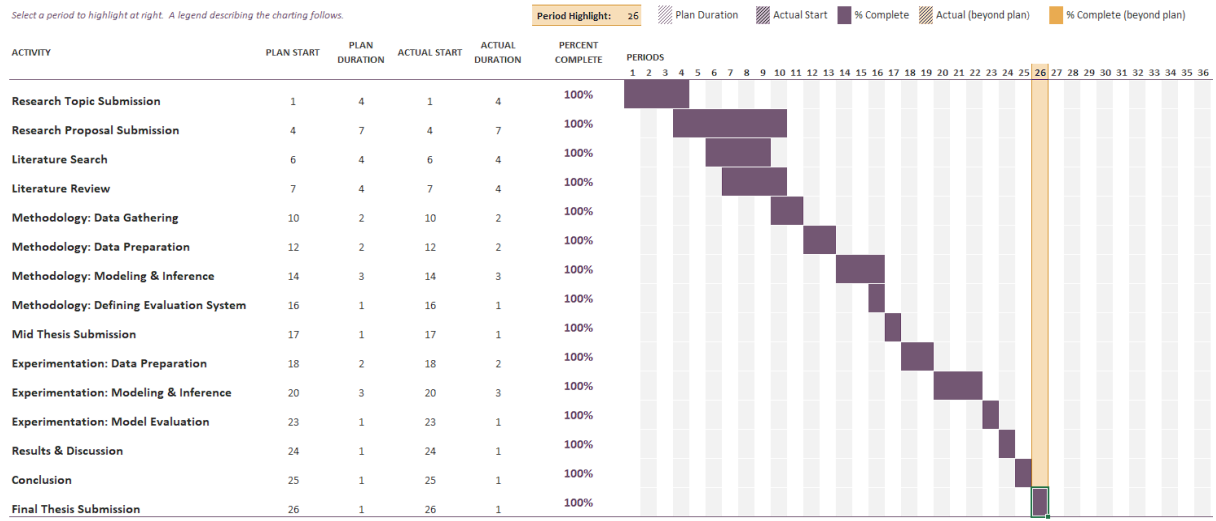

Figure A.1: Research Project Gantt Chart

*Note: 1 Period = 1 Calendar Week*

## Risk Mitigation and Contingency Plan

The completion risks of the thesis work and its associated contingencies are listed in Table A.1 below:

Table A.1: Risk and Contingency

| Risk | Contingency |
|------|-------------|
| Timelines may be affected due to health or personal issues. | Buffer time allocation to each subtask in project management; inform LJMU/UpGrad administration, and ask for an extension. |
| GPUs are unavailable for data generation and Machine Learning model training. | Use cloud GPUs. A nano version of the YOLO Pre-Trained Model may be used for training. |

# APPENDIX B: RESEARCH PROPOSAL

AUGMENTED REINFORCEMENT LEARNING FRAMEWORK FOR ENHANCING DECISION-MAKING IN
MACHINE LEARNING MODELS USING EXTERNAL AGENTS

SANDESH KUMAR SINGH

Research Proposal
MASTERS OF COMPUTER SCIENCE

AUGUST 2024

# Abstract

Training a Machine Learning model is similar to teaching students to make decisions. What if the model trains itself, just like a student learns more promptly after hand-holding in the initial learning phase? That is interesting, right? Reinforced Learning is one of the principal concepts of Artificial Intelligence and Machine Learning. It rectifies the errors induced in the previous iterations or epochs of the training cycle of a model. Although it can be utilized for different purposes, it cannot help in cases of a lack of a training dataset and is unable to explain its decisions.

The decision-making of the model will be enhanced by developing a new framework called Augmented Reinforcement Learning in this thesis. Much like the student sometimes needs a nudge in the right direction for his success, be it carrots or sticks. In the same way, an external agent will be involved in model training to use their expertise and enhance the model's decisions. Two different perspectives will be explored. First, from the model's standpoint. And second, from the agent's standpoint. In this thesis, the "Document Identification and Information Extraction" problem statement will be used to illustrate the framework. It will, in a later section, be generalized for a wider class of problems.

Implementation of the framework may differ from one problem statement to another. But the crux of it remains constant. Not only does it enhance the model's decision-making, but also greatly improve the explainability of the model. So, let's dive into the novel Augmented Reinforcement Learning framework.

# Table of Contents

# List of Tables

# List of Figures

# List of Abbreviations

| | |
|---|---|
| AI | Artificial Intelligence |
| ARL | Augmented Reinforcement Learning |
| CNN | Convolutional Neural Network |
| EDA | Exploratory Data Analysis |
| GPU | Graphics Processing Unit |
| MARL | Multi-Agent Reinforcement Learning |
| ML | Machine Learning |
| OCR | Optical Character Recognition |
| RAM | Random Access Memory |
| RL | Reinforcement Learning |
| RNN | Recurrent Neural Network |
| YOLO | You Only Look Once |

## 1. Background

Advancements in Artificial Intelligence and Machine Learning are driving the world toward "Singularity". The pace might be slow, but it will get there. To make sure it is on the right track, the decision-making process of the Machine Learning model should be made as similar to a human (an external agent) as possible. That's where Reinforcement Learning comes into the picture. Reinforcement Learning is a subfield of Machine learning and literally has changed the way of approaching sequential decision-making problems (Wang, X., 2024).

Reinforcement Learning is that variety of machine learning wherein models learn to make decisions by interacting with an environment. In a way, it's based on the paradigm of carrots and sticks. It, in principle, eradicates the wrong decisions of a Machine Learning model, correcting it in the next epoch. That is what constitutes a training cycle. It's important in making intelligent systems that are able to adapt and learn from interaction with their environment. Compared to other techniques, the model of Reinforcement Learning learns directly from experiences; hence, it becomes prepared for facing difficult situations.

The methods used for learning by an agent in reinforcement learning are generally of three types: value-based, policy-based, and model-based. Indeed, there have been enormous efforts directed toward this area, but often, value-based and policy-based evaluations are done using external agents. Now, the quite natural idea that would come to one's mind is that humans as external agents can be used to train Machine Learning models to enhance their decision-making abilities.

Recent research has shown that multiple agents and integrating reinforcement learning with other optimization techniques may improve learning efficiency and lead to better decisions. One of the works studied the role of multi-agent optimization in reinforcement learning and concluded that it "significantly enhances the learning efficiency and decision-making capabilities of the agents, especially in complex environments." (Wang, X., 2024).

Another such novel approach is the combination of Reinforcement Learning with genetic algorithms (Rohoullah et al., 2023). In the domain of medical sciences, Reinforcement Learning was also investigated for its potential to enhance information extraction from pathology reports (Park, B., 2022).

Another critical aspect in reinforcement learning is efficiency, more so in visually rich environments, where sample efficiency may vastly accelerate real-world deployment.

According to one study, sample efficiency is important in raising the speed of deployment of model-free reinforcement learning systems in visually rich environments (Yarats et al., 2021). Others insist on the need for efficiency in Reinforcement Learning. In this respect, efficiency in reinforcement learning is very important since it enables the minimization of the quantity of resources required for training while optimizing the performance of the learning agents (Blakeman, S., 2021).

These contributions build into a more powerful understanding of Reinforcement Learning and open up new directions for fitting human agents into the frameworks of Reinforcement Learning as independent decision-makers to further beef up decision-making processes. That means one can use multi-agent systems, genetic algorithms, and enhanced optimization techniques to come up with more efficient, effective, and adaptable models of Reinforcement Learning in complex scenarios.

This thesis will focus on how humans, as external agents, can become central in a machine learning model. Human intervention in the Augmented Reinforcement Learning framework would take place in all three stages: feedback, approval, and validation. Introducing humans to the Augmented Reinforcement Learning framework could enable optimizing and improving the accuracy of the Machine Learning model.

## 2. Related Work

Enhancement of decision-making in complex and dynamic environments is one of the critical challenges during the process of machine learning. Traditional machine learning models lack the capability of adaptation in complex scenarios because they were designed over static datasets and pre-defined rules. Now, reinforcement learning can become very promising in the resolution of this problem by letting the models learn from interactions with their environment and optimization of decision-making processes based on trial and error.

Decision making in Machine Learning models is important since it propels the efficiency and efficacy of automated systems cutting across various sectors. Poor decision-making can lead to sub-optimal outcomes alongside increased expenses and occasionally even poses safety risks. For example, bad decisions in health may imply the recovery of patients, while in finance, they could bring immense losses. Therefore, the improvement of decision-making in Machine Learning models is very critical to ensure the reliability and success of automated systems in real-world applications.

Inefficient decision-making in Machine Learning models has repercussions on multiple sectors. For example, in autonomous vehicles, bad decisions risk the safety of passengers and the efficiency of vehicle operation. In cases of industrial automation, it may cause production stoppages, leading to increased operational costs. Furthermore, regarding customer service, it reduces user satisfaction and loyalty. The societal impact is reduced faith in automated systems and, as a result, slow adoption of advanced technologies that may come with associated economic and safety risks.

There are many notable research studies done in this area. The need for the Augmented Reinforcement Learning framework will be better understood after discussing existing work regarding Reinforcement Learning. Interesting research that is very related to this topic explores multi-agent optimization, opportunistic exploration, and causal interpretation in Reinforcement Learning (Wang, X., 2024). The research has pointed out the role of multi-agent systems in the optimization of tasks belonging to the domain of Reinforcement Learning. According to them, because of opportunistic exploration, agents at runtime change the levels of exploration by taking cues from the environment. This foundational work explains how multiple agents may collaboratively improve the learning and decision-making processes in a framework of Reinforcement Learning.

Another innovative approach integrates genetic algorithms into reinforcement learning for automated pre-processing of images aimed at optimized text extraction (Rohoullah et al., 2023). Their work demonstrated that reinforcement learning holds a great potential to automate complicated tasks like image processing, which is very vital in a good number of fields of applications, including data extraction and analysis. Their approach further incorporates genetic algorithms into the reinforcement learning model to enhance efficiency and effectiveness.

Machine learning has also been applied to information extraction from pathology reports, while on the other hand, another study has proposed an adaptive offline value estimation method in Reinforcement Learning (Park, B., 2022). This paper strongly advocates for very useful applications of Reinforcement Learning in healthcare and shows how adaptive methods contribute to the value estimation process to bring about more accuracy and reliability in decision making within the offline scenarios of Reinforcement Learning.

The majority of attention in multi-agent Reinforcement Learning has been placed within the umbrella of how multiple agents could be trained at the same time for the optimization of collective performance (Morcos, A., 2022). This paper is mainly useful in scenarios in which collaboration and coordination among the agents are important to reach optimal results. This most especially marks the capability of multi-agent systems in scaling and providing more robustness to Reinforcement Learning models.

An interesting study addresses how to improve the sample efficiency in model-free Reinforcement Learning from images (Yarats et al., 2021). It shows that high improvements in the efficiency of Reinforcement Learning models can be gained to reduce the number of samples needed to train a model by incorporating data augmentation and representation learning techniques. This work is of significant importance for applications where it is costly or very hard to obtain large datasets.

Some of the works are concentrated on efficient Reinforcement Learning, both in humans and machines, gaining insights into how human cognitive processes can inform the design of Reinforcement Learning algorithms (Blakeman, S., 2021). Basically, this research closes a gap between human learning and machine learning, arguing that human-inspired strategies can be used in building more efficient and effective models of Reinforcement Learning.

The closest related research paper discusses the use of humans in the process of Reinforcement Learning. It focuses on how human interventions guide the learning process, improve sample efficiency, and enhance decision-making in complex environments (Huang et al., 2022). It proposes a few frameworks and algorithms pertaining to the effective incorporation of human insights into this process, which result in huge performance improvements compared with traditional methods for Reinforcement Learning.

Noticeably, Cooperative Multi-Agent Reinforcement Learning has also been approached. The paper contributes in-depth research on agent-based reinforcement learning for multi-agent systems; it further expresses various algorithms and frameworks that can enable cooperation and competition between the agents to address the challenges of scalability, communication, and coordination (Yang et al., 2021). What is just outstanding is the destination use for which the study is going to be used. These results have implications on an extremely broad range of applications, from autonomous vehicles to smart grids, that use agent-based concepts of reinforcement learning.

Reinforcement Learning, however, extends even to the finance market. Another paper investigates the application of agent-based reinforcement learning in financial markets. This paper proposes Reinforcement Learning agents that learn to adapt to the dynamics of the markets in order to make appropriate trading decisions (Wang et al., 2024). The results show that they are able to realize superior performance compared with traditional trading strategies, hence revealing the huge potential of Reinforcement Learning in financial applications.

Another interesting study focuses on how much trustworthiness and reliability are expected to come with Multi-Agent Reinforcement Learning (MARL) methods. It classifies the existing approaches to Multi-Agent Reinforcement Learning in a simple way and points to the recent advancements essentially in lines related to the robustness and safety of Multi-Agent Reinforcement Learning systems (Wang et al., 2022). Another important aspect discussed is the application areas wherein Multi-Agent Reinforcement Learning shows promise, such as autonomous driving and smart cities.

Essentially, the Multi-Agent Reinforcement Learning method shows decentralization aspects that are observed to be showing scalability and robustness (Sun et al., 2021). Agents in this algorithm operate without a central coordinator. The paper introduces algorithms to enable learning among agents to make independent decisions while executing coordinated behavior.

Some of the research has focused on the hierarchical structure of Reinforcement Learning. The field of reinforcement learning has been decomposing complex tasks into simpler subtasks such that an agent can learn and coordinate (Liu et al., 2022). In this paper, some of the hierarchical architectures are discussed that prove their efficiency by enhancing efficiency in learning and performance for multi-agent environments.

The research gap lies in the need for more effective methods to incorporate Reinforcement Learning into Machine Learning models for better decision-making. Most of the current techniques for Reinforcement Learning are associated with high computational requirements, long training time, and large amounts of interaction data. New frameworks of Reinforcement Learning should be invented which enhance the efficiency and effectiveness of decision-making without costing too much or requiring an impractical amount of data. Overall, the existing studies contribute to the progress of the field of Reinforcement Learning by investigating different methodologies and applications. It would further strengthen these decision-making processes in human intuition and expertise, besides bringing on board the indefatigable strengths of machine learning models if reinforcement learning is integrated with human agents, much like any other external agents for that matter. Though Augmented Reinforcement Learning allows for any kind of external agents in its framework, humans do play a key role.

It is a justified focus of research to integrate Reinforcement Learning into Machine Learning models for decision making, having a potential gain in capability for the automated system. Improved decision processes underpin the potential for more accurate, more reliable, and efficient outcomes for many different domains. This research is of high relevance if Machine Learning models are to be enabled to work in real dynamic environments in which decisions need to be made continuously and, often adaptively.

The justification comes many folds because the Augmented Reinforcement Learning framework solves challenges on multiple fronts. High computational demands can be brought down and reduce the lengthy training cycle of the Machine Learning model. Since these models have large amounts of data to learn from, this will make the interaction time with the data reduces by the insights and expertise provided by the external agents. This will allow the Machine Learning models to respond and fit into unpredictable situations. External agents could provide contextual understanding and intuitive decisions for which the model might respond poorly.

Truly transformative advancements will revolutionaries a number of sectors if solutions to the problem of how to enhance decision-making through Reinforcement Learning can be found. For health, this would mean better diagnostic accuracy and more effective, differentiated treatment plans for better patient outcomes. For autonomous driving, it would translate to safer and more efficient navigation systems. In finance, one gets an upper hand over perfecting trading strategies and risk management processes. All in all, it means greater operational efficiency, reduced costs, higher safety levels, and better user satisfaction.

By addressing this challenge, the community of Machine Learning will be pushing the envelope toward intelligent automation and hence creating smarter, more adaptive systems that could make optimal decisions within dynamic, complex environments. This progress will drive technological innovation and societal advancement, paving the way for the next generation of intelligent systems. If the dataset is concentrated, closely matching near the training dataset, a Machine Learning model works very well on the validation dataset. It has one critical challenge connected with the improvement of its decision-making in a complex environment. Because of the fixed dataset and predefined rules, it's hard to adapt efficiently to real-world problems and provide acceptable decisions for that particular scenario. The framework of Augmented Reinforcement Learning proposed seems to be a very promising approach. It can help to overcome this problem by forcing the Machine Learning model to learn from the external agents.

Effectiveness and efficiency of automated systems across sectors are based upon how much improvement can be made to the decision-making processes in the Machine Learning model. A bad decision-making process will have undesired results, increased cost, loss of time, and resources. Let us take the banking sector, for example; poor data entry might lead to credit decisions gone bad hence leading to financial losses by falling like a set of arranged dominoes. Therefore, the decision-making capabilities of the Machine Learning model need to be increased to ensure its reliability and success in real-world applications.

The proposed Augmented Reinforcement Learning framework is generalized for any problem statement. Be it autonomous vehicles, industrial automation, the banking sector, or medicine. The Augmented Reinforcement Learning framework can be applied anywhere. From a different perspective, there is a social impact: an increase in trust in automated systems, and more adoption of advanced technologies.

**3. Research Questions**

From the previous section, it's evident that many notable research works done in the field of Reinforcement Learning have used multi-agent model optimization (Wang, X., 2024), Genetic Algorithms (Rohoullah, R. and Joakim, M., 2023), Adaptive Offline value estimation (Park, B., 2022), and Deep Learning models (Renda, H.E.E., 2023). As mentioned earlier, interesting areas related to RL in humans and machines have also been studied (Blakeman, S., 2021).

Now, focusing specifically on utilizing external agents in RL, this thesis tries to answer the following research questions:

5. What is the Augmented Reinforcement Learning Framework? What are its different stages?

6. Machine Learning model faces a major problem of "Garbage-In, Garbage-Out". How does the Augmented Reinforcement Learning framework rectify this problem?

7. What role do the external agents (humans) play in the proposed framework?

8. How to implement the Augmented Reinforcement Learning framework in real-world problems?

## 4. Aim and Objectives

The main aim of this research is to propose an Augmented Reinforcement Learning framework that enhances the decision-making capabilities of a machine-learning model. Reinforcement Learning plays an important role in training a Machine Learning model. However, it is crucial to utilize humans as external agents and evaluators to reap better results from Reinforcement Learning. First, the definition of the framework will be provided. Then, to prove the hypothesis, this framework will be implemented for the problem statement "Document Identification and Information Extraction". This is a real-world problem, used in the banking sector.

The research objectives are formulated based on the aim of this study which are as follows:

5. To define the different stages of the Augmented Reinforcement Learning framework.
6. To understand how the proposed framework rectifies course correction of a model in the training cycle ("Garbage-In, Garbage-Out" problem).
7. To identify the role of external agents in the Augmented Reinforcement Learning framework.
8. To implement the framework in a real-life problem statement "Document Identification and Information Extraction".

## 5. Significance of the Study

Reinforcement Learning is a very active area of research. It has high significance in the field of Artificial Intelligence and Machine Learning. The Augmented Reinforcement Learning framework will aid in solving a number of key challenges. It will open up new opportunities for research and practical applications wherein humans are placed as external agents whose expertise, intuition, and knowledge can be utilized to make decisions and improve the model's accuracy and reliability. By incorporating humans as external agents, it capitalizes on their experience and intuition to build more resilient decision-making processes, hence improving the accuracy of the Reinforcement Learning models (Huang et al., 2022).

Sometimes, these limitations of the training dataset transfer to the Machine Learning model, affecting decisions with biases and errors. These biases may originate decisions that are hard to accept by society, reducing the confidence of Artificial Intelligence systems (Garcia and Fernández, 2023). This makes building trust in Artificial Intelligence hard on the part of society. This will be helpful if external agents are in a position to monitor how a model is being trained or how a valid data set is being fed to the model. This will help in fairer outcome.

This current research will attempt to fill up the gaps between Artificial Intelligence and humans by creating collaboration and a conducive environment. This makes Artificial Intelligence technologies more understandable and acceptable to users if the transparency and interpretability of Artificial Intelligence systems are improved (Silver et al., 2024). The framework brings together the power of Artificial Intelligence and human experience in the enhancement of problem-solving capabilities, ensuring Artificial Intelligence systems are aligned to human values and ethical standards (Lin et al., 2024). This therefore contributes to the development of Ethical Artificial Intelligence.

In that regard, the research work proposed can have significant advantages to society and individual citizens. This will make Artificial Intelligence systems much more reliable, fair, and aligned with human values so as to augment public trust in Artificial Intelligence technologies. This in turn facilitates ease in integrating Artificial Intelligence into everyday life for improving the quality of life with more accurate and ethical decision-making systems.

## 6. Scope of the Study

The scope of this research work is restricted to the following points:

8. This research will be completed in 16 weeks after submitting the research proposal document.

9. This research will include a definition and explanation of the Augmented Reinforcement Learning framework; and implementation of the framework in one real-life problem statement, the "Document Identification and Information Extraction".

10. The model training and evaluation will be conducted using a personal desktop with 8 GB RAM, 4 GB shared GPU, and AMD Ryzen 3 5000 series processor.

11. Keeping time constraints, document privacy issues, and government regulations in mind, the data collection procedure will not be done, instead the document synthesizer tool will be used.

12. All the documents/images in consideration will belong to the Indian region. Although, the model can be trained on any document. However, the documents of the Indian region are easier to generate via the synthesizer tool.

13. Evaluation of the model trained for the problem statement will be part of the research. This is necessary to compare the efficiency of the new model.

14. To evaluate the framework, merely a single problem would not suffice. A comparison of models from multiple problem statements is required. This will take more than 16 weeks. Keeping time constraints in mind, evaluation of the framework will not be part of this research and can be considered for future work.

## 7. Research Methodology

This thesis proposes a novel Augmented Reinforcement Learning framework. Our main aim is to validate how the proposed framework helps to enhance the model's efficiency. So, in this section, the following questions will be answered:

1. What dataset will be used for training, validating, and testing the model?
2. What is the Augmented Reinforcement Learning framework?
3. How to implement the Augmented Reinforcement Learning framework? (In this case for the "Document Identification and Information Extraction" problem statement)
4. How to prepare the dataset?
5. How to evaluate the Machine Learning model?

### 7.1 Dataset Description

As stated earlier, the "Document Identification and Information Extraction" problem statement will be used. This project depends on a dataset containing personal documents (which can include various document categories like Identity documents, Banking documents, Vehicle Documents, etc.), different organizations we contacted about this matter, but they refused to share the dataset because these types of documents have privacy issues, along with strict government regulations and policies. Moreover, people are reluctant to share these documents with others, even if it is for research. So direct and indirect data collection is not feasible. A synthetic dataset will be used for this purpose which is available at https://github.com/meetsandesh/synthetic_document_generator to prove my hypothesis. A sample of 5000 images for 5 different document types is available at https://github.com/meetsandesh/identification_document_dataset.

## 7.2 Augmented Reinforcement Learning Framework Description

### 7.2.1 Existing Flow of Machine Learning Model

In a typical Machine Learning model training, the entire process is divided in 12 stages. These stages are a superset of the entire process. A few of them can be skipped depending on various situations. This structured approach is supposed to ensure systematic development, leading to robust and reliable machine learning models. The flow chart of the whole process is shown in Figure 7.2.1. Let's have a look at each one of them.

#### 7.2.1.1 Problem Formulation

It starts by delineating the research problem through a deep study of existing methodologies, bringing about existing gaps and clearly identifying the objectives of the study, ensuring it is focused towards arriving at an output in conformity with the latest developments in reinforcement learning (Silver et al., 2024).

#### 7.2.1.2 Data Acquisition

It involves collecting various data sources, including public datasets, APIs, and real-time data streams. This data is highly instrumental in the training of the model; therefore, it needs to be complete and representative of a collection of diverse scenarios so that once developed, the model will be robust and reliable (Lin et al., 2024).

#### 7.2.1.3 Data Cleaning & Labeling

In this step, noise, inconsistencies, and irrelevant data are removed from the collected dataset (Yang et al., 2021). Accurate labeling is important for that and is mostly done by both automated tools and humans in a combination to ensure quality input into the model.

#### 7.2.1.4 Data Analysis

Exploratory Data Analysis (EDA) is conducted to understand the underlying patterns, distributions, and relationships of data (Zhao et al., 2024). This analysis guides feature engineering, points out possible biases, and ensures that the data is well-prepared for further stages.

#### 7.2.1.5 Data Preprocessing

It is where the data gets pre-processed with respect to normalization, scaling, and augmentation in order to get it ready for training a model. Techniques in image filtering and data augmentation are applied to improve generalization across different types and conditions of data.

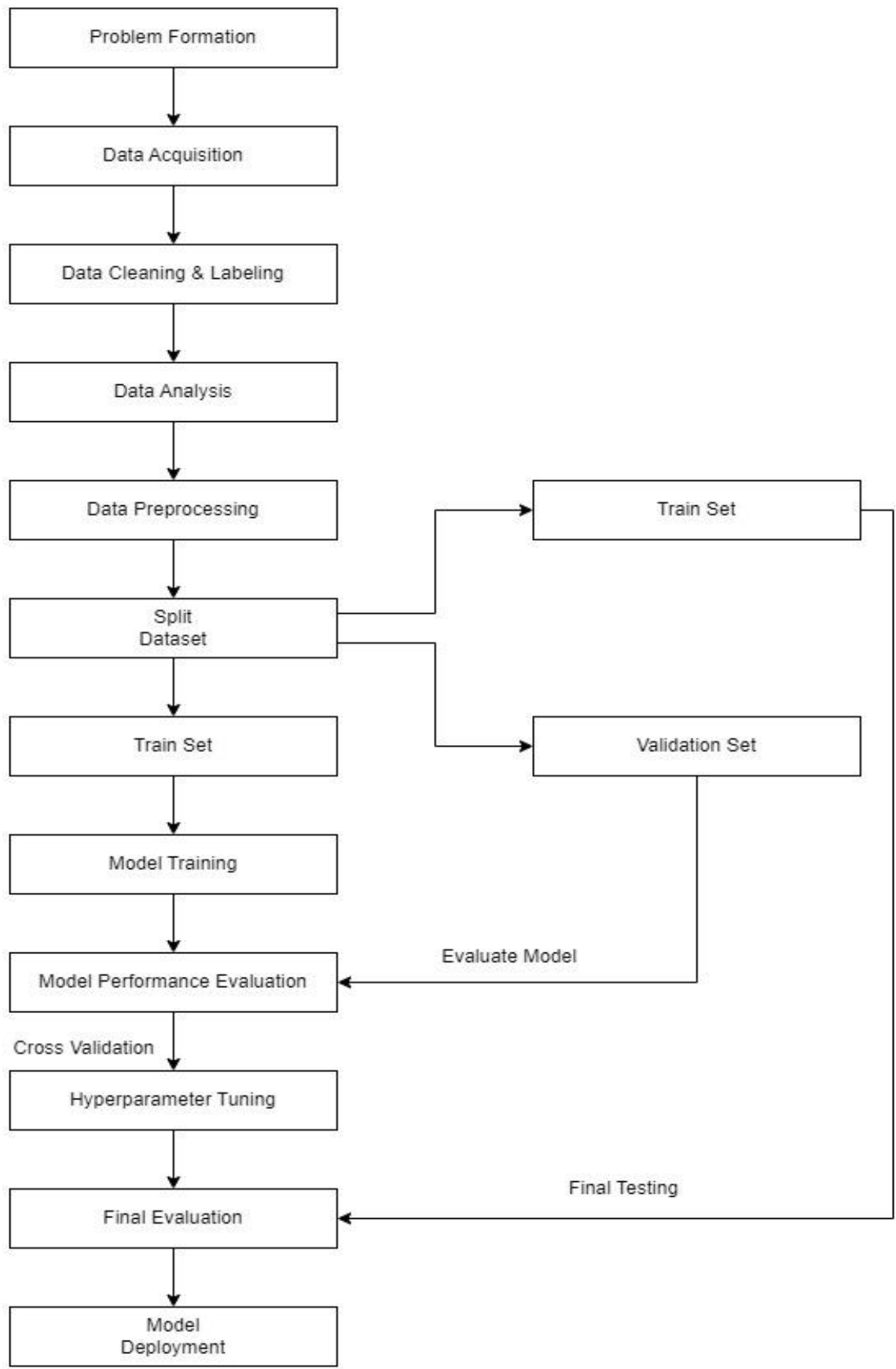

Figure 7.2.1 Typical Machine Learning Model Flow Chart

### 7.2.1.6    Split Dataset

It is divided into training, validation, and testing subsets by methods like stratified sampling. This will ensure that the model is tested on diverse data and hence generalize better to new unseen data during deployment.

### 7.2.1.7    Model Training

In this stage, algorithms like neural networks are used to train the model on the training data (Zhang et al., 2021). Gradient descent and other techniques would be used in optimizing the model's parameters with a view to reducing the error rate and hence improving accuracy.

### 7.2.1.8    Model Performance Evaluation

Evaluation metrics, such as precision, recall, and F1-scores, are indicative of a good knowledge of how the fitted model is performing. These metrics identify areas in which further improvement may be required by the model.

### 7.2.1.9    Cross Validation

K-fold cross-validation or other cross-validation techniques can be applied to confirm the model performance over various subsets of data (Rohoullah et al., 2023). This step is very important to detect overfitting and avoid a model that doesn't generalize.

### 7.2.1.10    Hyperparameter Tuning

Tuning involves the adjusting of a model's hyperparameters, for instance, learning rate and batch size, to find the right setting. It is techniques like grid search that will allow one to change through many different combinations in an orderly way for better model performance.

### 7.2.1.11    Final Evaluation

This version undergoes the final test by the test set, which assures predefined standards of performance from the model. In other words, this stage ensures that the model works in the real world, by proving it reliable and robust (Chen et al., 2023).

### 7.2.1.12    Model Deployment

The last step would be to then deploy the model into a production environment and monitor its performance. Tools such as Docker, Kubernetes, or cloud services help keep models scalable and adaptable (Kumar et al., 2022).

### 7.2.2  Proposed Modification in the Flow of Machine Learning Model

The proposed Augmented Reinforcement Learning framework suggests a significant change in the existing flow char as evident from the above figure. It incorporates external agents into the Machine Learning model development process, acting as a validator to enhance decision-making. The initial stages remain the same: Problem Formulation, Data Acquisition, Data Cleaning & Labeling, Data Analysis, Data Preprocessing, and splitting the dataset into Training Set and Validation Set.

After Model Training and Model Performance Evaluation, the process continues with Cross-Validation, Hyperparameter Tuning, Final Testing, and Final Evaluation. A critical new decision point is introduced before Model Deployment: Is the decision made by the model acceptable?

Here, External Agent 1 reviews the model's decisions. If the decisions are acceptable, the model is deployed. If not, the data or scenario is rejected and further analyzed by External Agent 2, who determines whether the rejected data constitutes a valid scenario. If External Agent 2 deems it valid, the rejected data is sent back for Rejected Data Augmentation, effectively creating a feedback loop to enhance the dataset and improve the model. If the data is invalid, it is sent to the Rejected Data Pipeline.

This integration of external agents ensures human insights, intuitions, and expertise are incorporated, improving the model's reliability and robustness by continuously refining the dataset and model performance based on real-world feedback.

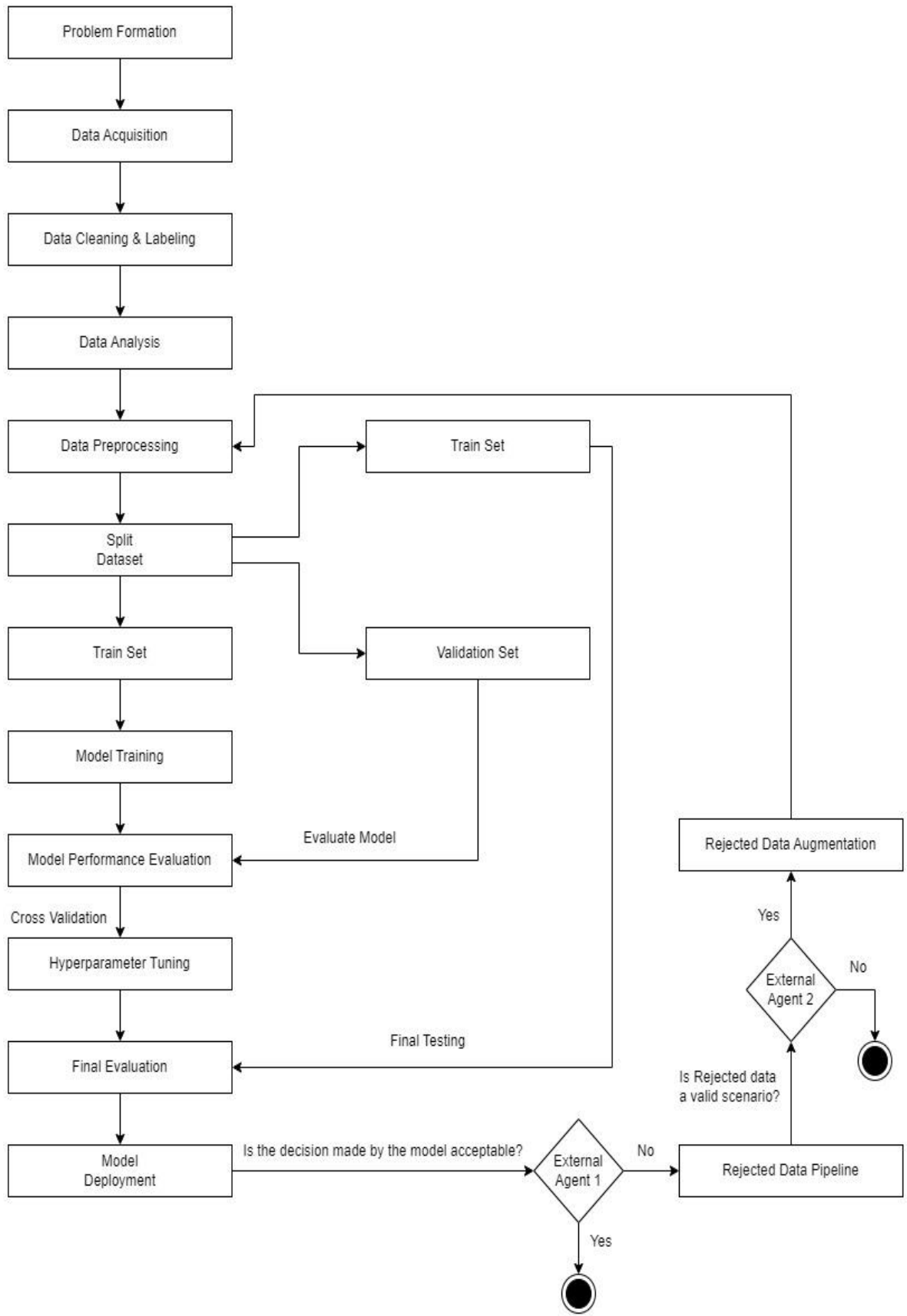

Figure 7.2.2 The Augmented Reinforcement Learning Framework Flow Chart

**7.3    Augmented Reinforcement Learning Framework Implementation**

The Augmented Reinforcement Learning framework will be implemented for the problem statement "Document Identification and Information Extraction". Considering the example of the banking sector. One of the jobs is data collection from potential customers and applying for a loan to facilitate lending. Currently, this is done using, lots and lots of data entry agents, whose sole job is to gather all the images/documents, and enter customers' data from the above-said documents in their customer acquisition systems. The breakdown of the whole problem statement is as follows:

1. Customers ask for banking services like loans, account opening, etc.
2. Banks ask for identification documents to verify the identity of the customers.
3. Banks, after receiving the documents, go through a painstaking process of manual data entry. The step is highly error-prone and costly for banks.
4. The data entry process requires some additional subtasks:
   a. Identification of the document
   b. Indexing of the various pages of documents
   c. Reading and verification of the information from the document
   d. Entering the above information in their respective systems
5. Finally providing the required service to the customers.

The problem statement is poised to solve the fourth step of the whole process. The following procedure will be done to achieve the automation of the fourth step:

1. A CNN-based Machine Learning model will be created to identify and index the respective document.
2. A novel templatization technique will be used to identify the information extraction area in the document.
3. An RNN-based model will be deployed to perform OCR for data extraction.
4. After model training, based on the model performance, the Augmented Reinforcement Learning framework will be employed to rectify the problems in the model. It will be taken care of in the next iteration of the training cycle.

**7.4     Data Preparation**

Since a specific class of images (or documents, if you may) is being targeted, there are two parts of the dataset on which need focus:

1. Documents Generation: There is no specific predefined template in which they are demanded/asked in the industry. However, image clarity is of utmost importance. Therefore, an image template is utilized for every document type to generate datasets. The minimum density of pixels used for experimentation is 96 dpi. However, the size of the image varies depending on the test cases. For this purpose, exact template-sized images and their A4 sheet counterparts will be used. After using the default template as the base case, the Augmentation technique will be employed to widen the test coverage. Some more filters like Greyscale and Random Localisation filters will also be used to generate more test cases.

   The following images will give an idea of how the dataset will looks like.

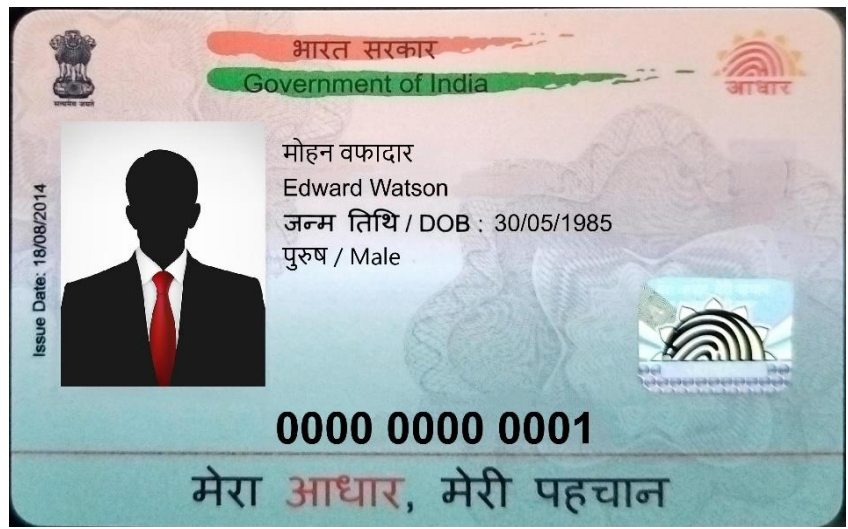

Figure 7.4.1 Sample Adhaar Card

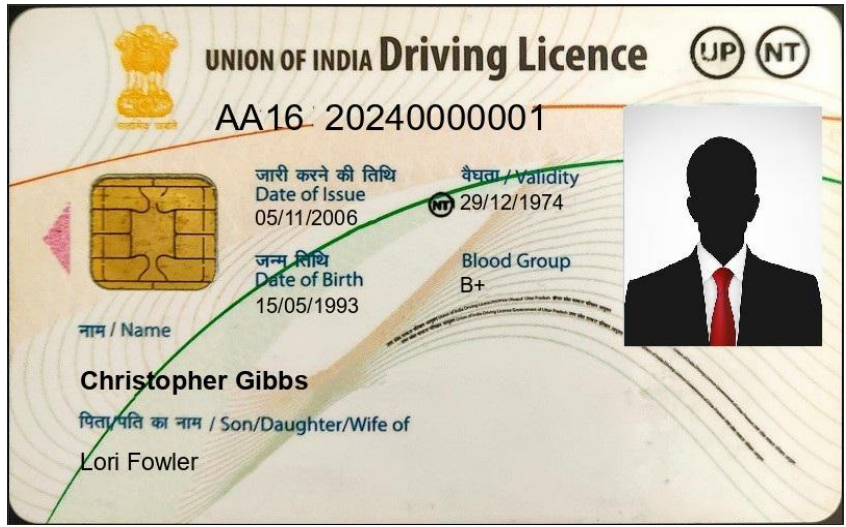

Figure 7.4.2 Sample Driving Licence

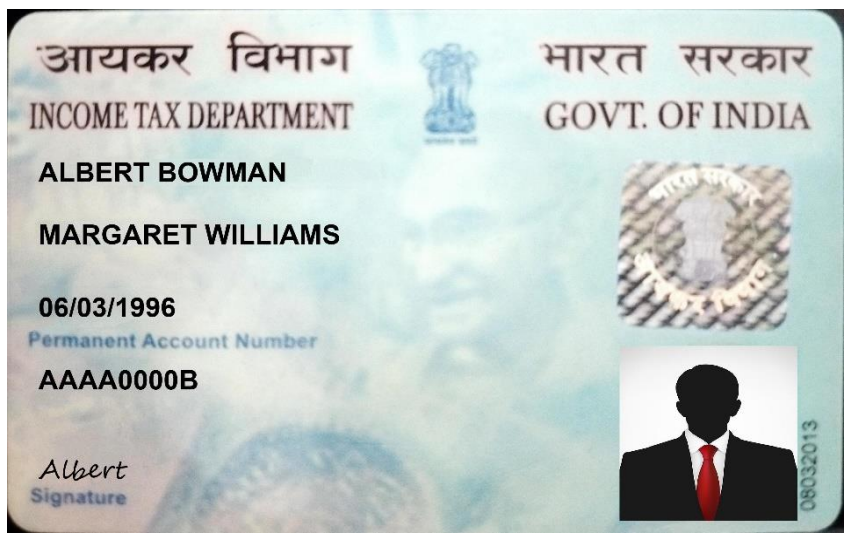

Figure 7.4.3 Sample PAN Card

Figure 7.4.4 Sample Passport

Figure 7.4.5 Sample Voter Card

2. Document Annotation: Using the Python library 'faker', identification-related data will be generated and embedded into the images and stored in a separate text file. This text file will later be used to verify the information extracted by the model.

## 7.5    Evaluation

The annotation made in the previous section will be used to test model accuracy. These annotations, created using the Python library "faker", contain identification-related data embedded into the images and store them in separate text files. These text files are ground truths to measure the performance of the model.

The evaluation begins by feeding the augmented dataset into the model. It comprises generated images of varied sizes and densities, filters like Greyscale, and Random Localization. The model processes the images to extract embedded information. Extracted information is then compared against ground-truth annotations.

Accuracy in this respect will be measured by the accuracy with which it extracts information from the images. Among others, key metrics to use include Precision, Recall, and F1-score, which collectively provide a more complete view of model performance.

The precision measures the correctness of the extracted data. It is the ratio of true positive (TP) results to the total number of positive results (TP + FP), where FP represents false positives (Jain, 2023). The formula is:

$$Precision = \frac{TP}{TP + FP}$$

The Recall measures the model's ability to capture all relevant information. It is defined as the ratio of true positive (TP) results to the total number of actual positives (TP + FN), where FN represents false negatives (Jain, 2023). The formula is:

$$Recall\ = \frac{TP}{TP + FN}$$

The combination of the precision and the recall formulates the F1 score. The **F1 score** provides a balanced measure of precision and recall. It is the harmonic mean of precision and recall, ensuring a comprehensive performance indicator (Jain, 2023). The formula is:

$$F1\ score\ = \frac{2 \cdot Precision \cdot Recall}{Precision + Recall}$$

Additionally, the evaluation considers the model's performance across different image qualities and augmentations, ensuring it is robust and reliable in various real-world scenarios.

## 8. Required Resources

### 8.1      Hardware Requirements

The following are the hardware requirements to complete the thesis work:

1. A computer having internet access, capable of:
    a. Web Browsing
    b. Document Writing
    c. Code Compilation and Execution
2. For model training without a dedicated GPU, it's advisable to have a computer with AMD RYZEN 3 5000 series, a minimum of 16 GB of RAM, and 4 GB of GPU.
3. For faster training, it's advisable to have access to a dedicated 16 GB of GPU.

### 8.2      Software Requirements

The following are the software requirements to complete the thesis work:

1. Microsoft Word, or any other report-writing software.
2. Google Chrome, or any other web browser.
3. IDE, for coding.
4. Anaconda Navigator, version – 2.5.2
5. Python, version – 3.8.19
6. YOLO, version – v8
7. ultralytics, version – 8.0.58
8. numpy, version – 1.24.2
9. torch, version – 2.3.0
10. torchvision, version – 2.3.0
11. torchaudio, version – 0.18.0
12. easyocr, version – 1.7.1
13. opencv-python, version – 4.9.0.80
14. matplotlib, version – 3.7.5
15. jproperties, version – 2.1.1

## 9. Research Plan

### 9.1　　Project Gantt Chart

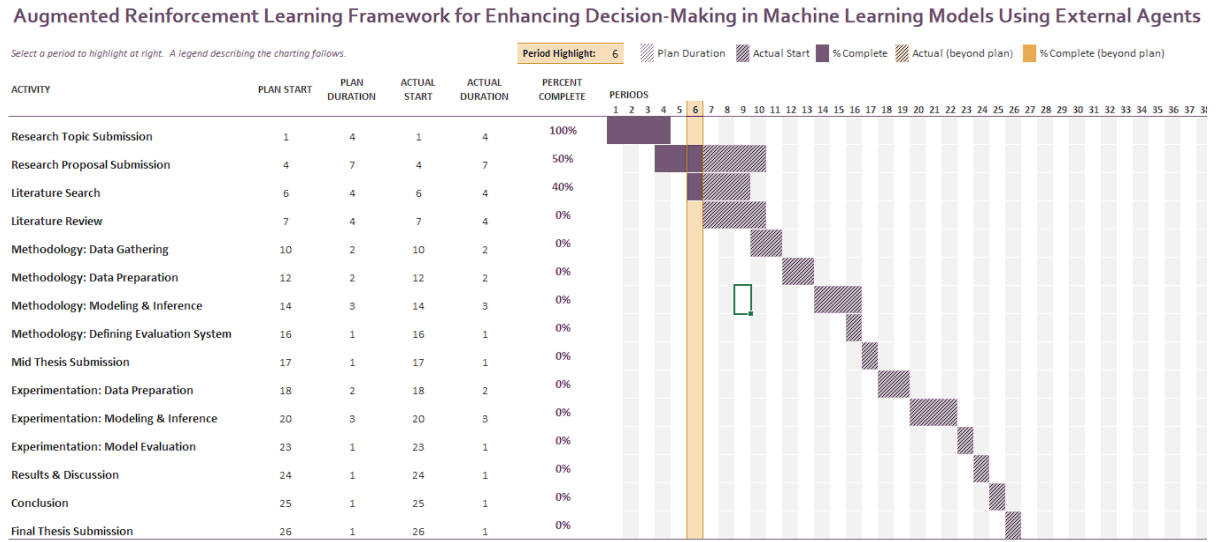

Figure 9.1.1 Research Project Gantt Chart

**Note:** *1 Period = 1Calendar Week*

### 9.2　　Risk Mitigation and Contingency Plan

The completion risks of the thesis work and its associated contingencies are listed below:

**Table 9.2.1 Risk and Contingency**

| Risk | Contingency |
|------|-------------|
| Timelines may be affected due to health or personal issues. | Buffer time allocation to each subtask in project management; inform LJMU/UpGrad administration, and ask for an extension. |
| GPUs are unavailable for data generation and Machine Learning model training. | Use cloud GPUs. A nano version of the YOLO Pre-Trained Model may be used for training. |

# APPENDIX C: ETHICS FORMS

No ethics form is required for this thesis because as stated in Chapter 3 earlier, all the data is being gathered via document synthesizer toolkit , the document generation code base is provided in chapter 3 as well as it is available at https://github.com/meetsandesh/synthetic_document_generator. The detailed description of the tool and its relevant code is provided in Chapter 3 and Chapter 4.